\documentclass{article}

\usepackage[numbers]{natbib}

\usepackage[final]{neurips_2022}

\usepackage[utf8]{inputenc} 
\usepackage[T1]{fontenc}    
\usepackage{xr-hyper}
\usepackage{hyperref}
\usepackage{url}            
\usepackage{amsfonts}       
\usepackage{nicefrac}       
\usepackage{microtype}
\usepackage{xcolor}

\usepackage{graphicx}
\usepackage{subfigure}
\usepackage{booktabs} 
\usepackage{stackengine} 
\usepackage{float} 
\usepackage{wrapfig} 

\usepackage{multirow}
\usepackage{array}
\newcolumntype{P}[1]{>{\centering\arraybackslash}p{#1}}

\usepackage{amsmath}
\mathchardef\mhyphen="2D 
\usepackage{dsfont} 
\newcommand{\probP}{\mathds{P}}
\newcommand{\given}{\:\vert\:} 
\newcommand{\stopgrad}[1]{\text{stop\_grad} (#1)}
\newcommand{\indicator}[1]{\mathds{1} (#1)}
\newcommand{\Expect}{\operatorname{\mathbb{E}}}
\newcommand{\Var}{\mathrm{Var}}
\DeclareMathOperator*{\plim}{plim}

\definecolor{medblue}{rgb}{0,0,.75}
\usepackage[noend]{algpseudocode}
\usepackage{algorithm,algorithmicx}
\algrenewcommand\alglinenumber[1]{\sf\tiny\color{medblue}{#1}\quad}
\algrenewcommand\algorithmicrequire{\textbf{Input:}}
\algrenewcommand\algorithmicensure{\textbf{Output:}}
\algdef{SE}[DOWHILE]{Do}{doWhile}{\algorithmicdo}[1]{\algorithmicwhile\ #1}%

\usepackage{amsmath}
\usepackage{amssymb}
\usepackage{mathtools}
\usepackage{amsthm}

\usepackage{enumitem} 

\usepackage[capitalize,noabbrev]{cleveref}

\theoremstyle{plain}

\theoremstyle{definition}

\theoremstyle{remark}

\usepackage[textsize=tiny]{todonotes}

\makeatletter
\newcommand*{\addFileDependency}[1]{
  \typeout{(#1)}
  \@addtofilelist{#1}
  \IfFileExists{#1}{}{\typeout{No file #1.}}
}
\makeatother

\newcommand{\myparagraph}[1]{\noindent\textbf{#1.}\,}

\def \figpath {figures/}

\title{TabNAS: Rejection Sampling for \\ Neural Architecture Search on Tabular Datasets}

\author{Chengrun Yang$^1$, Gabriel Bender$^1$, Hanxiao Liu$^1$, Pieter-Jan Kindermans$^1$, \\ \textbf{Madeleine Udell$^2$, Yifeng Lu$^1$, Quoc V. Le$^1$, Da Huang$^1$} \\
{\texttt{\{chengrun, gbender, hanxiaol, pikinder\}@google.com},}\\ {\texttt{udell@stanford.edu, \{yifenglu, qvl, dahua\}@google.com}} \\
[1ex]
$^1$ Google Research, Brain Team $^2$ Stanford University\\
}

\begin{document}
\maketitle
\begin{abstract}
The best neural architecture for a given machine learning problem 
depends on many factors: not only the complexity and structure of the dataset,
but also on resource constraints including latency, compute, energy consumption, etc. 
Neural architecture search (NAS) for tabular datasets is an important but under-explored problem.
Previous NAS algorithms designed for image search spaces incorporate resource constraints directly into the reinforcement learning (RL) rewards. 
However, for NAS on tabular datasets, this protocol often discovers suboptimal architectures.
This paper develops TabNAS, a new and more effective approach to handle resource constraints in tabular NAS
using an RL controller motivated by the idea of rejection sampling.
TabNAS immediately discards any architecture that violates the resource constraints
without training or learning from that architecture.
TabNAS uses a Monte-Carlo-based correction to the RL policy gradient update to account for this extra filtering step.
Results on several tabular datasets 
demonstrate the superiority of TabNAS over previous reward-shaping methods:
it finds better models that obey the constraints.
\end{abstract}

\section{Introduction}
\label{sec:intro}
To make a machine learning model better, one can scale it up. 
But larger networks are more expensive as measured by inference time, memory, energy, etc,
and these costs limit the application of large models: 
training is slow and expensive, 
and inference is often too slow to satisfy user requirements.

Many applications of machine learning in industry 
use tabular data, e.g., in finance, advertising and medicine. 
It was only recently that deep learning has achieved parity with classical tree-based models in these domains~\cite{gorishniy2021revisiting,kadra2021well}.
For vision, optimizing models for practical deployment often relies on Neural Architecture Search (NAS).
Most NAS literature targets convolutional networks on vision benchmarks~\cite{liu2018darts,cai2018proxylessnas,howard2019searching,ying2019bench}.
Despite the practical importance of tabular data, however, NAS research on this topic is quite limited~\cite{erickson2020autogluon,egele2021agebo}. (See Appendix~\ref{appsec:literature} for a more comprehensive literature review.)

Weight-sharing reduces the cost of NAS by training a \emph{SuperNet} that is the superset of all candidate architectures \cite{bender2018understanding}.
This trained SuperNet is then used to estimate the quality of each candidate architecture or \emph{child network} by allowing activations in only a subset of the components of the SuperNet and evaluating the model. 
Reinforcement learning (RL) has shown to efficiently find the most promising child networks~\cite{pham2018efficient,cai2018proxylessnas,bender2020can} for vision problems.

\begin{figure}
\centering
\subfigure{\includegraphics[width=.9\linewidth]{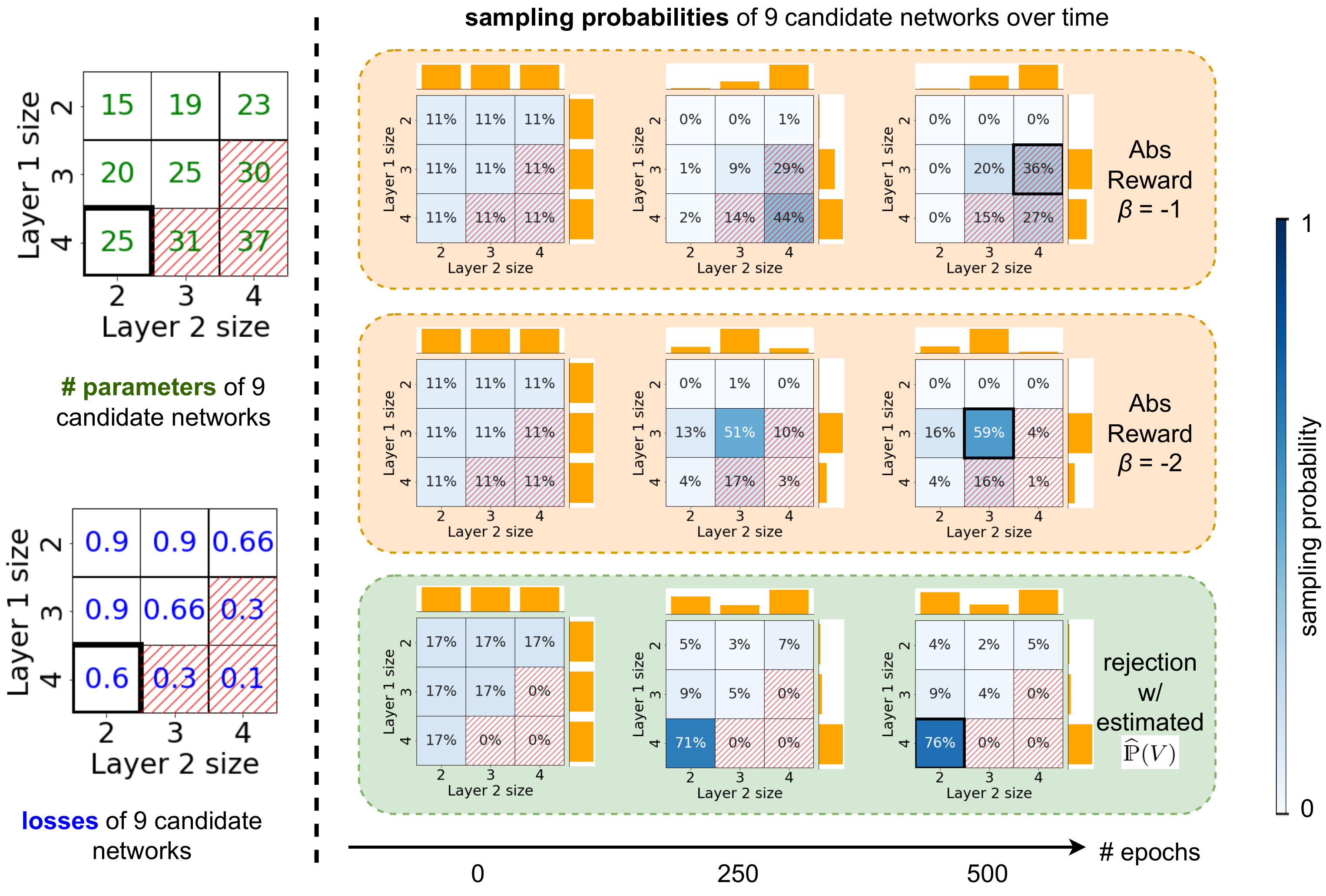}}
\caption{\textbf{A toy example for tabular NAS} in a 2-layer search space with a 2-dimensional input and a limit of 25 parameters.
Each cell represents an architecture.
The left half shows the number of parameters and loss of each candidate in the search space.
Infeasible architectures have red-striped cells. 
The bottom left cell (bold border) is the global optimum with size 4 for the first hidden layer and size 2 for the second. 
The right half shows the change of sampling probabilities in NAS with different RL rewards. 
Sampling probabilities are shown both as percentages in cells, and intensity indicated by the right colorbar.
Orange bars on the top and right sides show the (independent) sampling probability distributions of size candidates for individual layers.
With the Abs Reward, the sampling probability of each architecture is the product of sampling probabilities of each layer; with the rejection-based reward, the probability of an infeasible architecture is 0, and probabilities of feasible architectures are normalized to sum to 1.
At epoch 500, the cell squared in bold shows the architecture picked by the corresponding RL controller. 
RL with the Abs Reward $Q(y) + \beta |T(y) / T_0 - 1|$ proposed in TuNAS~\cite{bender2020can} either converges to a feasible but suboptimal architecture ($\beta=-2$, middle row) or violates the resource constraint ($\beta=-1$, top row). 
Other latency-aware reward functions show similar failures. 
In contrast, TabNAS converges to the optimum (bottom row).
}
\label{fig:toy_example}
\end{figure}

In our experiments, we show that a direct application of approaches designed for vision to tabular data often fails. 
For example, the TuNAS \cite{bender2020can} approach from vision struggles to find the optimal architectures for tabular datasets (see experiments). 
The failure is caused by the interaction of the search space and the factorized RL controller. 
To understand why, consider the following toy example with 2 layers, illustrated in Figure~\ref{fig:toy_example}. 
For each layer, 
we can choose a layer size of $2$, $3$, or $4$, 
and the maximum number of parameters is set to 25. 
The optimal solution is to set the size of the first hidden layer to 4 and the second to 2. 
Finding this solution with RL is difficult with a cost penalty approach. 
The RL controller is initialized with uniform probabilities. 
As a result, it is quite likely that the RL controller will initially be penalized heavily when choosing option 4 for the first layer, since two thirds of the choices for the second layer will result in a model that is too expensive. 
As a result, option 4 for the first layer is quickly discarded by the RL controller and we get stuck in a local optimum. 

This co-adaptation problem is caused by the fact that existing NAS methods for computer vision often use factorized RL controllers, which force all choices to be be made independently. While factorized controllers can be optimized easily and are parameter-efficient, they cannot capture all of the nuances in the loss landscape. A solution to this could be to use a more complex model such as an LSTM (e.g., ~\cite{pham2018efficient,cai2018efficient}). However, LSTMs are often much slower to train and are far more difficult to tune.

Our proposed method, TabNAS, uses a solution inspired by rejection sampling. 
It updates the RL controller only when the sampled model satisfies the cost constraint. 
The RL controller is then discouraged from sampling poor models within the cost constraint and encouraged to sample the high quality models.
Rather than penalizing models that violate the constraints, the controller silently discards them. This trick allows the RL controller to see the true constrained loss landscape, in which having some large layers is beneficial,
allowing TabNAS to efficiently find global (not just local) optima for tabular NAS problems.
Our contributions can be summarized as follows:
\begin{itemize}[leftmargin=2em,topsep=0pt,partopsep=1ex,parsep=0ex]
\item We identify failure cases of existing resource-aware NAS methods on tabular data and provide evidence this failure is due to the cost penalty in the reward together with the factorized space.
\item We propose and evaluate an alternative: a rejection sampling mechanism that ensures the RL controller only selects architectures that satisfy resource constraint.
This extra rejection step allows the RL controller to explore parts of the search space that would otherwise be overlooked.
\item The rejection mechanism also introduces a systematic bias into the RL gradient updates, which can skew the results.
To compensate for this bias, we introduce a theoretically motivated and empirically effective correction into the gradient updates. 
This correction can be computed exactly for small search spaces
and efficiently approximated by Monte-Carlo sampling otherwise. 
\item We show the resulting method, TabNAS, automatically learns whether a bottleneck structure is needed in an optimal architecture, and if needed, where to place the bottleneck in the network.
\end{itemize}

\begin{figure}
\begin{minipage}[t]{.48\linewidth}
\centering
\subfigure{\includegraphics[width=.9\linewidth]{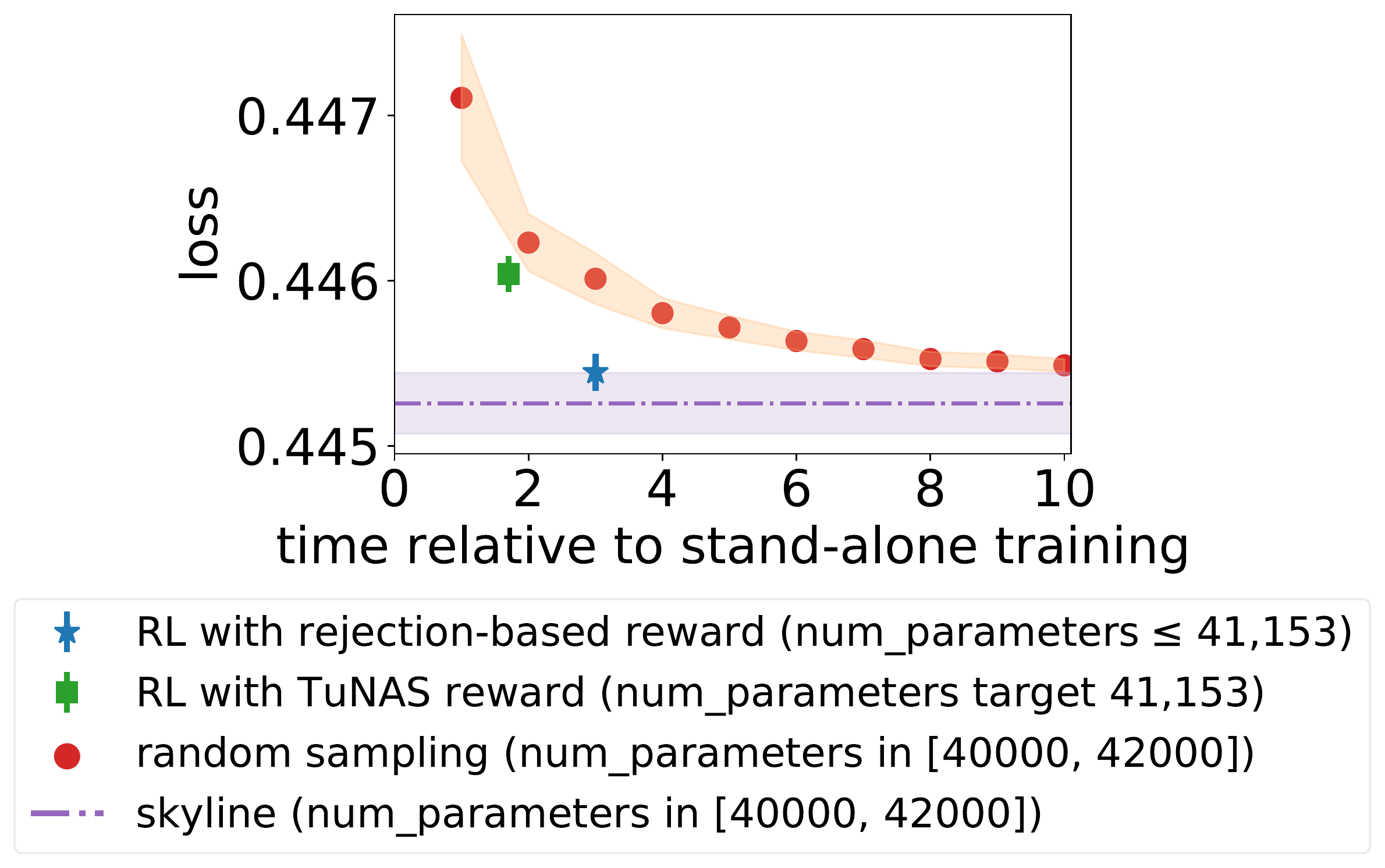}}
\vspace{-1em}
\caption{TabNAS reward distributionally outperforms random search and resource-aware Abs Reward on the Criteo dataset within a 3-layer search space.
All error bars and shaded regions are 95\% confidence intervals.
The x axis is the time relative to train time for a single architecture. 
The y axis is the validation loss.
More details in Appendix~\ref{appsec:distributional_performance_comparison_details}.
}
\label{fig:comparison_with_random_criteo_3_layers}
\end{minipage}
\hspace{.02\linewidth}
\begin{minipage}[t]{.48\linewidth}
\centering
\subfigure{\includegraphics[width=\linewidth]{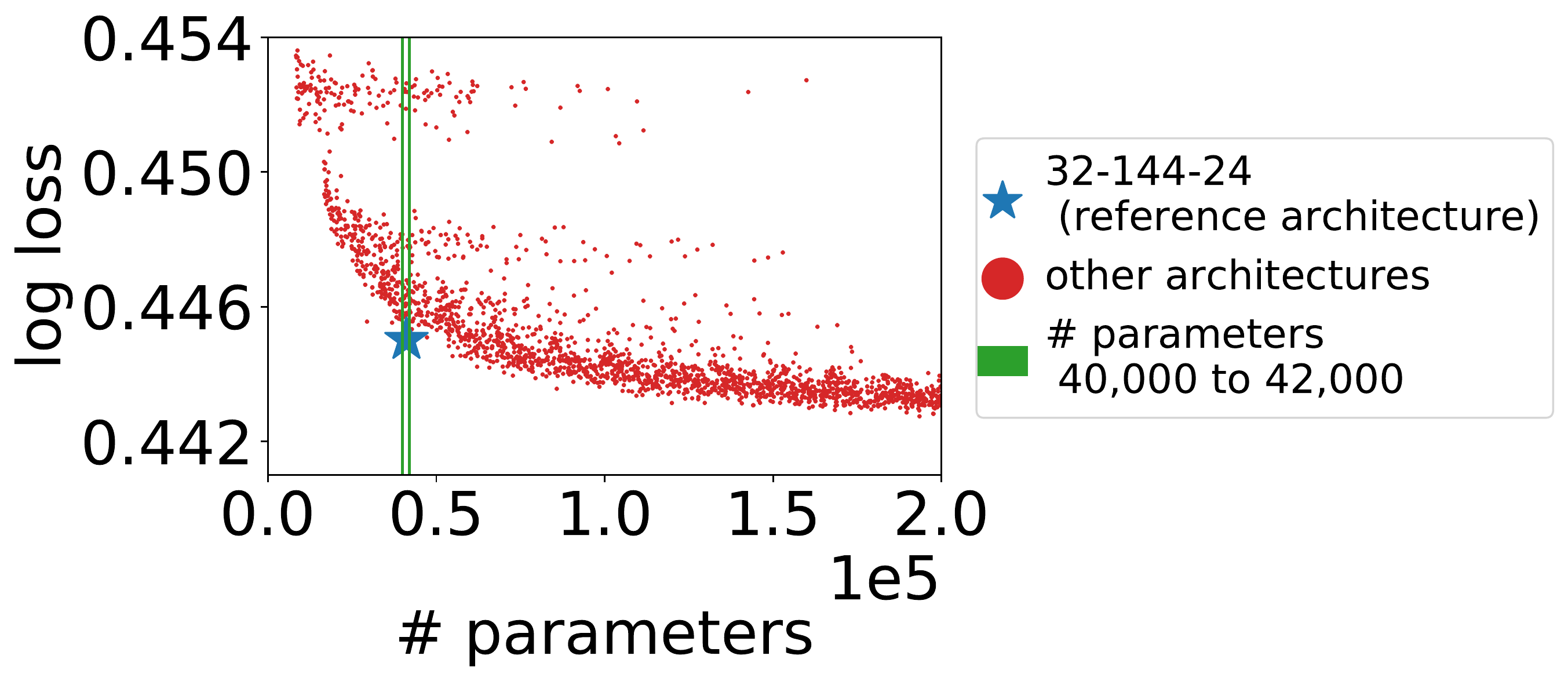}}
\caption{
Validation loss (logistic) vs.~number of parameters on Criteo with a 3-layer search space.
The standard deviation (std) of architecture performance for different runs is 0.0002, so architectures whose performance difference is larger than $2$std are qualitatively different with high probability. 
The search space and Pareto-optimal architectures are shown in Appendix~\ref{appsec:tradeoff_plot_details}.
}
\label{fig:tradeoff_criteo_3_layers}
\end{minipage}
\end{figure}

These contributions form TabNAS, our RL-based weight-sharing NAS with rejection-based reward. TabNAS robustly and efficiently finds a feasible architecture with optimal performance within the resource constraint.
Figure~\ref{fig:comparison_with_random_criteo_3_layers} shows an example.

\section{Notation and terminology}
\label{sec:term}
\myparagraph{Math basics}
We define $[n] = \{1, \ldots, n\}$ for a positive integer $n$.
With a Boolean variable $\mathcal{X}$, the indicator function $\indicator{\mathcal{X}}$ equals 1 if $\mathcal{X}$ is true, and 0 otherwise.
$|S|$ denotes the cardinality of a set $S$; $\stopgrad{f}$ denotes the constant value (with gradient 0) corresponding to a differentiable quantity $f$, and is equivalent to \texttt{tensorflow.stop\_gradient(f)} in TensorFlow~\cite{tensorflow2015-whitepaper} or \texttt{f.detach()} in PyTorch~\cite{paszke2019pytorch}.
$\subseteq$ and $\subset$ denote subset and strict subset, respectively.
$\nabla$ denotes the gradient with respect to the variable in the context.

\myparagraph{Weight, architecture, and hyperparameter}
We use \emph{weights} to refer to the parameters of the neural network.
The \emph{architecture} of a neural network is the structure of how nodes are connected; examples of architectural choices are hidden layer sizes and activation types.
\emph{Hyperparameters} are the non-architectural parameters that control the training process of either stand-alone training or RL, including learning rate, optimizer type, optimizer parameters, etc. 

\myparagraph{Neural architecture}
A neural network with specified architecture and hyperparameters is called a \emph{model}. 
We only consider fully-connected feedforward networks (FFNs) in this paper, since they can already achieve SOTA performance on tabular datasets~\cite{kadra2021well}. 
The number of hidden nodes after each weight matrix and activation function is called a \emph{hidden layer size}. 
We denote a single network in our search space with hyphen-connected choices. 
For example, when searching for hidden layer sizes, in the space of 3-hidden-layer ReLU networks, 32-144-24 denotes the candidate where the sizes of the first, second and third hidden layers are 32, 144 and 24, respectively.
We only search for ReLU networks; for brevity, we will not mention the activation function type in the sequel.

\myparagraph{Loss-resource tradeoff and reference architectures}
In the hidden layer size search space, the validation loss in general decreases with the increase of the number of parameters, giving the loss-resource tradeoff (e.g., Figure~\ref{fig:tradeoff_criteo_3_layers}). 
Here loss and number of parameters serve as two \emph{costs} for NAS.
Thus there are Pareto-optimal models that achieve the smallest loss among all models with a given bound on the number of parameters.
With an architecture that outperforms others with a similar or fewer number of parameters, we do resource-constrained NAS with the number of parameters of this architecture as the resource target or constraint.
We call this architecture the \emph{reference architecture} (or \emph{reference}) of NAS, and its performance the \emph{reference performance}.
We do NAS with the goal of \emph{matching} (the size and performance of) the reference. 
Note that the RL controller only has knowledge of the number of parameters of the reference, and is not informed of its hidden layer sizes.

\myparagraph{Search space}
When searching $L$-layer networks, we use capital letters like $X=X_1 \mhyphen \dots \mhyphen X_L$ to denote the random variable of sampled architectures, in which $X_i$ is the random variable for the size of the $i$-th layer. 
We use lowercase letters like $x = x_1 \mhyphen \dots \mhyphen x_L$ to denote an architecture sampled from the distribution over $X$, in which $x_i$ is an instance of the $i$-th layer size.
When there are multiple samples drawn, we use a bracketed superscript to denote the index over samples: $x^{(k)}$ denotes the $k$-th sample.
The search space $S = \{s_{ij}\}_{i \in [L], j \in [C_i]}$ has $C_i$ choices for the $i$-th hidden layer, in which $s_{ij}$ is the $j$-th choice for the size of the $i$-th hidden layer: for example, when searching for a one-hidden-layer network with size candidates \{5, 10, 15\}, we have $s_{13} = 15$.

\myparagraph{Reinforcement learning}
The RL algorithm learns the set of logits $\{\ell_{ij}\}_{i \in [L], j \in [C_i]}$, in which $\ell_{ij}$ is the logit associated with the $j$-th choice for the $i$-th hidden layer.
With a fully factorized distribution of layer sizes (we learn a separate distribution for each layer), the probability of sampling the $j$-th choice for the $i$-th layer $p_{ij}$ is given by the SoftMax function: $p_{ij} = \exp (\ell_{ij}) / \sum_{j \in [C_i]} \exp (\ell_{ij})$.
In each RL step, we sample an architecture $y$ to compute the single-step RL objective $J(y)$, and update the logits with $\nabla J(y)$: an unbiased estimate of the gradient of the RL value function.

\myparagraph{Resource metric and number of parameters}
We use the number of parameters, which can be easily computed for neural networks, as a cost metric in this paper. However, our approach does not depend on the specific cost used, and can be easily adapted to other cost metrics. 

\section{Methodology}
\label{sec:meth}

Our NAS methodology can be decomposed into three main components: weight-sharing with layer warmup, REINFORCE with one-shot search, and Monte Carlo (MC) sampling with rejection.

As an overview, our method starts with a SuperNet, which is a network that layer-wise has width equal to the largest choice within the search space.
We first stochastically update the weights of the entire SuperNet to ``warm up'' over the first 25\% of search epochs.
Then we alternate between updating the shared model weights (which are used to estimate the quality of different child models) and the RL controller (which focuses the search on the most promising parts of the space).
In each iteration, we first sample a child network from the current layer-wise probability distributions and update the corresponding weights within the SuperNet (weight update). We then sample another child network to update the layerwise logits that give the probability distributions (RL update).
The latter RL update is only performed if the sampled network is feasible, in which case we use rejection with MC sampling to update the logits with a sampling probability conditional on the feasible set.

To avoid overfitting, we split the labelled portion of a dataset into training and validation splits.
Weight updates are carried out on the training split; RL updates are performed on the validation split.

\subsection{Weight sharing with layer warmup}
The weight-sharing approach has shown success on various computer vision tasks and NAS benchmarks~\cite{pham2018efficient,bender2018understanding,cai2018proxylessnas,bender2020can}. 
To search for an FFN on tabular datasets, we build a SuperNet where the size of each hidden layer is the largest value in the search space. 
Figure~\ref{fig:supernet_illustration} shows an example.
When we sample a child network with a hidden layer size $\ell_i$ smaller than the SuperNet, we only use the first $\ell_i$ hidden nodes in that layer to compute the output in the forward pass and the gradients in the backward pass.
Similarly, in RL updates, only the weights of the child network are used to estimate the quality reward that is used to update logits.

In weight-sharing NAS, warmup helps to ensure that the SuperNet weights are sufficiently trained to properly guide the RL updates \cite{bender2020can}. 
With probability $p$, we train all weights of the SuperNet, and with probability $1-p$ we only train the weights of a random child model.
When we run architecture searches for FFNs, we do warmup in the first 25\% epochs, during which the probability $p$ linearly decays from 1 to 0 (Figure~\ref{fig:warmup_prob}).
The RL controller is disabled during this period.

\subsection{One-shot training and REINFORCE}
\label{sec:meth_reinforce}
We do NAS on FFNs with a REINFORCE-based algorithm.
Previous works have used this type of algorithm to search for convolutional networks on vision tasks~\cite{tan2019mnasnet, cai2018proxylessnas, bender2020can}.
When searching for $L$-layer FFNs, we learn a separate probability distribution over $C_i$ size candidates for each layer.
The distribution is given by $C_i$ logits via the SoftMax function. 
Each layer has its own independent set of logits.
With $C_i$ choices for the $i$th layer, where $i=1, 2, \ldots, L$, there are $\prod_{i \in [L]} C_i$ candidate networks in the search space but only $\sum_{i \in [L]} C_i$ logits to learn.
This technique significantly reduces the difficulty of RL and make the NAS problem practically tractable~\cite{cai2018proxylessnas,bender2020can}. 

The REINFORCE-based algorithm trains the SuperNet weights and learns the logits $\{\ell_{ij}\}_{i \in [L], j \in [C_i]}$ that give the sampling probabilities $\{\ell_{ij}\}_{i \in [L], j \in [C_i]}$ over size candidates by alternating between weight and RL updates.
In each iteration, we first sample a child network $x$ from the SuperNet and compute its training loss in the forward pass. 
Then we update the weights in $x$ with gradients of the training loss computed in the backward pass.
This weight update step trains the weights of $x$. 
The weights in architectures with larger sampling probabilities are sampled and thus trained more often.
We then update the logits for the RL controller by sampling a child network $y$ that is independent of the network $x$ from the same layerwise distributions, computing the quality reward $Q(y)$ as $1 - \textit{loss}(y)$ on the validation set, and then updating the logits with the gradient of $J(y) = \stopgrad{Q(y) - \bar{Q}} \log \probP(y)$: the product of the advantage of $y$'s reward over past rewards (usually an exponential moving average) and the log-probability of the current sample. 

The alternation creates a positive feedback loop that trains the weights and updates the logits of the large-probability child networks; thus the layer-wise sampling probabilities gradually converge to more deterministic distributions, under which one or several architectures are finally selected. 

Details of a resource-oblivious version is shown as Appendix~\ref{appsec:pseudocode} Algorithm~\ref{alg:reinforce}, which does not take into account a resource constraint.
In Section~\ref{sec:meth_mc}, we show an algorithm that combines Monte-Carlo sampling with rejection sampling, which serves as a subroutine of Algorithm~\ref{alg:reinforce} by replacing the probability in $J(y)$ with a conditional version.

\subsection{Rejection-based reward with MC sampling}
\label{sec:meth_mc}

Only a subset of the architectures in the search space $S$ will satisfy resource constraints; $V$ denotes this set of feasible architectures.
To find a feasible architecture, a resource target $T_0$ is often used in an RL reward. Given an architecture $y$, a resource-aware reward combines its quality $Q(y)$ and resource consumption $T(y)$ into a single reward. 
MnasNet~\cite{tan2019mnasnet} proposes the rewards $Q(y) (T(y) / T_0)^\beta$ and $Q(y) \max\{1, (T(y) / T_0)^\beta\}$ while TuNAS~\cite{bender2020can} proposes the absolute value reward (or Abs Reward) $Q(y) + \beta |T(y) / T_0 - 1|$. 
The idea behind is to encourage models with high quality with respect the resource target. 
In these rewards $\beta$ is a hyperparameter that needs careful tuning.

\begin{figure}
\begin{minipage}[t]{.48\linewidth}
\centering
\subfigure{\includegraphics[width=.51\linewidth]{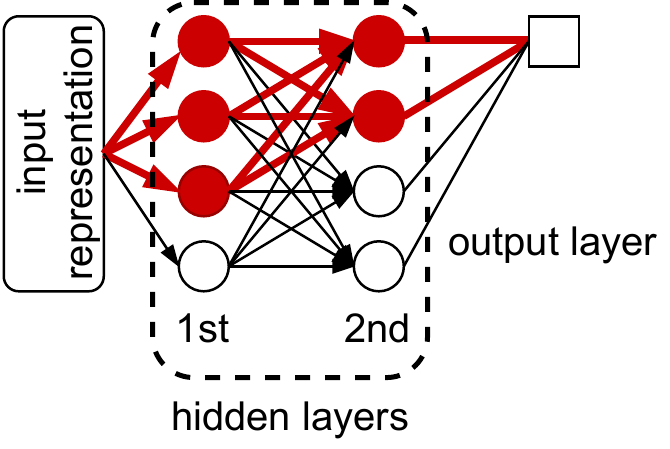}}
\caption{Illustration of weight-sharing on two-layer FFNs for a binary classification task. 
Edges denote weights; arrows at the end of lines denote ReLU activations; circles denote hidden nodes; the square in the output layer denotes the output logit.
The size of each hidden layer can be one of \{2, 3, 4\}, thus the SuperNet is a two-layer FFN with size 4-4. 
At this moment, the controller picks the child network 3-2, thus only the first 3 hidden nodes in the first hidden layer and the first 2 hidden nodes in the second hidden layer, together with the connected edges (in red), are enabled to compute the output logits.
}
\label{fig:supernet_illustration}
\end{minipage}
\hspace{.02\linewidth}
\begin{minipage}[t]{.48\linewidth}
\centering
\subfigure[Warmup probability]{\label{fig:warmup_prob}\includegraphics[width=.45\linewidth]{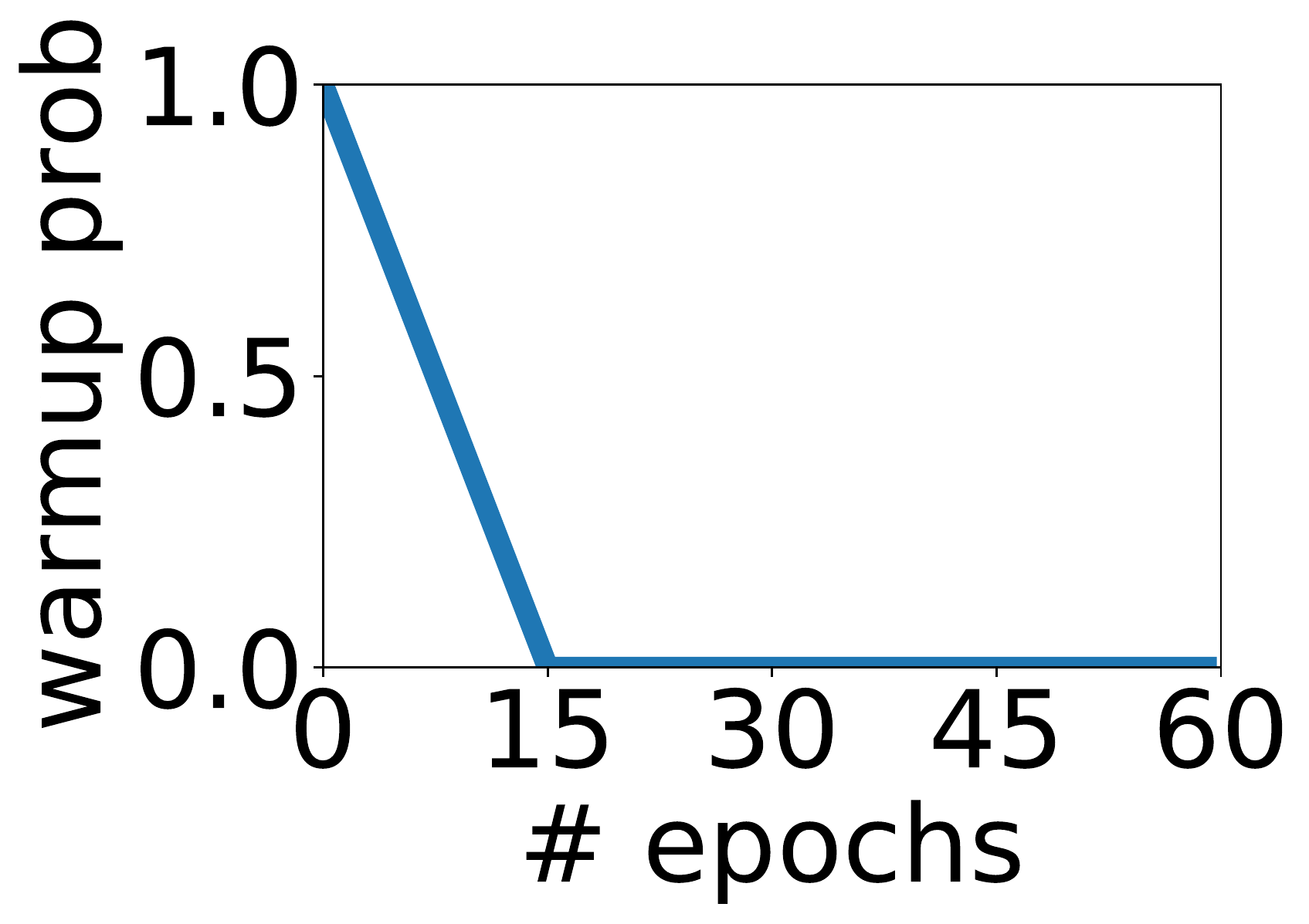}}
\hspace{.03\linewidth}
\subfigure[Valid probability]{\label{fig:valid_prob}\includegraphics[width=.45\linewidth]{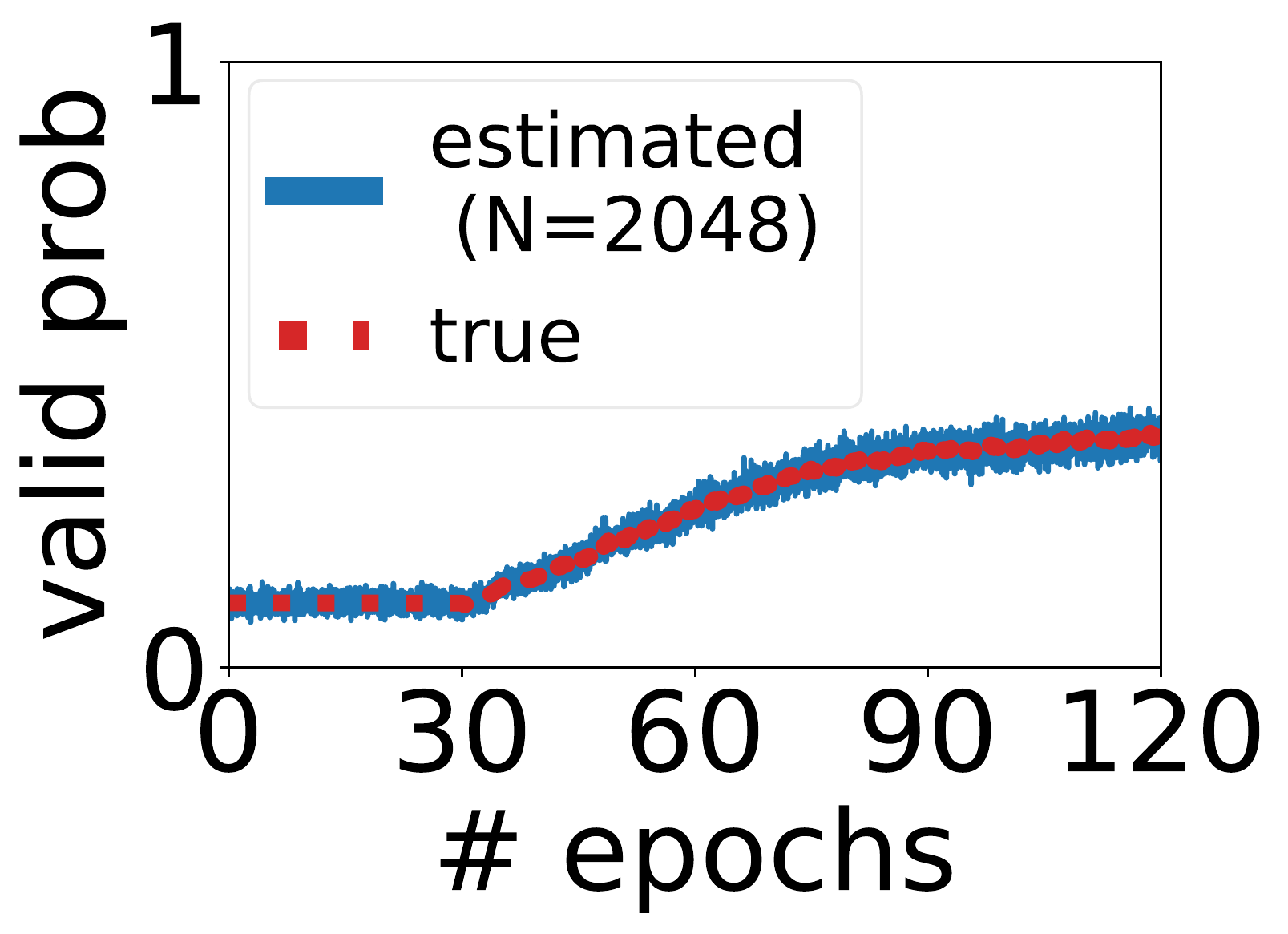}}

\caption{Examples of layer warmup and valid probabilities. 
Figure~\subref{fig:warmup_prob} shows our schedule: linearly decay from 1 to 0 in the first 25\% epochs.
Figure~\subref{fig:valid_prob} shows an example of the change of true and estimated valid probabilities ($\probP(V)$ and $\widehat{\probP}(V)$) in a successful search, with 8,000 architectures in the search space and the number of MC samples $N=1024$.
Both probabilities are (nearly) constant during warmup before RL starts, then increase after RL starts because of rejection sampling.
}
\label{fig:warmup_and_valid_probs}
\end{minipage}
\end{figure}

We find that on tabular data, RL controllers using these resource-aware rewards above can struggle to discover high quality structures. 
Figure~\ref{fig:toy_example} shows a toy example in the search space in Figure~\ref{fig:supernet_illustration}, in which we know the validation losses of each child network and only train the RL controller for 500 steps.
The optimal network is 4-2 among architectures with number of parameters no more than 25, but the RL controller rarely chooses it. 
In Section~\ref{sec:failure_mode}, we show examples on real datasets.

This phenomenon reveals a gap between the true distribution we want to sample from and the distributions obtained by sampling from this factorized search space:
\begin{itemize}[leftmargin=2em,topsep=0pt,partopsep=1ex,parsep=0ex]
\item We \emph{only} want to sample from the set of feasible architectures $V$, whose distribution is $\{\probP(y \given y \in V)\}_{y \in V}$.
The resources (e.g., number of parameters) used by an architecture, and thus its feasibility, is determined jointly by the sizes of all layers. 
\item On the other hand, the factorized search space learns a separate (independent) probability distribution for the choices of each layer. While this distribution is efficient to learn, independence between layers discourages an RL controller with a resource-aware reward from choosing a bottleneck structure.
A bottleneck requires the controller to select large sizes for some layers and small for others. But decisions for different layers are made independently, and both very large and very small layer sizes, considered independently, have poor expected rewards:
small layers are estimated to perform poorly, while large layers easily exceed the resource constraints.
\end{itemize}

To bridge the gap and efficiently learn layerwise distributions that take into account the architecture feasibility, we propose a rejection-based RL reward for Algorithm~\ref{alg:reinforce}.
We next sketch the idea; detailed pseudocode is provided as Algorithm~\ref{alg:mc} in Appendix~\ref{appsec:pseudocode}.

REINFORCE optimizes a set of logits $\{\ell_{ij}\}_{i \in [L], j \in [C_i]}$ which define a probability distribution $p$ over architectures. 
In the original algorithm, we sample a random architecture $y$ from $p$ and then estimate its quality $Q(y)$. 
Updates to the logits $\ell_{ij}$ take the form $\ell_{ij} \gets \ell_{ij} + \eta \frac{\partial}{\partial \ell_{ij}} J(y)$, where $\eta$ is the learning rate, $\overline{Q}$ is a moving average of recent rewards, and
$J(y) = \stopgrad{Q(y) - \overline{Q}} \cdot \log \mathbb{P}(y)$.
If $y$ is better (worse) than average, then $Q(y) - \overline{Q}$ will be positive (negative), so the REINFORCE update will increase (decrease) the probability of sampling the same architecture in the future.

In our new REINFORCE variant, motivated by rejection sampling, we do not update the logits when $y$ is infeasible. 
When $y$ is feasible, we replace the probability $\probP(y)$ in the REINFORCE update equation with the conditional probability $\probP(y \given y \in V) = \probP(y) / \probP(y \in V)$. So $J(y)$ becomes
\begin{equation}
J(y) = \stopgrad{Q(y) - \overline{Q}} \cdot \log \left[ \probP(y) / \probP(y \in V) \right].
\end{equation}
We can compute the probability of sampling a feasible architecture $\probP(V) := \probP(y \in V)$ exactly when the search space is small, 
but this computation is too expensive when the space is large.
Instead, we replace the exact probability $\probP(y)$ with a differential approximation $\widehat{\probP}(y)$ obtained with Monte-Carlo (MC) sampling.
In each RL step, we sample $N$ architectures $\{z^{(k)}\}_{k \in [N]}$ within the search space with a proposal distribution $q$ and estimate $\probP(V)$ as
\begin{equation}
\widehat{\probP}(V) = \frac{1}{N} \sum_{k \in [N]} \frac{p^{(k)}}{q^{(k)}} \cdot \indicator{z^{(k)} \in V}.
\end{equation}
For each $k \in [N]$, $p^{(k)}$ is the probability of sampling $z^{(k)}$ with the factorized layerwise distributions and so is differentiable with respect to the logits. In contrast, $q^{(k)}$ is the probability of sampling $z^{(k)}$ with the proposal distribution, and is therefore non-differentiable.

$\widehat{\probP}(V)$ is an unbiased and consistent estimate of $\probP(V)$; $\nabla \log [\probP(y) / \widehat{\probP}(V)]$ is a consistent estimate of $\nabla \log[\probP(y \given y \in V)]$ (Appendix~\ref{appsec:proofs}).
A larger $N$ gives better results (Appendix~\ref{appsec:hyperparameter_tuning}); in experiments, we need smaller than the size of the sample space to get a faithful estimate (Figure~\ref{fig:valid_prob}, Appendix~\ref{appsec:more_failure_cases} and~\ref{appsec:ablation}) because neighboring RL steps can correct the estimates of each other.
We set $q = \stopgrad{p}$ in experiments for convenience: use the current distribution over architectures for MC sampling. 
Other distributions that have a larger support on $V$ may be used to reduce sampling variance (Appendix~\ref{appsec:proofs}). 

At the end of NAS, we pick as our final architecture the layer sizes with largest sampling probabilities if the layerwise distributions are deterministic, 
or sample from the distributions $m$ times and pick $n$ feasible architectures with the largest number of parameters if not. 
Appendix~\ref{appsec:pseudocode} Algorithm~\ref{alg:sample_to_get_solution} provides the full details.
We find $m=500$ and $n \leq 3$ suffice to find an architecture that matches the reference (optimal) architecture in our experiments.

In practice, the distributions often (almost) converge after twice the number of epochs used to train a stand-alone child network. Indeed the distributions are often useful after training the same number of epochs in that the architectures found by Algorithm~\ref{alg:sample_to_get_solution} are competitive. 
Figure~\ref{fig:toy_example} shows TabNAS finds the best feasible architecture, 4-2, in our toy example, using $\widehat{\probP}(V)$ estimated by MC sampling.

\section{Experimental results}
\label{sec:experiments}
Our implementation can be found at \url{https://github.com/google-research/tabnas}.
We ran all experiments using TensorFlow on a Cloud TPU v2 with 8 cores. 
We use a 1,027-dimensional input representation for the Criteo dataset and 180 features for Volkert\footnote{Our paper takes these features as given.
It is worth noting that methods proposed in feature engineering works like~\cite{khawar2020autofeature} and~\cite{liu2020autofis} are complementary to and can work together with TabNAS.}.
The best architectures in our FFN search spaces already produce near-state-of-the-art results; details in Appendix~\ref{appsec:more_details_real_datasets}.
More details of experiment setup and results in other search spaces can be found in Appendix~\ref{appsec:experiment_setup} and~\ref{appsec:more_failure_cases}.
Appendix~\ref{appsec:tabulated_performance_of_rewards_on_datasets} tabulates the performance of all RL rewards on all tabular datasets in our experiments.
Appendix~\ref{appsec:comparison_with_bo_and_es} shows a comparison with Bayesian optimization and evolutionary search in similar settings;
Ablation studies in Appendix~\ref{appsec:ablation} show TabNAS components collectively deliver desirable results; Appendix~\ref{appsec:hyperparameter_tuning} shows TabNAS has easy-to-tune hyperparameters.

\subsection{When do previous RL rewards fail?}
\label{sec:failure_mode}

Section~\ref{sec:meth_mc} discussed the resource-aware RL rewards and highlighted a potential failure case.
In this section, we show several failure cases of three resource-aware rewards, $Q(y) (T(y) / T_0)^\beta$, $Q(y) \max\{1, (T(y) / T_0)^\beta\}$, and the Abs Reward $Q(y) + \beta |T(y) / T_0 - 1|$, on our tabular datasets.

\subsubsection{Criteo -- 3 layer search space}
\label{sec:failure_criteo_3_layer}
We use the 32-144-24 reference architecture (41,153 parameters). 
Figure~\ref{fig:tradeoff_criteo_3_layers} gives an overview of the costs and losses of all architectures in the search space. 
The search space requires us to choose one of 20 possible sizes for each hidden layer; details in Appendix~\ref{appsec:more_failure_cases}. 
The search has $1.7\times$ the cost of a stand-alone training run.

\begin{figure}[t]
\centering
\subfigure[Layer 1 (finally 32)]{\label{fig:sampling_prob_heatmap_layer_1_criteo_3_layers_autoflow}\includegraphics[width=.23\linewidth]{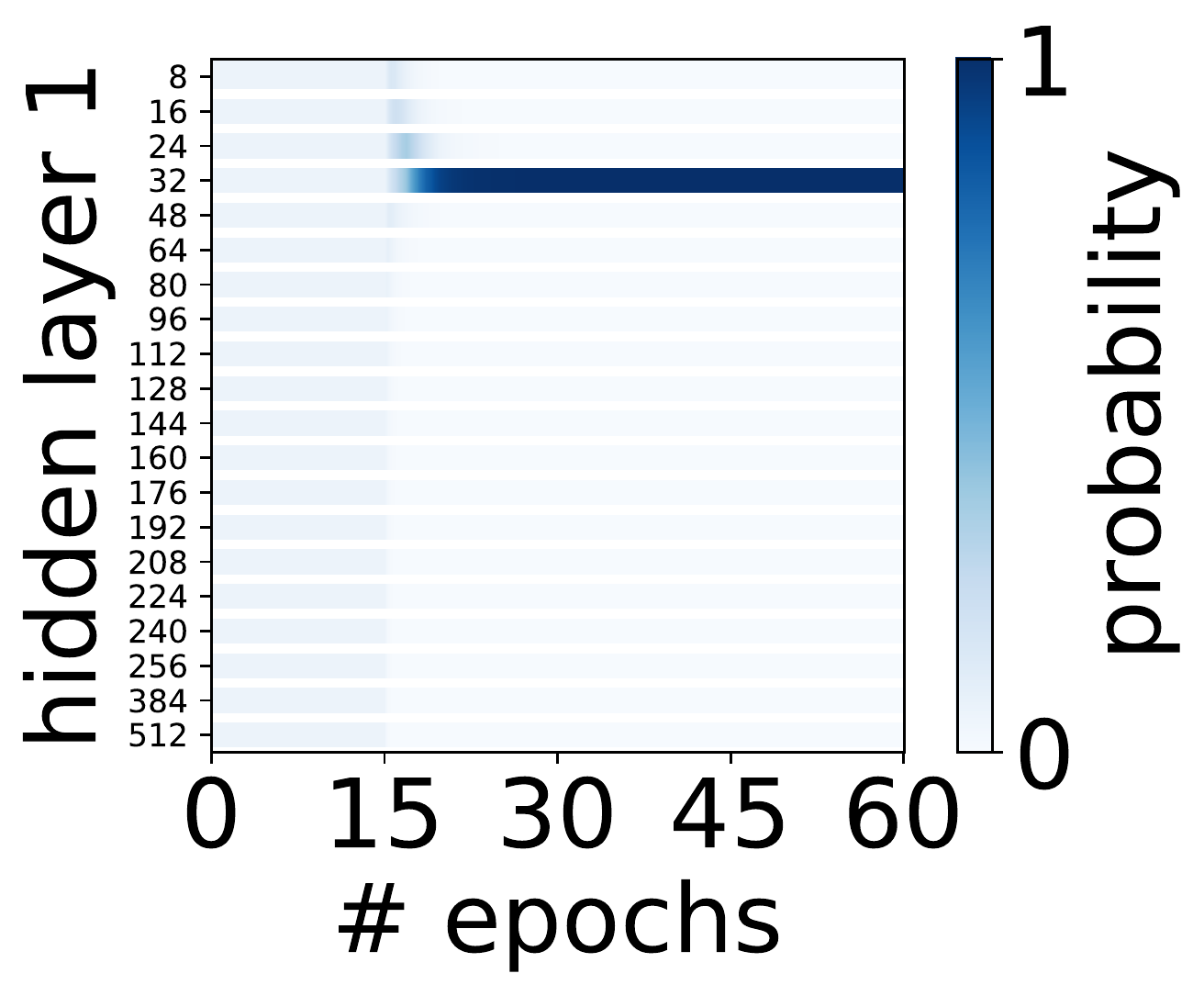}}
\hspace{.01\linewidth}
\subfigure[Layer 2 (finally 64)]{\label{fig:sampling_prob_heatmap_layer_2_criteo_3_layers_autoflow}\includegraphics[width=.23\linewidth]{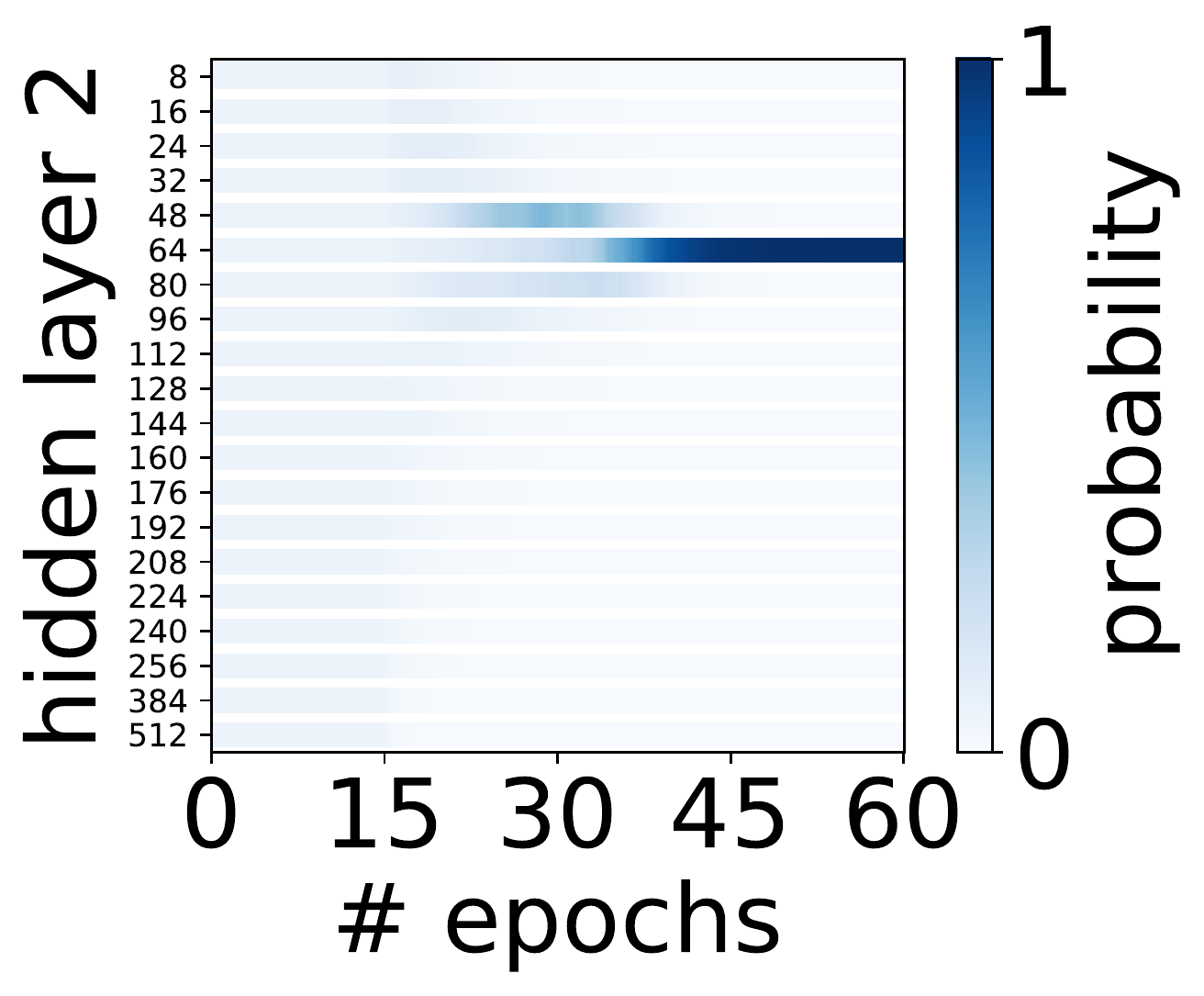}}
\hspace{.01\linewidth}
\subfigure[Layer 3 (finally 96)]{\label{fig:sampling_prob_heatmap_layer_3_criteo_3_layers_autoflow}\includegraphics[width=.23\linewidth]{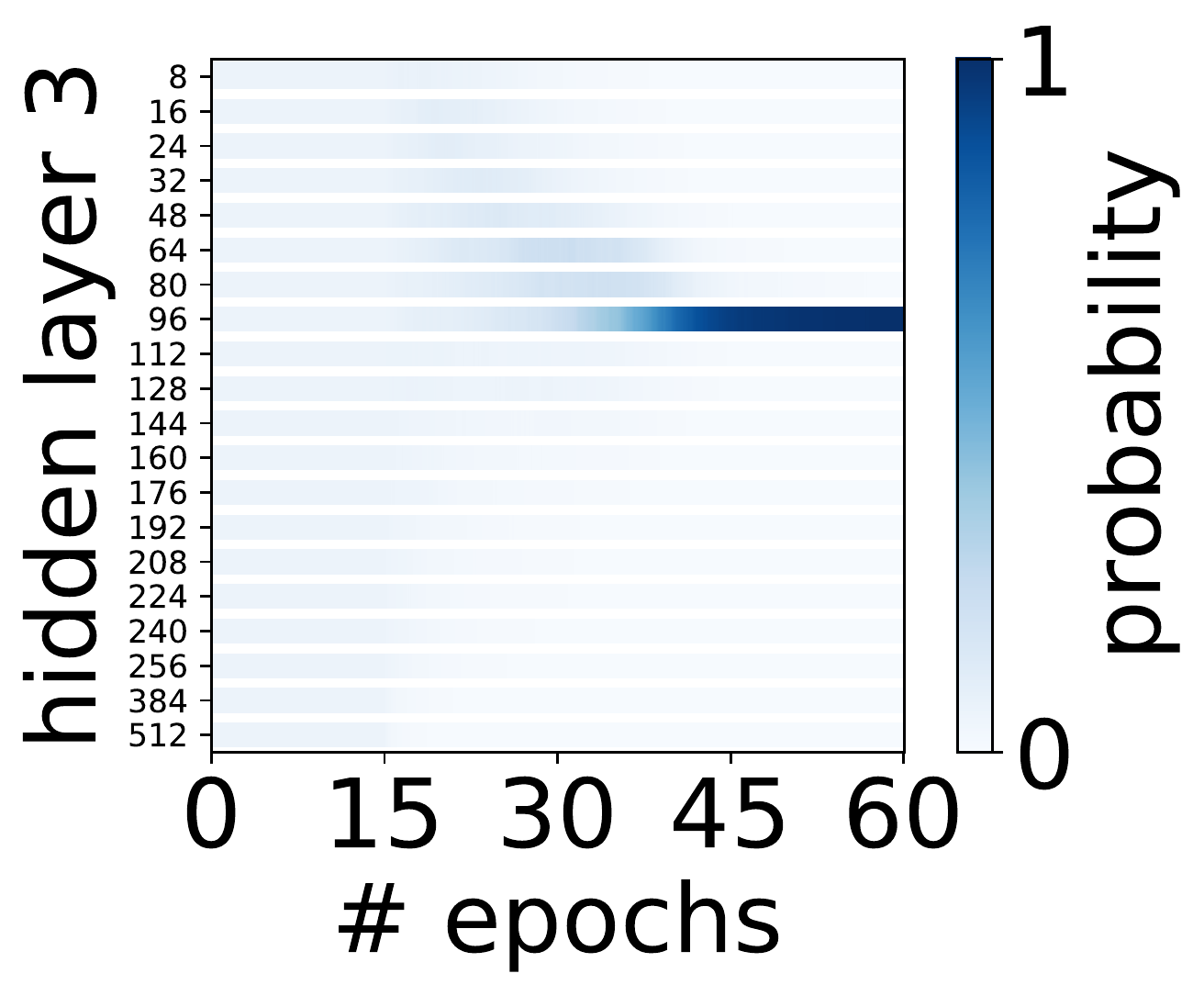}}
\hspace{.01\linewidth}
\subfigure[retrain performance]{\label{fig:retrain_3_layers_autoflow}\includegraphics[width=.23\linewidth]{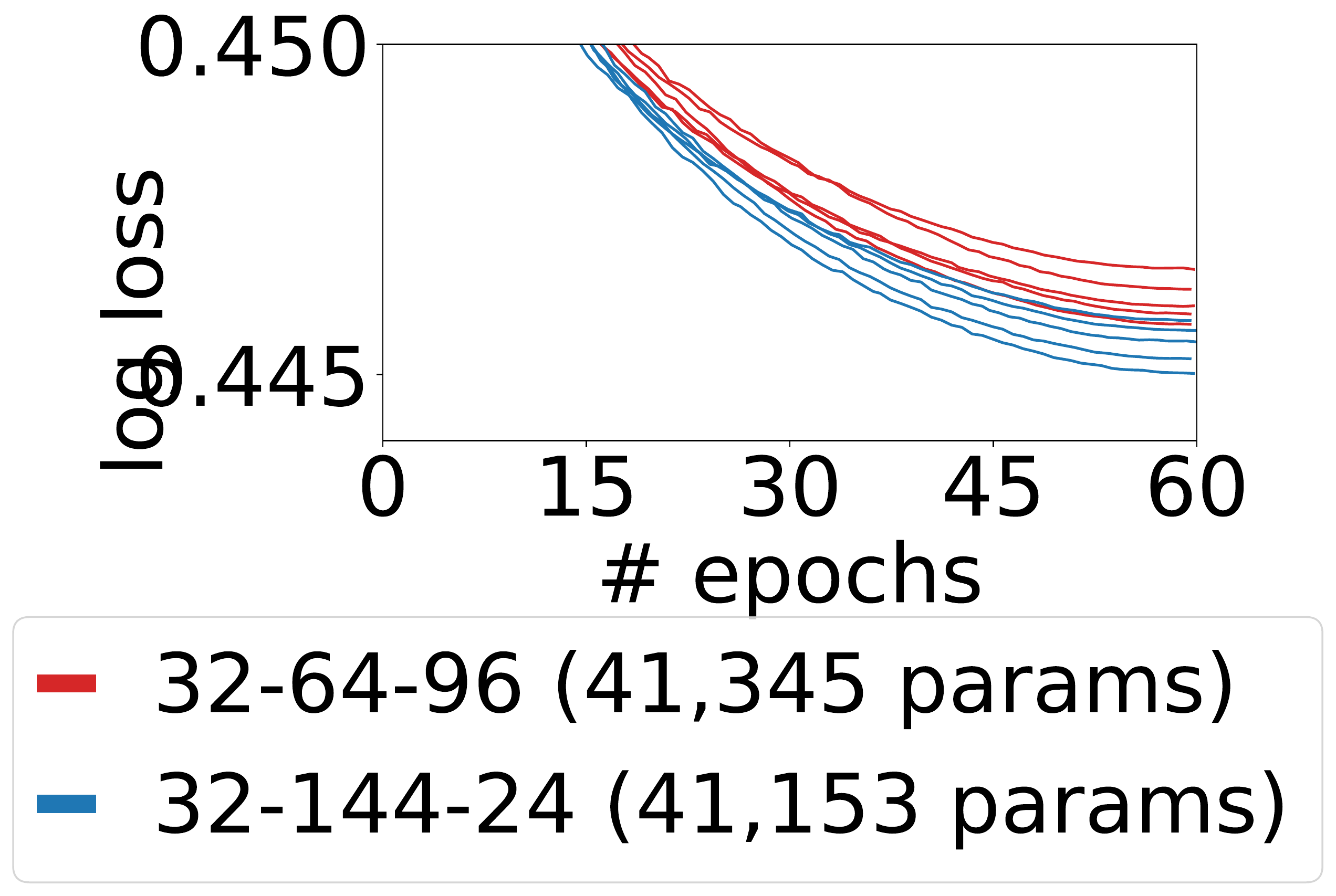}}

\caption{\textbf{Failure case of the Abs Reward on Criteo} in a search space of 3-layer FFNs. 
The change of sampling probabilities and comparison of retrain performance between the 32-144-24 reference and the 32-64-96 architecture found with the $Q(y) + \beta |T(y) / T_0 - 1|$ Abs Reward, the target for the reward was 41,153 parameters.
Repeated runs of the same search find the same architecture.
Figure~\ref{fig:retrain_3_layers_autoflow} shows the retrain validation losses of 32-64-96 (NAS-found) and 32-144-24 (reference).
}
\label{fig:sampling_prob_and_retrain_criteo_3_layers_autoflow}
\end{figure}

\myparagraph{Failure of latency rewards}
Figure~\ref{fig:sampling_prob_and_retrain_criteo_3_layers_autoflow} shows the sampling probabilities from the search when using the Abs Reward, and the retrain validation losses of the found architecture 32-64-96.
In Figures~\ref{fig:sampling_prob_heatmap_layer_1_criteo_3_layers_autoflow} --~\ref{fig:sampling_prob_heatmap_layer_3_criteo_3_layers_autoflow}, the sampling probabilities for the different choices are uniform during warmup and then converge quickly. The final selected model (32-64-96) is much worse than the reference model (32-144-24) even though the reference model is actually less expensive.
We also observed similar failures for the MnasNet rewards. With the MnasNet rewards, the RL controller also struggles to find a model within $\pm 5\%$ of the constraint despite a grid search of the RL parameters (details in Appendix~\ref{appsec:experiment_setup}).
In both cases, almost all found models are worse than the reference architecture.

\begin{wrapfigure}{r}{.44\linewidth}
\vspace{-1.5em}
\centering
\subfigure{\label{fig:calibration_criteo_3_layers}\includegraphics[width=.48\linewidth]{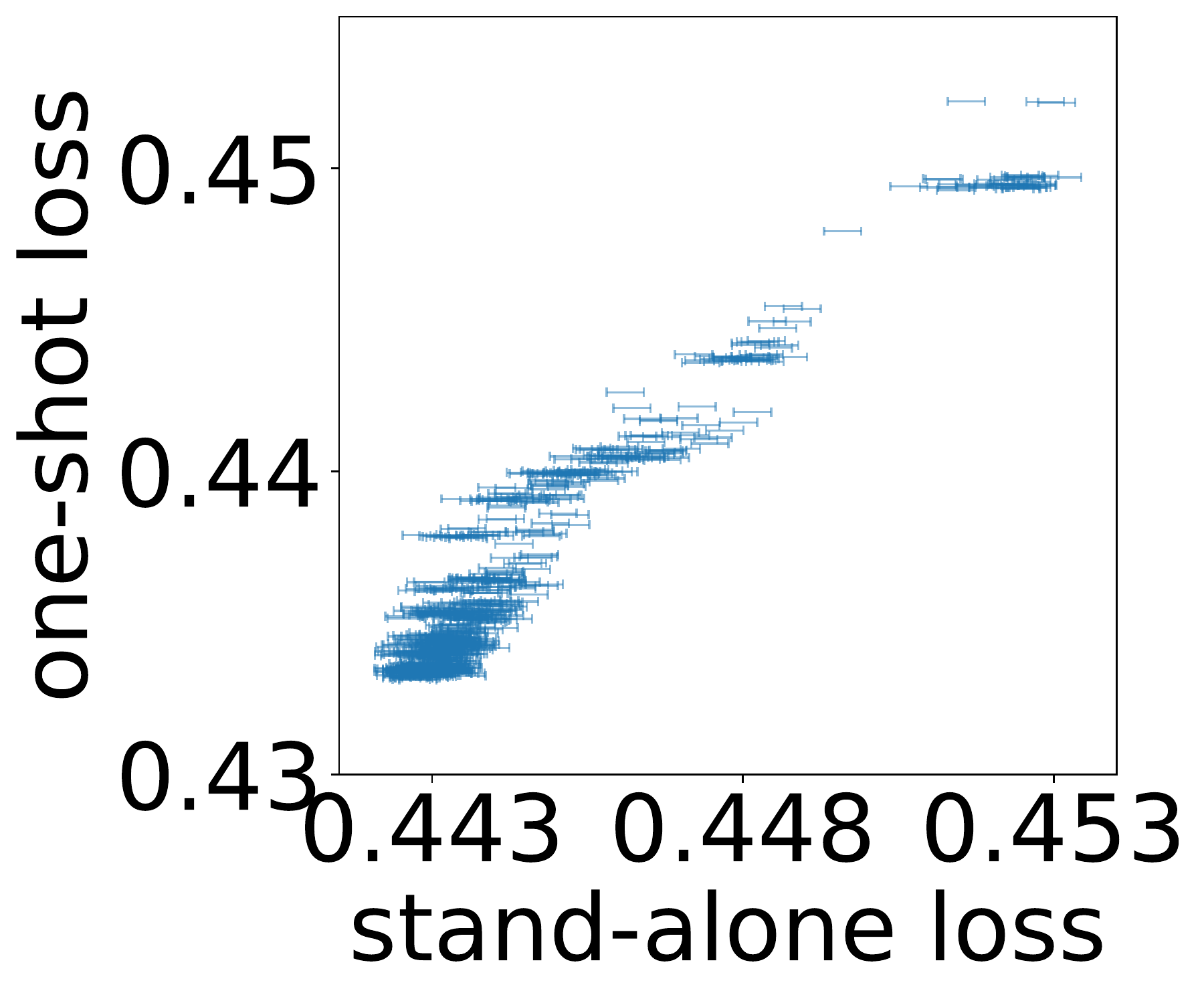}}
\hspace{.01\linewidth}
\subfigure{\label{fig:sampling_prob_heatmap_layer_1_criteo_3_layers_autoflow_further_delayed_RL}\includegraphics[width=.48\linewidth]{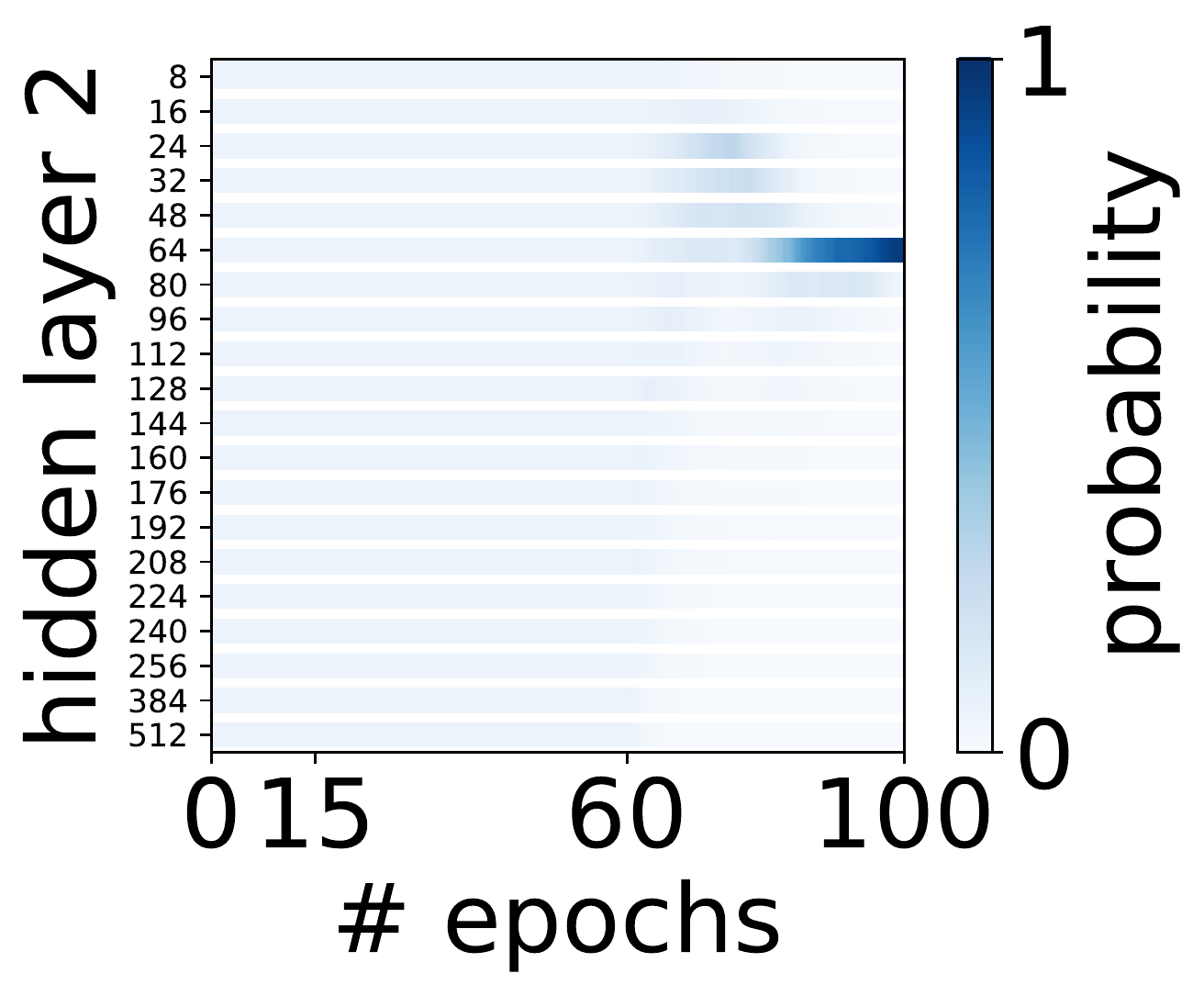}}

\caption{Left: 3-layer Criteo SuperNet calibration after 60 epochs (search space in Appendix~\ref{appsec:experiment_setup}): Pearson correlation is 0.96.
The one-shot loss is validation loss of each child network with weights taken from a SuperNet trained with the same hyperparameters as in Figure~\ref{fig:sampling_prob_and_retrain_criteo_3_layers_autoflow} but with no RL in the first 60 epochs; the stand-alone loss of each child network is computed by training the same architecture with the same hyperparameters from scratch, and has std 0.0003. 
Right: change in probabilities in layer 2 after 60 epochs of SuperNet training and 40 of RL. Note the rapid changes due to RL.
}
\label{fig:calibration_and_continuing_search}
\vspace{-2em}
\end{wrapfigure}
\myparagraph{The RL controller is to blame}
To verify that a low quality SuperNet was not the culprit, we trained a SuperNet without updating the RL controller, and manually inspected the quality of the resulting SuperNet. The sampling probabilities for the RL controller remained uniform throughout the search; the rest of the training setup was kept the same. 
At the end of the training, we compare two sets of losses on each of the child networks: the validation loss from the SuperNet (\emph{one-shot loss}), and the validation loss from training the child network from scratch.
Figure~\ref{fig:calibration_criteo_3_layers} shows that there is a strong correlation between these accuracies; Figure~\ref{fig:sampling_prob_heatmap_layer_1_criteo_3_layers_autoflow_further_delayed_RL} shows RL that starts from the sufficiently trained SuperNet weights in~\ref{fig:calibration_criteo_3_layers} still chooses the suboptimal choice 64.
This suggests that the suboptimal search results on Criteo are likely due to issues with the RL controller, rather than issues with the one-shot model weights.
In a 3 layer search space we can actually find good models without the RL controller, but in a 5 layer search space, we found an RL controller whose training is interleaved with the SuperNet is important to achieve good results.

\begin{wrapfigure}{r}{.55\linewidth}
\vspace{-4em}
\centering
\subfigure[$\beta=-10$, RL learning rate $\eta = 0.001$]{\label{fig:retrain_volkert_4_layers_autoflow_44-56-30-44}\includegraphics[width=.49\linewidth]{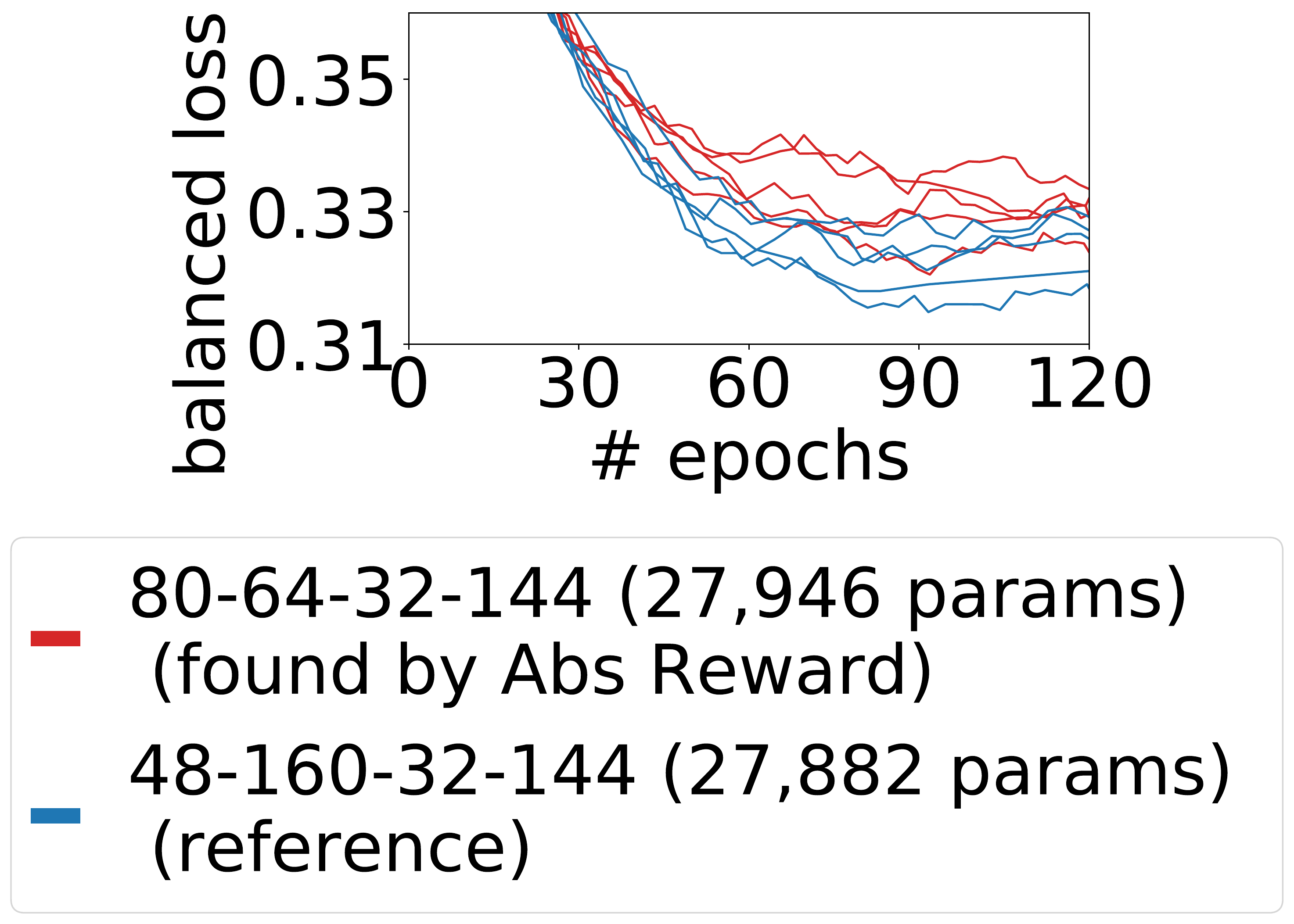}}
\subfigure[$\beta=-25$, RL learning rate $\eta = 0.001$]{\label{fig:retrain_volkert_4_layers_autoflow_32-30-64-76}\includegraphics[width=.49\linewidth]{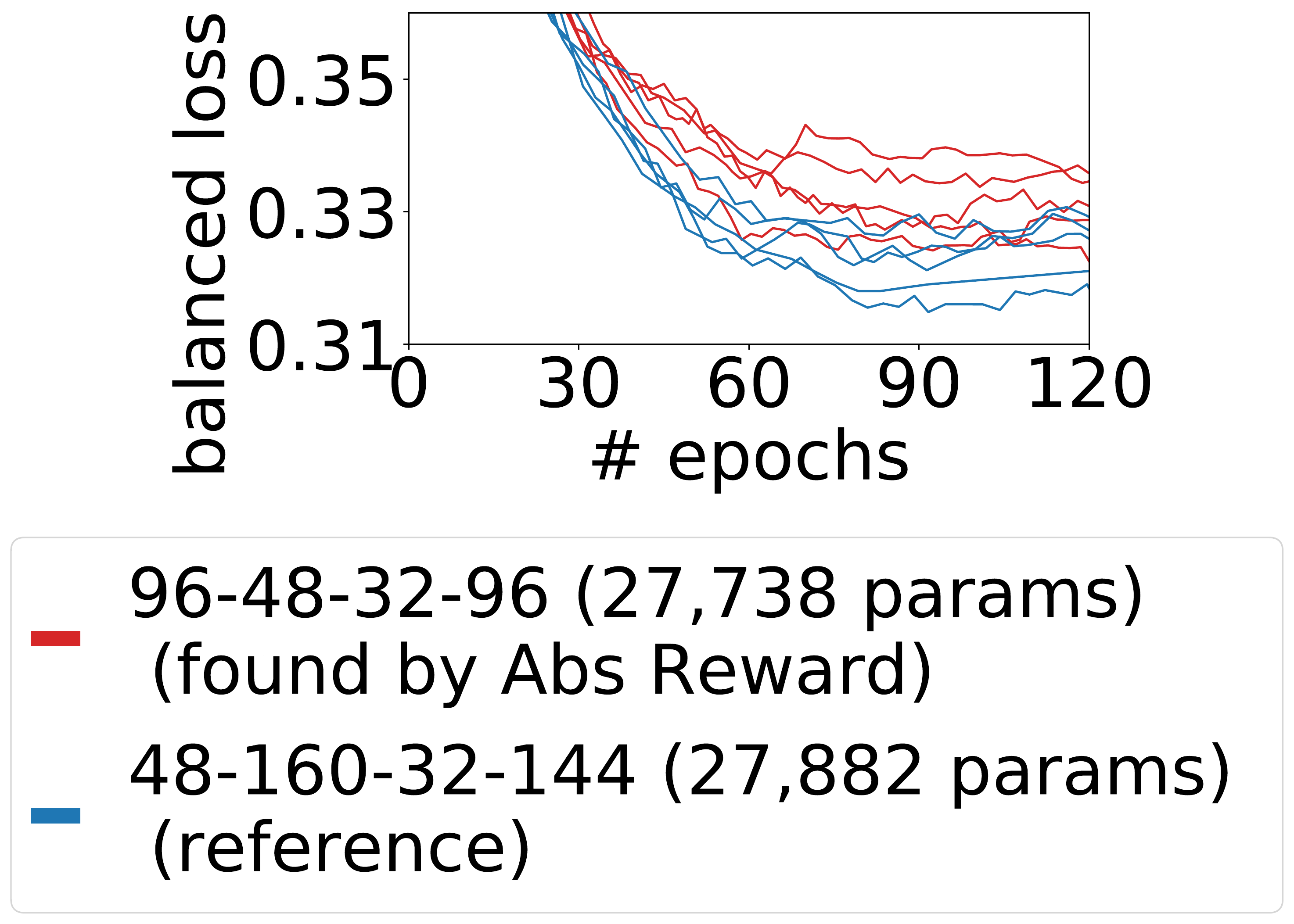}}

\caption{Abs Reward misses the global optimum on Volkert. 
Figure shows the retrain validation losses of two 
architectures found by the Abs Reward vs.~the 48-160-32-144 reference.
}
\label{fig:retrain_volkert_4_layers_autoflow}
\end{wrapfigure}
\subsubsection{Volkert -- 4 layer search space}
\label{sec:failure_volkert_4_layer}
We search for 4-layer and 9-layer networks on the Volkert dataset; details in Appendix~\ref{appsec:more_failure_cases}. 
For resource-aware RL rewards, we ran a grid search over the RL learning rate and $\beta$ hyperparameter. 
The reference architecture for the 4 layer search space is 48-160-32-144 with 27,882 parameters. 
Despite a hyperparameter grid search, it was difficult to find models with the right target cost reliably using the MnasNet rewards. 
Using the Abs Reward (Figure~\ref{fig:retrain_volkert_4_layers_autoflow}), searched models met the target cost but their quality was suboptimal, and the trend is similar to what has been shown in the toy example (Figure~\ref{fig:toy_example}): a smaller $|\beta|$ gives an infeasible architecture that is beyond the reference number of parameters, and a larger $|\beta|$ gives an architecture that is feasible but suboptimal. 

\subsubsection{A common failure pattern}
Apart from Section~\ref{sec:failure_criteo_3_layer} and~\ref{sec:failure_volkert_4_layer}, more examples in search spaces of deeper FFNs can be found in Appendix~\ref{appsec:more_failure_cases}. 
In cases on Criteo and Volkert where where the RL controller with soft constraints cannot match the quality of the reference architectures, the reference architecture often has a bottleneck structure.
For example, with a 1,027-dimensional input representation, the 32-144-24 reference on Criteo has bottleneck 32; with 180 features, the 48-160-32-144 reference on Volkert has bottleneck 48 and 32.
As the example in Section~\ref{sec:meth_mc} shows, the wide hidden layers around the bottlenecks get penalized harder in the search, and it is thus more difficult for RL with the Abs Reward to find a model that can match the reference performance.
Also, Appendix~\ref{appsec:tradeoff_plot_details} shows the Pareto-optimal architectures in the tradeoff points in Figure~\ref{fig:tradeoff_criteo_3_layers} often have bottleneck structures, so resource-aware RL rewards in previous NAS practice may have more room for improvement than previously believed.

\subsection{NAS with TabNAS reward}
\label{sec:exp_mc}

With proper hyperparameters (Appendix~\ref{appsec:hyperparameter_tuning}), our RL controller with TabNAS reward finds the global optimum when RL with resource-aware rewards produces suboptimal results.

\begin{figure}
\centering
\subfigure[Layer 1 (finally 32)]{\label{fig:sampling_prob_heatmap_layer_1_criteo_3_layers_rejection}\includegraphics[width=.22\linewidth]{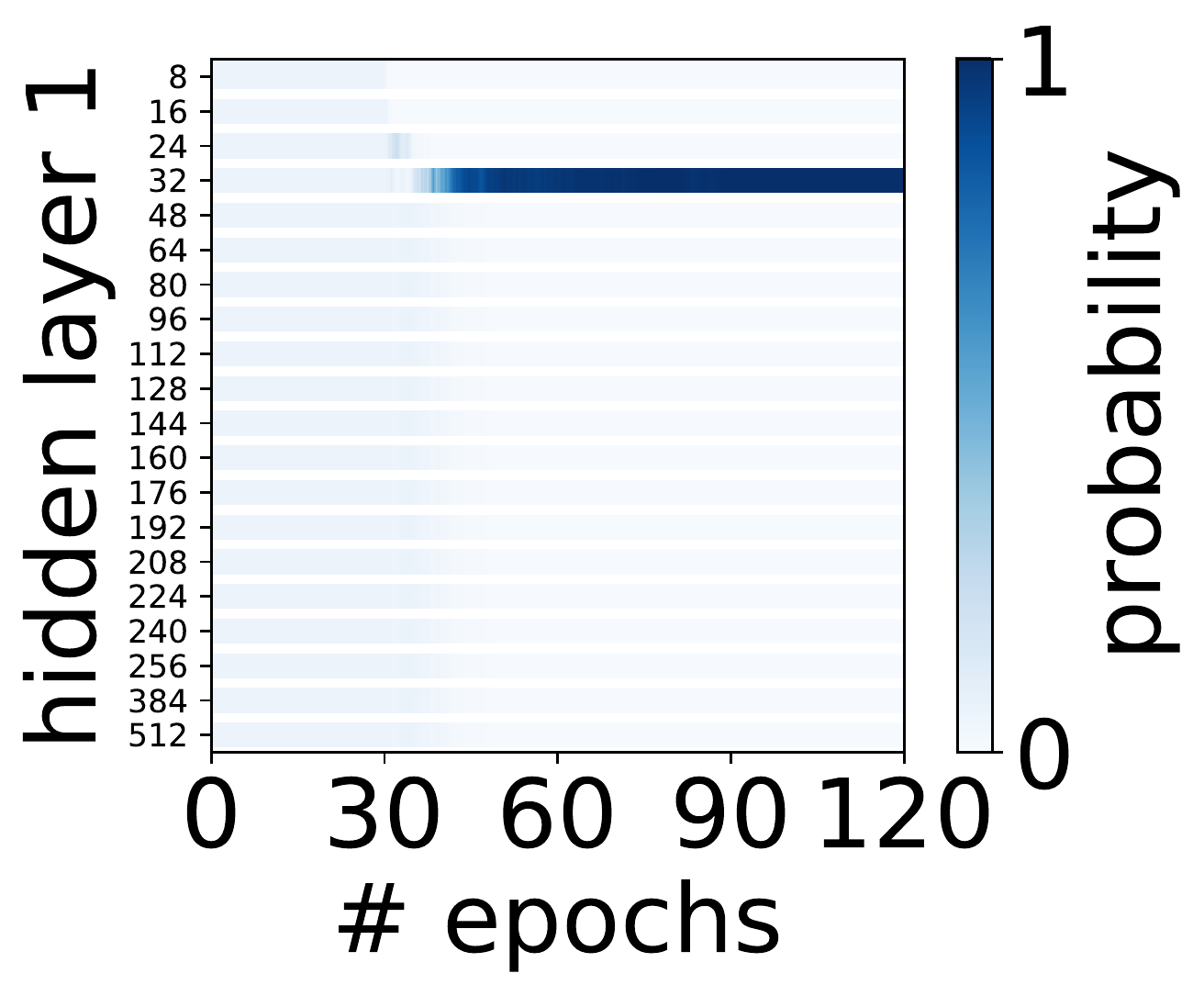}}
\hspace{.02\linewidth}
\subfigure[Layer 2 (finally 144)]{\label{fig:sampling_prob_heatmap_layer_2_criteo_3_layers_rejection}\includegraphics[width=.22\linewidth]{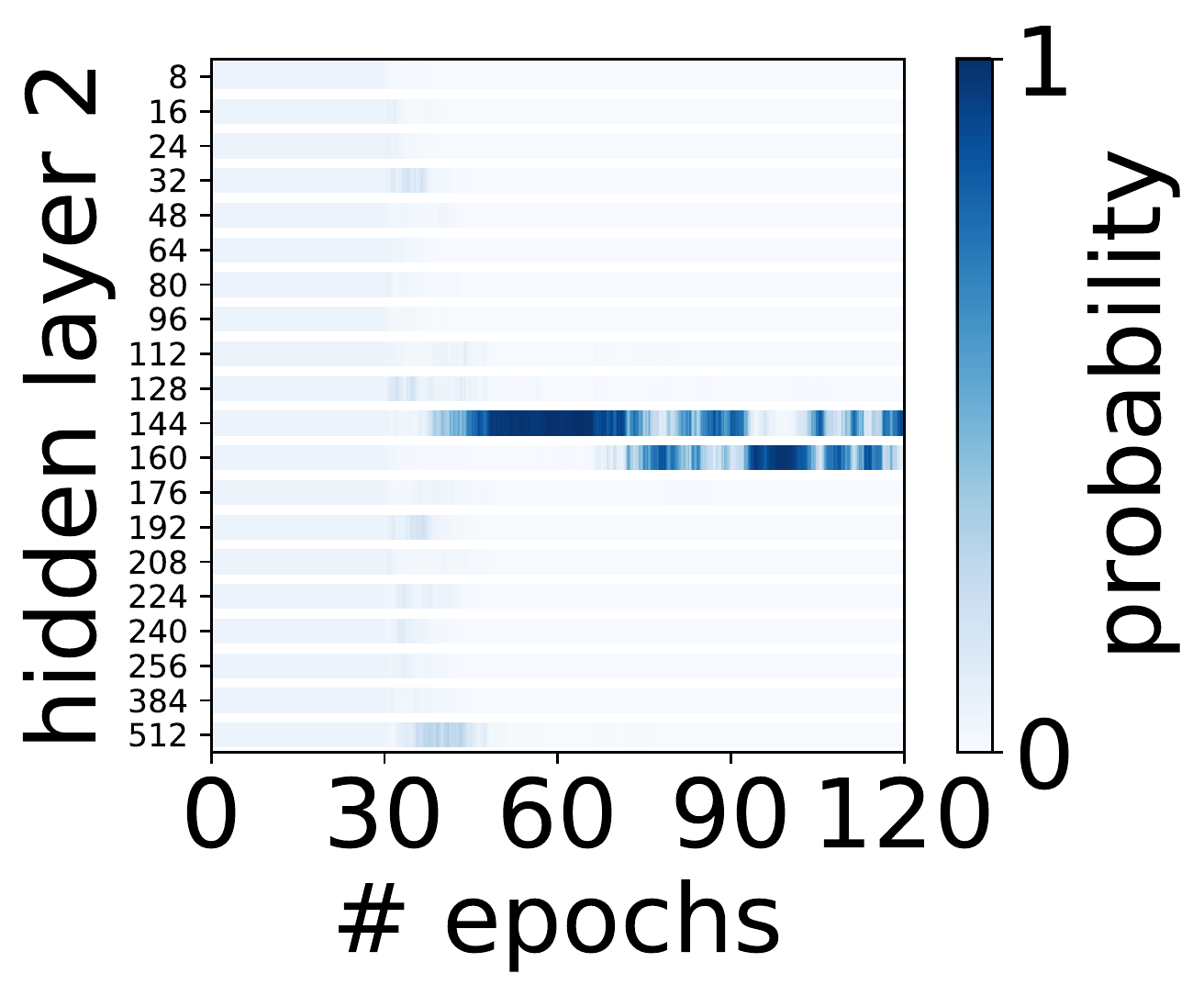}}
\hspace{.02\linewidth}
\subfigure[Layer 3 (finally 24)]{\label{fig:sampling_prob_heatmap_layer_3_criteo_3_layers_rejection}\includegraphics[width=.22\linewidth]{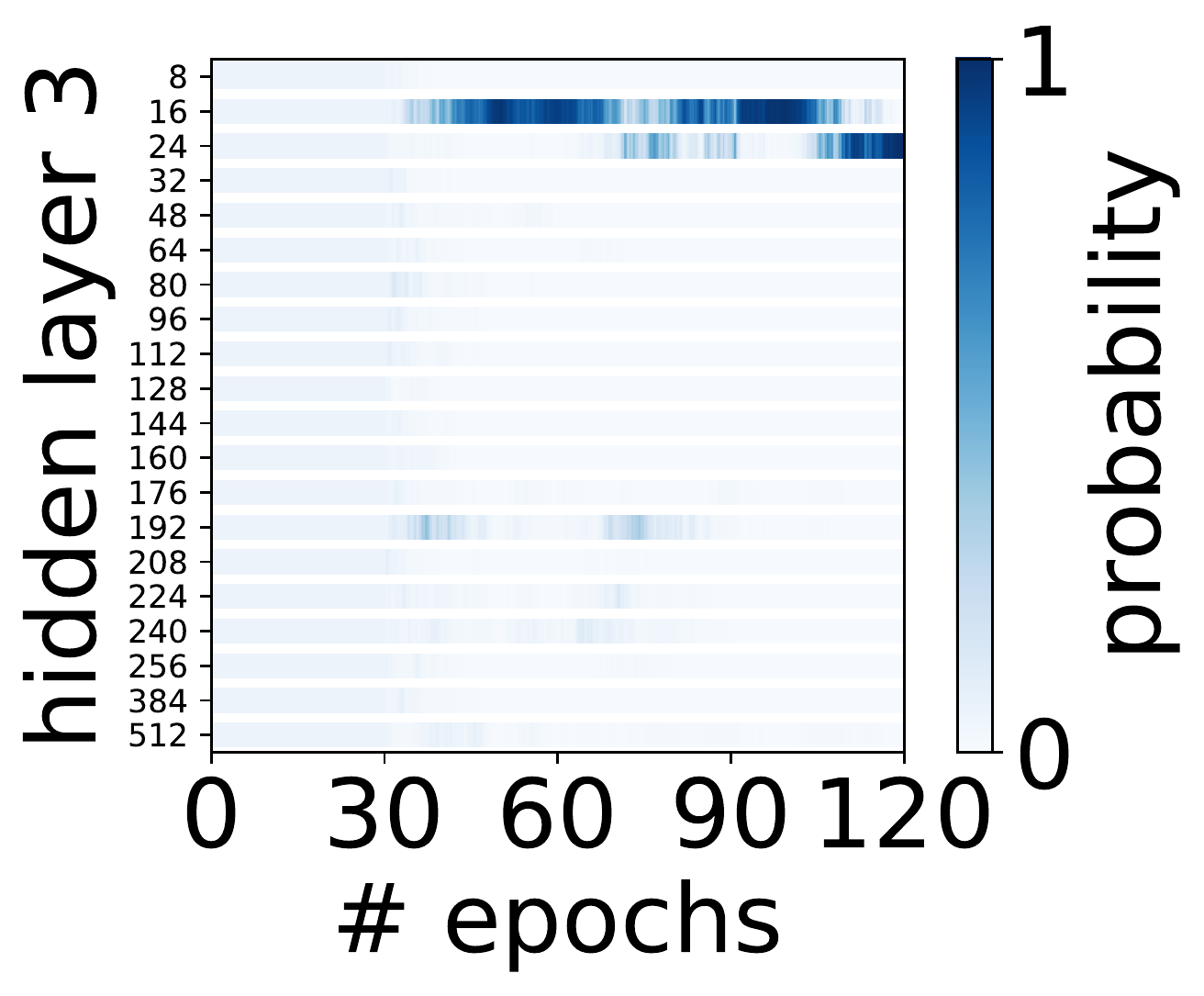}}
\hspace{.02\linewidth}
\subfigure[valid probabilities]{\label{fig:valid_prob_criteo_3_layers}\includegraphics[width=.23\linewidth]{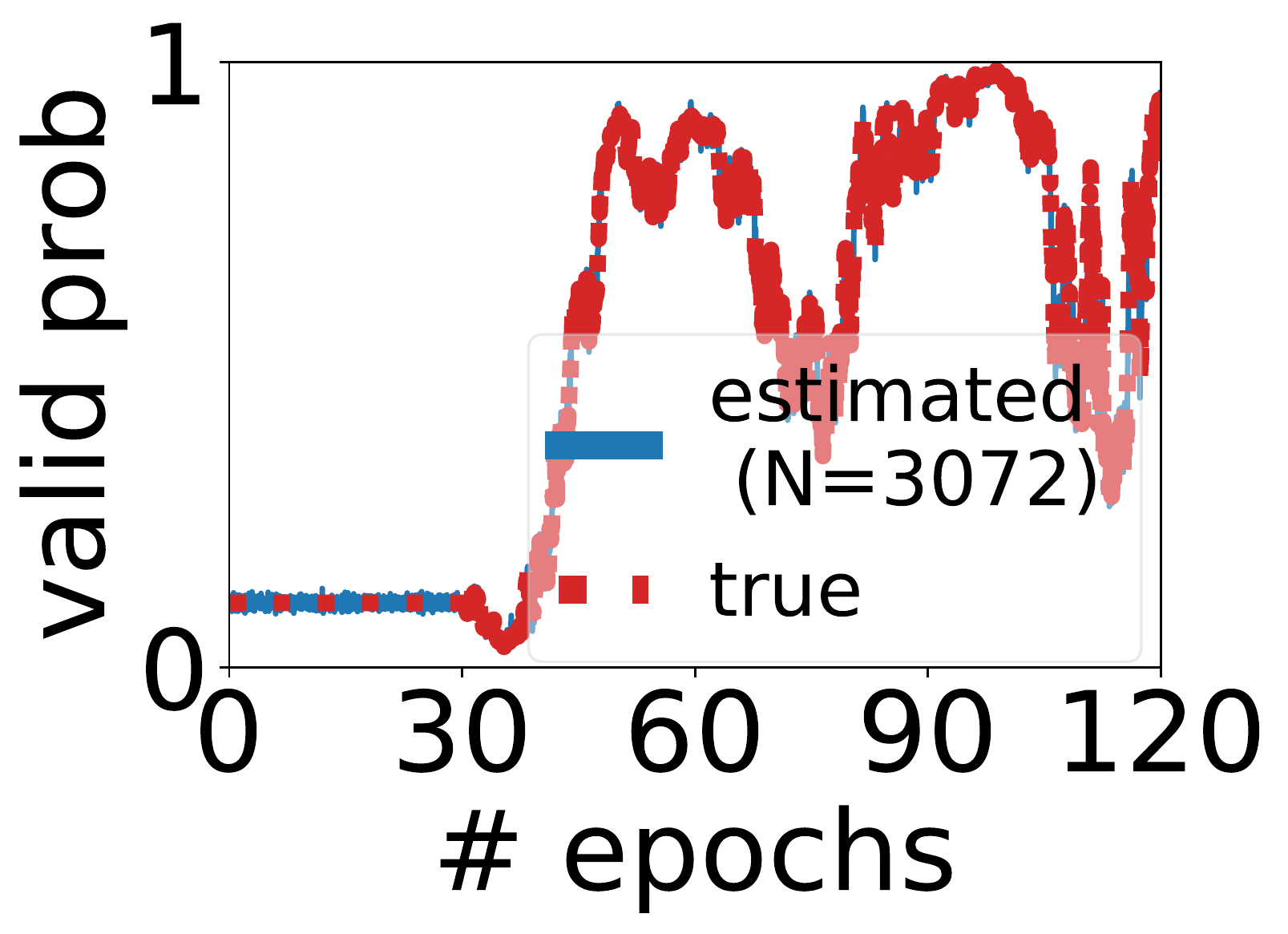}}

\caption{\textbf{Success case:} on Criteo in a search space of 3-layer FFNs, Monte-Carlo sampling with rejection eventually finds 32-144-24, the reference architecture, with RL learning rate 0.005 and number of MC samples 3,072.
Figure~\ref{fig:valid_prob_criteo_3_layers} shows the change of true and estimated valid probabilities.
}
\label{fig:sampling_prob_and_valid_prob_criteo_3_layers_rejection}
\end{figure}

TabNAS does not introduce a resource-aware bias in the RL reward (Section~\ref{sec:meth_mc}). 
Instead, it uses conditional probabilities to update the logits in feasible architectures.
We run TabNAS for 120 epochs with RL learning rate 0.005 and $N=3072$ MC samples.\footnote{The 3-layer search space has $20^3=8000$ candidate architectures, which is small enough to compute $\probP(V)$ exactly. However, MC can scale to larger spaces which are prohibitively expensive for exhaustive search (Appendix~\ref{appsec:more_failure_cases}).}
The RL controller converges to two architectures, 32-160-16 (40,769 parameters, with loss 0.4457 $\pm$ 0.0002) and 32-144-24 (41,153 parameters, with loss 0.4455 $\pm$ 0.0003), after around 50 epochs of NAS, then oscillates between these two solutions (Figure \ref{fig:sampling_prob_and_valid_prob_criteo_3_layers_rejection}). After 120-epochs, we sample from the layerwise distribution and pick the largest feasible architecture: the global optimum 32-144-24.

On the same hardware, the search takes $3\times$ the runtime of stand-alone training.
Hence, as can be seen in Figure \ref{fig:comparison_with_random_criteo_3_layers}, the proposed architecture search method is much faster than a random baseline.

\subsection{TabNAS automatically determines whether bottlenecks are needed}
\label{sec:tabnas_automatically_decides_bottleneck}
Previous NAS works like MnasNet and TuNAS (often or only on vision tasks) often have inverted bottleneck blocks~\cite{sandler2018mobilenetv2} in their search spaces.
However, the search spaces used there have a hard-coded requirement that certain layers must have bottlenecks. 
In contrast, our search spaces permit the controller to \emph{automatically determine} whether to use bottleneck structures based on the task under consideration.
TabNAS automatically finds high-quality architectures, both in cases where bottlenecks are needed and in cases where they are not. 
This is important because networks with bottlenecks do not always outperform others on all tasks. 
For example, the reference architecture 32-144-24 outperforms the TuNAS-found 32-64-96 on Criteo, but the reference 64-192-48-32 (64,568 parameters, 0.0662 $\pm$ 0.0011) is on par with the TuNAS-and-TabNAS-found 96-80-96-32 (64,024 parameters, 0.0669 $\pm$ 0.0013) on Aloi.
TabNAS automatically finds an optimal (bottleneck) architecture for Criteo, and automatically finds an optimal architecture that does not necessarily have a bottleneck structure for Aloi.
Previous reward-shaping rewards like the Abs Reward only succeed in the latter case.

\subsection{Rejection-based reward outperforms Abs Reward in NATS-Bench size search space}
Although we target resource-constrained NAS on tabular datasets in this paper, our proposed method is not specific to NAS on tabular datasets.
In Appendix~\ref{appsec:comparison_in_natsbench_size_search_space}, we show the rejection-based reward in TabNAS outperforms RL with the Abs Reward in the size search space of NATS-Bench~\cite{dong2021nats}, a NAS benchmark on vision tasks.

\section{Conclusion}
We investigate the failure of resource-aware RL rewards to discover optimal structures in tabular NAS 
and propose TabNAS for tabular NAS in a constrained search space.
The TabNAS controller uses a rejection mechanism to compute the policy gradient updates
from feasible architectures only,
and uses Monte-Carlo sampling to reduce the cost of debiasing this rejection-sampling approach.
Experiments show TabNAS finds better architectures than previously proposed RL methods with resource-aware rewards in resource-constrained searches.

Many questions remain open. For example:
1) Can the TabNAS strategy find better architectures on other types of tasks such as vision and language?
2) Can TabNAS improve RL results for more complex architectures?
3) Is TabNAS useful for resource-constrained RL problems more broadly?

\begin{ack}
This work was done when Madeleine Udell was a visiting researcher at Google. 
The authors thank Ruoxi Wang, Mike Van Ness, Ziteng Sun, Xuanyi Dong, Lijun Ding, Yanqi Zhou, Chen Liang, Zachary Frangella, Yi Su, and Ed H.~Chi for helpful discussions, and thank several anonymous reviewers for useful comments. 
\end{ack}

\bibliographystyle{plainnat}
\bibliography{scholar}

\appendix
\section{Previous work}
\label{appsec:literature}
\subsection{Neural architecture search (NAS)}
Neural architecture search (NAS)~\cite{zoph2016neural} stems from the resurgence of deep learning.
It focuses on tuning architectural hyperparameters in neural networks, like hidden layer sizes in feedforward networks or convolutional kernel sizes in convolutional networks. 
The earliest NAS papers \cite{zoph2016neural,zoph2018learning} trained thousands of network architectures from scratch for a single search. 
Due to the high costs involved, many works have proposed different methods to reduce the search cost. 
Most of these proposals are based around two complementary (but often intertwined) high-level strategies.

The first strategy is to reduce the time needed to evaluate each architecture seen during a search. For example, instead of training each network architecture from scratch, we can train a SuperNet -- a single set of shared model weights that can be used to evaluate and rank any different candidate architecture in the search space~\cite{bender2018understanding, liu2018darts,shi2020bridging,zhao2021few}.
Other approaches include the use of network morphism~\cite{kandasamy2018neural,elsken2018efficient} to initialize weights of candidate networks that are close to previous instances.

The second strategy is to reduce the number of architectures we need to evaluate during a search.
Proposed methods include reinforcement learning algorithms~\cite{zoph2016neural, cai2018proxylessnas, bender2020can} that learn a probability distribution over candidate architectures, evolutionary search~\cite{liu2018progressive,elsken2018efficient,guo2020single,awad2020differential,real2019regularized} that pursues more promising architectures from ancestors, and Bayesian optimization~\cite{kandasamy2018neural,jin2019auto,white2019bananas,zhou2019bayesnas,shi2020bridging} and parametric models~\cite{zhang2018graph, wen2020neural} that directly predict network performance. 

Resource constraints are prevalent in deep learning.
Finding architectures with outstanding performance and low costs are important to both NAS research and application.
Apart from the surrogate models above that transfer knowledge across network candidates to avoid exhaustive search, specific techniques have been adopted to find networks with a good balance of performance and resource consumption.
A popular method is to add regularizers~\cite{wu2019fbnet,tan2019mnasnet,bender2020can,lyu2021resource} that penalize expensive architectures.
With hard resource constraints, greedy submodular maximization~\cite{xiong2019resource} and heuristic scaling methods~\cite{hu2020tf} were used to grow or shrink networks during the search, so as to ensure the chosen candidate architecture obeys the constraint.
TabNAS in this work operates under the hard resource constraint, and can find the global optima in the feasible set of architectures.

\subsection{Deep learning on tabular datasets}
Deep neural networks are gaining popularity on tabular datasets in academia and industry.
In the pursuit of designing better architectures for tabular deep learning,
an earlier line of work~\cite{ke2018tabnn,arik2019tabnet,popov2019neural,abutbul2020dnf,chang2021node} mimic the structure of tree-based models~\cite{chen2016xgboost,ke2017lightgbm}.
Some other works use the attention mechanism~\cite{huang2020tabtransformer,somepalli2021saint,gorishniy2021revisiting} or are based on feedforward~\cite{kadra2021well,gorishniy2021revisiting} or residual networks~\cite{gorishniy2021revisiting}.

While automated machine learning (AutoML) on tabular datasets has been addressed from multiple perspectives and with different approaches~\cite{feurer2020auto,olson2016tpot,fusi2018probabilistic,yang2019oboe,erickson2020autogluon}, NAS on tabular datasets is less explored, partly due to insufficient understanding of promising architectures and a lack of benchmarks~\footnote{To disambiguate, the phrase ``tabular NAS benchmark'' in previous NAS benchmark literature~\cite{ying2019bench,mehta2022bench} often refers to tabulated performance of architectures on vision and language tasks.}.
\citet{egele2021agebo} interleaves NAS with aging evolution~\cite{real2019regularized} in a search space with multiple branches and hyperparameter tuning with Bayesian optimization.
We show TabNAS can find architectures as simple as FFNs with a few layers that have outstanding performance and obey the resource constraint.

\section{Algorithm pseudocode}
\label{appsec:pseudocode}

We show pseudocode of the algorithms introduced in Section~\ref{sec:meth}.

\begin{algorithm}[H]
\caption{(Resource-Oblivious) One-Shot Training and REINFORCE}
\label{alg:reinforce}
\begin{algorithmic}[1]
\Require{search space $S$, weight learning rate $\alpha$, RL learning rate $\eta$}
\Ensure{sampling probabilities $\{p_{ij}\}_{i \in [L], j \in [C_i]}$}
\State initialize logits $\ell_{ij} \gets 0$, $\forall i \in [L], j \in [C_i]$
\State initialize quality reward moving average $\bar{Q} \gets 0$
\State layer warmup
\For {$\text{iter}=1$ {\bfseries to} max\_iter}
\State $p_{ij} \gets \exp (\ell_{ij}) / \sum_{j \in [C_i]} \exp (\ell_{ij})$, $\forall i \in [L], j \in [C_i]$
\State \Comment{weight update}
\For{$i=1$ to $L$}
\State $x_i \gets$ the $i$-th layer size sampled from $\{s_{ij}\}_{j \in [C_i]}$ with distribution $\{p_{ij}\}_{j \in [C_i]}$
\EndFor
\State $\textit{loss}(x) \gets$ the (training) loss of $x = x_1 \mhyphen \dots \mhyphen x_L$ on the training set
\State $w \gets w - \alpha \nabla \textit{loss}(x)$, in which $w$ is the weights of $x$ \Comment{can be replaced with optimizers other than SGD}
\State \Comment{RL update}
\For{$i=1$ to $L$}
\State $y_i \gets$ the $i$-th layer size sampled from $\{s_{ij}\}_{j \in [C_i]}$ with distribution $\{p_{ij}\}_{j \in [C_i]}$
\EndFor
\State $Q(y) \gets 1 - \textit{loss}(y)$, the quality reward of $y = y_1 \mhyphen \dots \mhyphen y_L$ on the validation set
\State RL reward $r(y) \gets Q(y)$\hspace{1em}\Comment{can be replaced with resource-aware rewards introduced in Section~\ref{sec:meth_mc}}
\State $J(y) \gets \stopgrad{r(y) - \bar{Q}} \log \probP(y)$ \hspace{1em}\Comment{can be replaced with Algorithm~\ref{alg:mc} when resource-constrained}
\State $\ell_{ij} \gets \ell_{ij} + \eta \nabla J(y)$, $\forall i \in [L], j \in [C_i]$ \Comment{can be replaced with optimizers other than SGD}
\State $\bar{Q} \gets \frac{\gamma*\bar{Q} + (1-\gamma) * Q(y)}{\gamma*\bar{Q} + 1-\gamma}$ \Comment{update moving average with $\gamma=0.9$}
\EndFor
\end{algorithmic}
\end{algorithm}

\begin{algorithm}[H]
\caption{Rejection with Monte-Carlo (MC) Sampling}
\label{alg:mc}
\begin{algorithmic}[1]
\State {\bfseries Input:} number of MC samples $N$, feasible set $V$, MC proposal distribution $q$, quality reward moving average $\bar{Q}$, sampled architecture for RL in the current step $y = y_1 \mhyphen y_2 \mhyphen \dots \mhyphen y_L$, current layer size distribution over $\{s_{ij}\}_{j \in [C_i]}$ with probability $\{p_{ij}\}_{j \in [C_i]}$
\State {\bfseries Output:} $J(y)$
\If{$y$ is feasible}
\State $Q(y) =$ the quality reward of $y$
\State $\probP(y) := \prod\limits_{i \in [L]} \probP(Y_i = y_i)$
\For{$i=1$ to $L$}
\State $\{z_i^{(k)}\}_{k \in [N]} \gets$ $N$ samples of the $i$-th layer size, sampled from $\{s_{ij}\}_{j \in [C_i]}$ with distribution $\{p_{ij}\}_{j \in [C_i]}$
\EndFor
\State $p_i^{(k)} := \probP(Z_i = z_i^{(k)})$, $\forall i \in [L], k \in [N]$
\State $p^{(k)} := \prod\limits_{i \in [L]} p_i^{(k)}$, $\forall k \in [N]$
\State $\widehat{\probP}(V) \gets \frac{1}{N} \sum\limits_{k \in [N], z^{(k)} \in V} \frac{p^{(k)}}{q^{(k)}}$, in which $z^{(k)} := z_1^{(k)} \mhyphen \dots \mhyphen z_L^{(k)}$
\State $J(y) \gets \stopgrad{Q(y) - \bar{Q}} \log\frac{\probP(y)}{\widehat{\probP}(V)}$
\Else
\State $J(y) \gets 0$
\EndIf
\end{algorithmic}
\end{algorithm}

\begin{figure}[H]
\centering
\subfigure{\includegraphics[width=.35\linewidth]{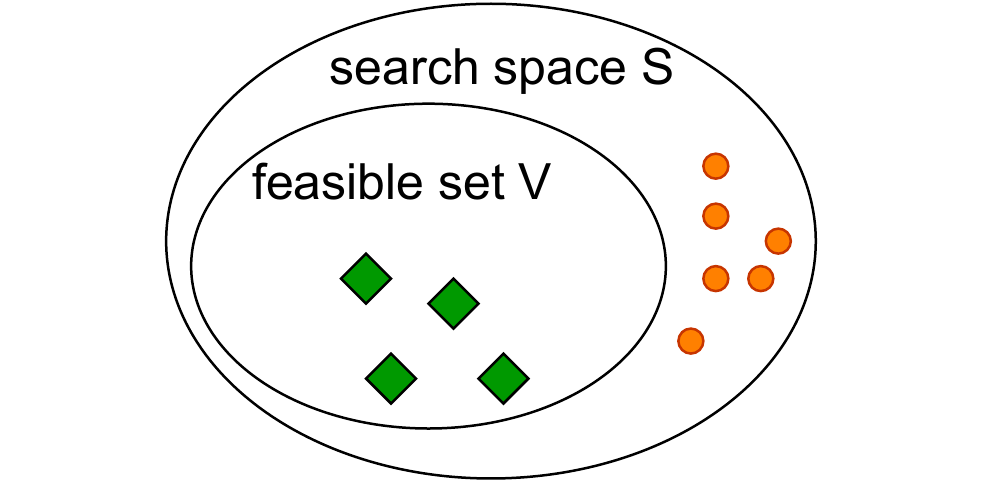}}
\vspace{-1em}
\caption{Illustration of the feasible set $V$ within the search space $S$.
Each green diamond or orange dot denotes a feasible or infeasible architecture, respectively.
}
\label{fig:feasible_set_illustration}
\end{figure}

\begin{algorithm}[H]
\caption{Sample to Return the Final Architecture}
\label{alg:sample_to_get_solution}
\begin{algorithmic}[1]
\State {\bfseries Input:} sampling probabilities $\{p_{ij}\}_{i \in [L], j \in [C_i]}$ returned by Algorithm~\ref{alg:reinforce}, number of desired architectures $n$, number of samples to draw $m$
\State {\bfseries Output:} the set of $n$ selected architectures $A$
\For{$i=1$ to $L$}
\State $\{x_i^{(k)}\}_{k \in [m]} \gets$ $m$ samples of the $i$-th layer size, sampled from $\{s_{ij}\}_{j \in [C_i]}$ with distribution $\{p_{ij}\}_{j \in [C_i]}$
\EndFor
\State $F := \{k \in [m] \mid x_1^{(k)} \mhyphen x_2^{(k)} \mhyphen \dots \mhyphen x_L^{(k)} \in V\}$
\State $A \gets$ $n$ unique architectures in $F$ with largest numbers of parameters
\end{algorithmic}
\end{algorithm}

Notice that in Algorithm~\ref{alg:reinforce}, we show the weight and RL updates with the stochastic gradient descent (SGD) algorithm; in our experiments on the toy example and real datasets, we use Adam for both updates as in ProxylessNAS~\cite{cai2018proxylessnas} and TuNAS~\cite{bender2020can}, since it synchronizes convergence across different layer size choices, and slows down the learning which would otherwise converge too rapidly.

\section{Details of experiment setup}
\label{appsec:experiment_setup}
\subsection{Toy example}
\label{appsec:more_details_toy_example}

We use the Adam optimizer with $\beta_1=0.9$, $\beta_2=0.999$ and $\epsilon=0.001$ to update the logits.
When we use the Abs Reward, the results are similar when $\eta \geq 0.05$, while the RL controller with $\eta < 0.05$ converges too slow or is hard to converge.
When we use the rejection-based reward, we use RL learning rate $\eta=0.1$; other $\eta$ values with which RL converges give similar results.

\subsection{Real datasets}
\label{appsec:more_details_real_datasets}
Table~\ref{table:dataset_details} shows the datasets we use.
Datasets other than Criteo~\footnote{https://ailab.criteo.com/download-criteo-1tb-click-logs-dataset/} come from the OpenML dataset repository~\cite{OpenML2013}.
For Criteo, we randomly split the labeled part (45,840,617 points) into 90\% training (41,258,185 points) and 10\% validation (4,582,432 points); for the other datasets, we randomly split into 80\% training and 20\% validation\footnote{The ranking of validation losses among architectures under such splits is almost the same as that of test losses under 60\%-20\%-20\% training-validation-test splits.}.
The representations we use for Criteo are inspired by DCN-V2~\cite{wang2021dcn}.

\begin{table}[H]
\footnotesize
\caption{Dataset details}
\label{table:dataset_details}
\begin{center}
\begin{small}
\begin{tabular}{cccccP{5cm}}
\toprule
 name & \# points & \multicolumn{2}{c}{\# features} & \# classes & embedding we use for each feature\\
 \cmidrule(lr){3-4}
 & & numerical & categorical \\
 \midrule
Criteo & 51,882,752 & 13 & 26 & 2 & original values for each numerical, 39-dimensional for each categorical\\
Volkert & 58,310 & 180 & 0 & 10 & original values\\
Aloi & 108,000 & 128 & 0 & 1,000 & original values \\
Connect-4 & 67,557 & 0 & 42 & 3 & 2-dimensional for each categorical \\
Higgs & 98,050 & 28 & 0 & 2 & original values \\
\bottomrule
\end{tabular}
\end{small}
\end{center}
\end{table}

Table~\ref{table:hyperparameter_details} shows the hyperparameters we use for stand-alone training and NAS, found by grid search.
With these hyperparameters, the best architecture in each of our search spaces (introduced in Appendix~\ref{appsec:tradeoff_plot_details}) has performance that is within $\pm 5\%$ of the best performance in~\citet{kadra2021well} Table 2, and we achieve these scores with FFNs that only have 5\% parameters of the ones there. 
The Adam optimizer has hyperparameters $\beta_1=0.9$, $\beta_2=0.999$ and $\epsilon=0.001$.
We use layer normalization \cite{ba2016layer} for all datasets.
We use balanced error (weighted average of classification errors across classes) for all other datasets\footnote{The performance ranking of architectures under the balanced error metric is almost the same as under logistic loss. Also, the balanced error metric is only for reporting the final validation losses; both weight and RL updates use logistic loss.} as in~\citet{kadra2021well}, except for Criteo, on which we use logistic loss as in~\citet{wang2021dcn}.

\begin{table}[H]
\footnotesize
\caption{Weight training hyperparameter details}
\label{table:hyperparameter_details}
\begin{center}
\begin{small}
\begin{tabular}{P{1.4cm}P{0.5cm}P{1.2cm}P{1.8cm}P{2.2cm}P{2cm}c}
\toprule
name & batch size & learning rate & learning rate schedule & optimizer & \# training epochs & metric \\
\midrule
Criteo & 512 & 0.002 & cosine decay & Adam & 60 & log loss\\
Volkert & 32 & 0.01 & constant & SGD with momentum 0.9 & 120 & balanced error \\
Aloi & 128 & 0.0005 & constant & Adam & 50 & balanced error\\
Connect-4 & 32 & 0.0005 & cosine decay & Adam & 60 & balanced error\\
Higgs & 64 & 0.05 & constant & SGD & 60 & balanced error\\
\bottomrule
\end{tabular}
\end{small}
\end{center}
\end{table}

We use constant RL learning rates for NAS. The Connect-4\footnote{https://www.openml.org/d/40668} and Higgs\footnote{https://www.openml.org/d/23512} datasets are easy for both the Abs Reward and rejection-based reward, in the sense that small FFNs with fewer than 5,000 parameters can achieve near-SOTA results ($\pm 5\%$ of the best accuracy scores listed in~\citet{kadra2021well} Table 2, except that we do 80\%-20\% training-validation splits and use original instead of standardized features), and RL-based weight-sharing NAS with either reward can find architectures that match the Pareto-optimal reference architectures.
The Aloi dataset\footnote{https://www.openml.org/d/42396} needs more parameters (more than 100k), but the other observations are similar to on Connect-4 and Higgs.
Thus we omit the corresponding results. 

The factorized search spaces we use for NAS are:
\begin{itemize}[leftmargin=2em,topsep=0pt,partopsep=1ex,parsep=0ex]
    \item Criteo: Each layer has 20 choices \{8, 16, 24, 32, 48, 64, 80, 96, 112, 128, 144, 160, 176, 192, 208, 224, 240, 256, 384, 512\}.
    \item Volkert\footnote{https://www.openml.org/d/41166}, 4-layer networks: Each layer has 20 choices \{8, 16, 24, 32, 48, 64, 80, 96, 112, 128, 144, 160, 176, 192, 208, 224, 240, 256, 384, 512\}.
    \item Volkert, 9-layer networks: Each layer has 12 choices \{8, 16, 24, 32, 48, 64, 80, 96, 112, 128, 144, 160\}.
    This search space has fewer choices for each hidden layer than the 4-layer counterpart, but the size of the search space is over $3 \times 10^4$ times larger. 
\end{itemize}

Our goal is not to achieve state-of-the-art (SOTA) accuracy, 
but to find the best architecture that obeys a resource upper bound. 
This mimics a resource-constrained setting that is common in practice.
Impressively, our method does nearly match SOTA performance, as 
our search space has architectures that are close to the best in previous literature. 
For example:
\begin{itemize}[leftmargin=2em,topsep=0pt,partopsep=1ex,parsep=0ex]
\item On Criteo: The best architecture in our search space achieves public and private scores 0.45284 and 0.45283 on Kaggle, ranking 20/717 on the leaderboard\footnote{https://www.kaggle.com/competitions/criteo-display-ad-challenge/leaderboard}. 
\item On Volkert: The best architecture in our search space has balanced accuracy 0.695, within 2\% of the best in~\citet{kadra2021well} and better than most other works used for comparison in that work. 
The differences in settings are that we use original features instead of standardized, and we achieve this score with an FFN that only has 5\% parameters of the one there.
\item On Aloi: The best architecture in our search space has balanced accuracy 0.957, within 2\% of the best in~\citet{kadra2021well} and better than most other works used for comparison in that work. 
Again, we use original features instead of standardized, and we achieve this score with an FFN that only has 5\% parameters of the one used there. 
And the 0.957 balanced accuracy score is also within 1\% of the best in~\citet{gorishniy2021revisiting}. 
The difference is that we use an FFN that only has $<$10\% as many parameters as the one used there. 
\end{itemize}

\subsubsection{More details on the tradeoff plot (Figure~\ref{fig:tradeoff_criteo_3_layers})}
\label{appsec:tradeoff_plot_details}
Each search space we use for exhaustive search and NAS has a fixed number of hidden layers. 
Resource-constrained NAS in a search space with varying number of hidden layers is an interesting problem for future studies.
On each dataset, we randomly sample, train and evaluate architectures in the search space with the number of parameters fall within a range, in which there is a clear tradeoff between loss and number of parameters.
These ranges are: 
\begin{itemize}
    \item Criteo: 0 -- 200,000
    \item Volkert, 4-layer networks: 15,000 -- 50,000
    \item Volkert, 9-layer networks: 40,000 -- 100,000
\end{itemize}

\begin{figure}[t]
\centering
\subfigure[Tradeoff on Criteo, in 4 layer search space]{\label{fig:tradeoff_criteo_4_layers}\includegraphics[width=.45\linewidth]{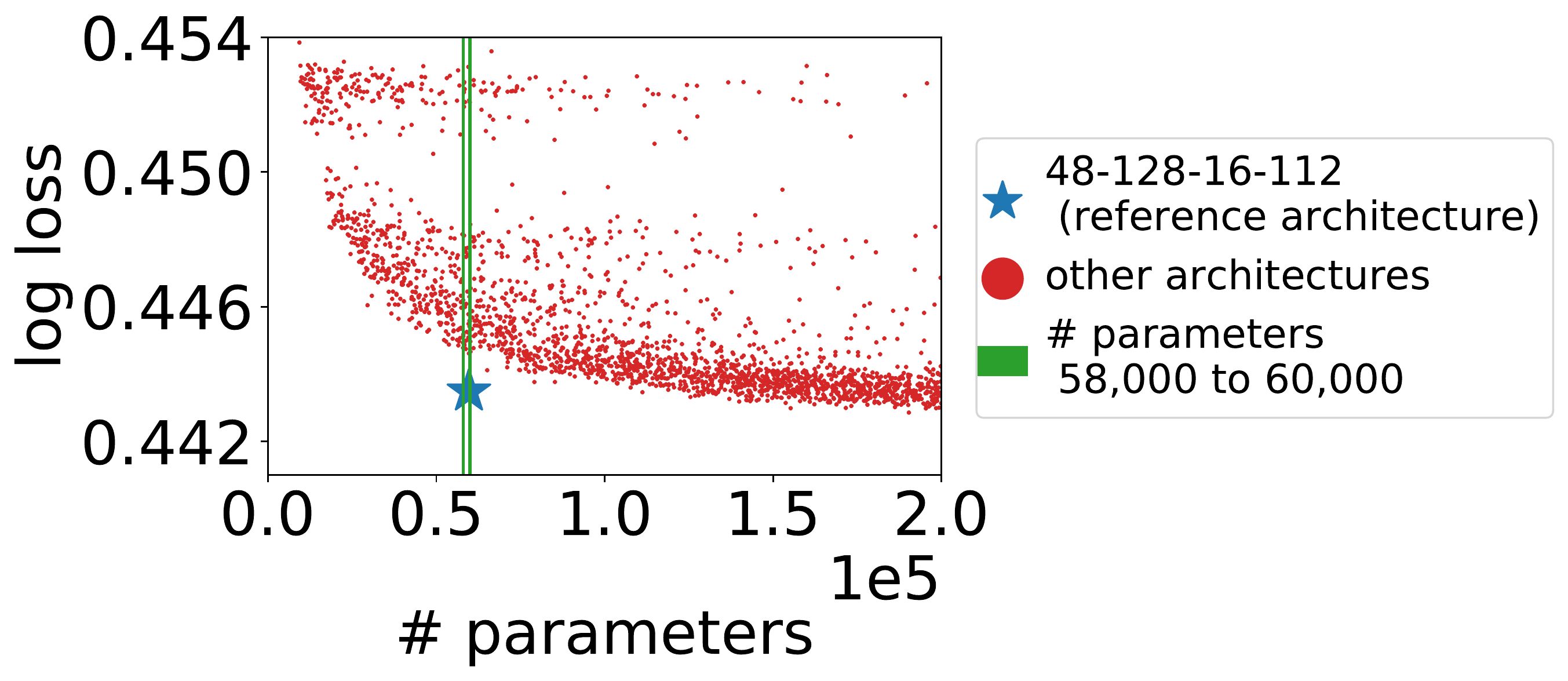}}
\hspace{.03\linewidth}
\subfigure[Tradeoff on Criteo, in 5 layer search space]{\label{fig:tradeoff_criteo_5_layers}\includegraphics[width=.45\linewidth]{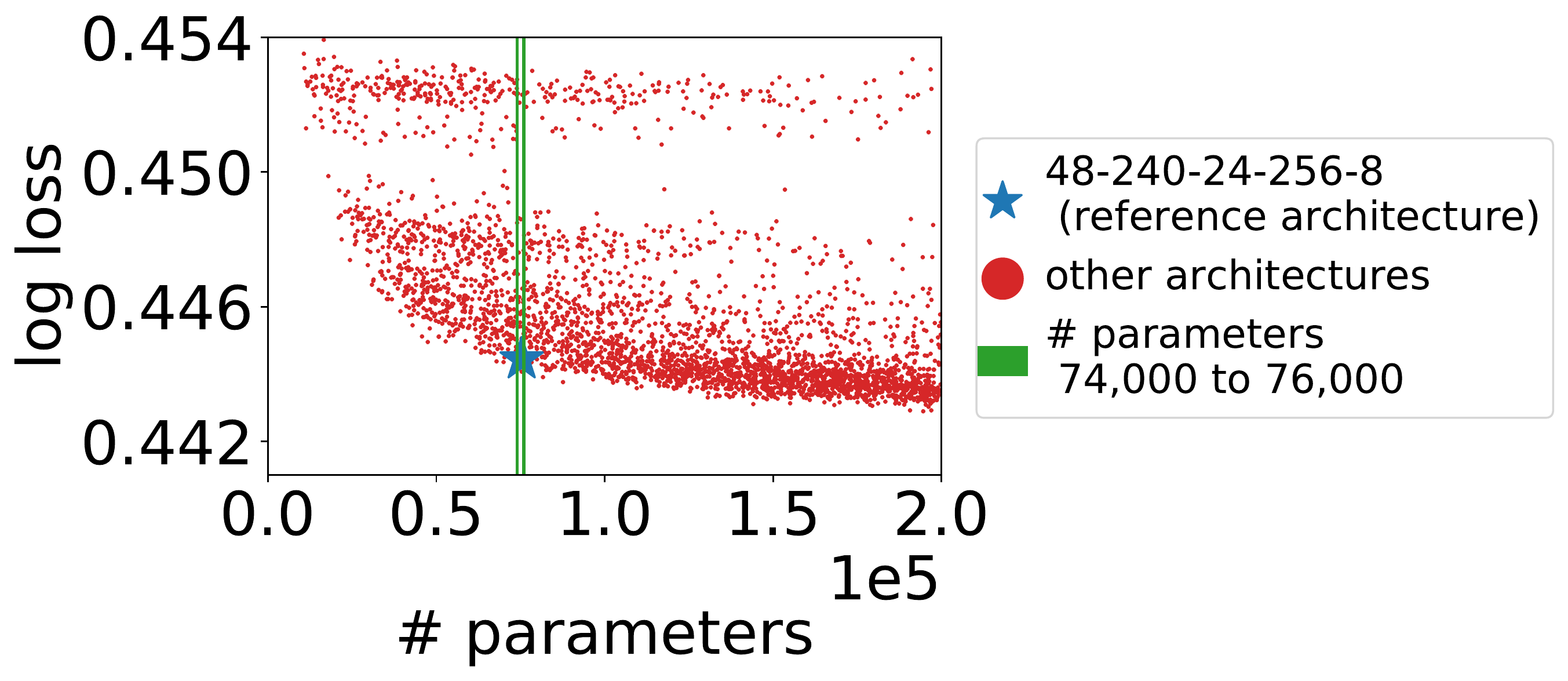}}

\subfigure[Tradeoff on Volkert, in 4 layer search space]{\label{fig:tradeoff_volkert_4_layers}\includegraphics[width=.45\linewidth]{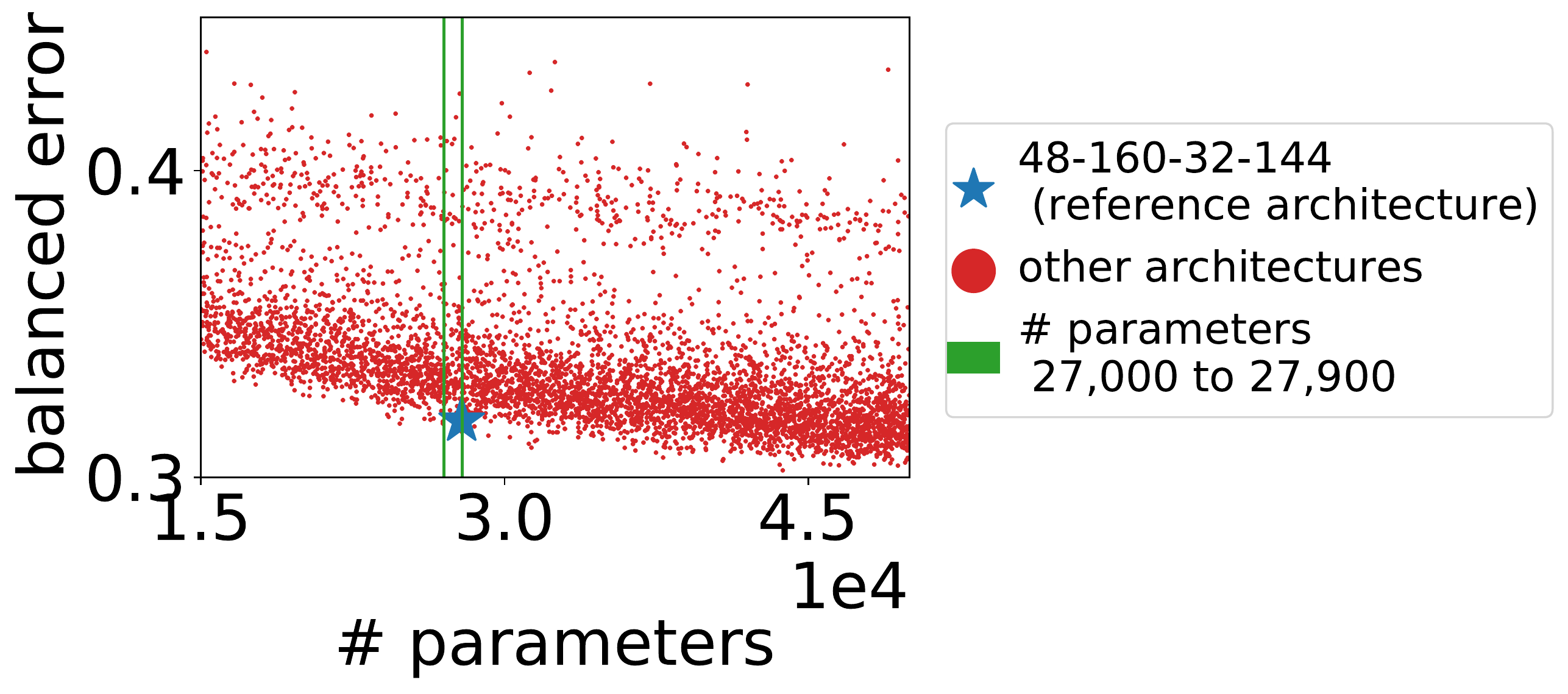}}
\hspace{.03\linewidth}
\subfigure[Tradeoff on Volkert, in 9 layer search space]{\label{fig:tradeoff_volkert_9_layers}\includegraphics[width=.45\linewidth]{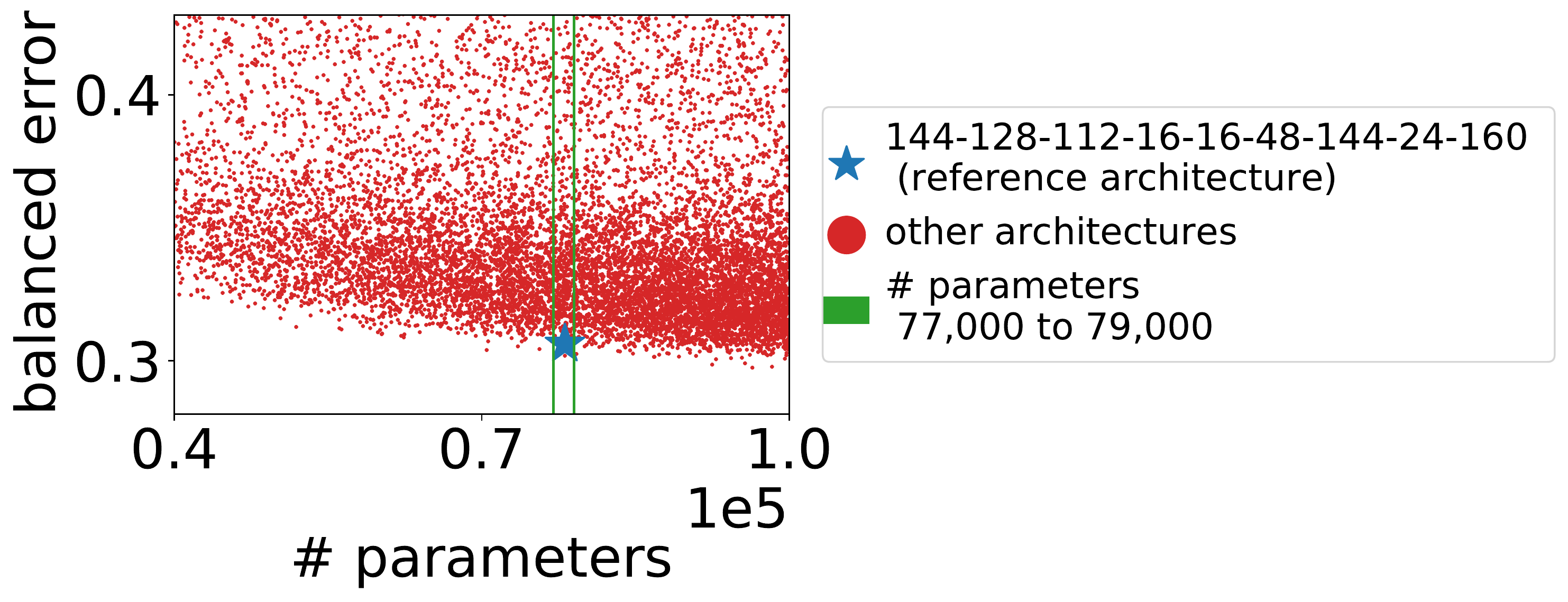}}

\caption{Tradeoffs between validation loss and number of parameters in four search spaces.
}
\label{fig:tradeoff_more}
\end{figure}

Figure~\ref{fig:tradeoff_more} shows the tradeoffs between loss and number of parameters in these search spaces. 
When training each architecture 5 times, the standard deviation (std) across different runs is 0.0002 for Criteo\footnote{On Criteo, ``a 0.001-level improvement (of logistic loss) is considered significant''~\cite{wang2021dcn}.} and 0.004 for Volkert, meaning that the architectures whose performance difference is larger than $2\times$ std are qualitatively different. 
We use Pareto-optimal architectures as the reference of resource-constrained NAS: we want an architecture that both matches (or even beats\footnote{Note that the Pareto optimality of the reference architecture is determined by only one round of random search. Thus because of the randomness across multiple training runs, the other architectures are likely to beat the reference architecture: a ``regression toward the mean''.}) the performance of the reference architecture and has no more parameters than the reference.
Most Pareto-optimal architectures in Figure~\ref{fig:tradeoff_more} have the bottleneck structure; Table~\ref{table:pareto_optimal_arch_examples} shows some examples.

\begin{table}[t]
\footnotesize
\caption{Some Pareto-optimal architectures in Figure~\ref{fig:tradeoff_more}.
All architectures shown here and almost all other Pareto-optimal architectures have the bottleneck structure.
}
\begin{center}
\begin{tabular}{ccccc}
\toprule
& search space & Pareto-optimal architecture & number of parameters & loss \\
\midrule
Figure~\ref{fig:tradeoff_criteo_4_layers} & Criteo 4-layer & 32-144-24-112 & 44,041 & 0.4454 \\
Figure~\ref{fig:tradeoff_criteo_4_layers} & Criteo 4-layer & 48-112-8-80 & 56,537 & 0.4448 \\
Figure~\ref{fig:tradeoff_criteo_4_layers} & Criteo 4-layer & 48-384-16-176 & 77,489 & 0.4441 \\
Figure~\ref{fig:tradeoff_criteo_4_layers} & Criteo 4-layer & 96-144-32-240 & 125,457 & 0.4433 \\
Figure~\ref{fig:tradeoff_criteo_4_layers} & Criteo 4-layer & 96-384-48-16 & 155,217 & 0.4430 \\ [1ex]
Figure~\ref{fig:tradeoff_criteo_5_layers} & Criteo 5-layer & 32-240-16-8-96 & 45,769 & 0.4451 \\
Figure~\ref{fig:tradeoff_criteo_5_layers} & Criteo 5-layer & 48-128-64-16-128 & 67,217 & 0.4446 \\
Figure~\ref{fig:tradeoff_criteo_5_layers} & Criteo 5-layer & 48-256-16-8-384 & 69,977 & 0.4443 \\
Figure~\ref{fig:tradeoff_criteo_5_layers} & Criteo 5-layer & 64-144-48-96-160 & 102,497 & 0.4437 \\
Figure~\ref{fig:tradeoff_criteo_5_layers} & Criteo 5-layer & 96-512-24-256-48 & 179,449 & 0.4430\\ [1ex]
Figure~\ref{fig:tradeoff_volkert_4_layers} & Volkert 4-layer & 48-112-16-24 & 16,642 & 0.3314 \\
Figure~\ref{fig:tradeoff_volkert_4_layers} & Volkert 4-layer & 32-112-24-224 & 20,050 & 0.3269 \\
Figure~\ref{fig:tradeoff_volkert_4_layers} & Volkert 4-layer & 48-160-32-144 & 27,882 & 0.3149 \\
Figure~\ref{fig:tradeoff_volkert_4_layers} & Volkert 4-layer & 48-256-24-112 & 31,330 & 0.3097 \\
Figure~\ref{fig:tradeoff_volkert_4_layers} & Volkert 4-layer & 80-208-32-64 & 40,778 & 0.3054 \\ [1ex]
Figure~\ref{fig:tradeoff_volkert_9_layers} & Volkert 9-layer & 64-64-160-48-16-144-16-8-48 & 40,482 & 0.3250 \\
Figure~\ref{fig:tradeoff_volkert_9_layers} & Volkert 9-layer & 80-144-32-112-32-8-128-8-144 & 43,290 & 0.3238 \\
Figure~\ref{fig:tradeoff_volkert_9_layers} & Volkert 9-layer & 112-144-32-32-24-24-24-128-32 & 51,890 & 0.3128 \\
Figure~\ref{fig:tradeoff_volkert_9_layers} & Volkert 9-layer & 144-128-112-16-16-48-144-24-160 & 78,114 & 0.3019 \\
Figure~\ref{fig:tradeoff_volkert_9_layers} & Volkert 9-layer & 160-144-144-32-112-32-48-32-144 & 94,330 & 0.3010 \\
\bottomrule
\end{tabular}
\end{center}
\label{table:pareto_optimal_arch_examples}
\end{table}

\subsubsection{More details on TPU implementation}
When we run one-shot NAS on a TPU that has multiple TPU cores (for example, each Cloud TPU-v2 we use has 8 cores), each core samples an architectures independently, and we use the average loss and reward for weight and RL updates, respectively.
This means our algorithm actually samples multiple architectures in each iteration and uses the \texttt{tensorflow.tpu.cross\_replica\_sum()} method to compute their average effect on the gradient.
Since only a fraction of architectures are feasible in each search space, we set the losses and rewards given by the infeasible architectures to 0 before averaging, so that we are equivalently only averaging across the sampled architectures that are feasible. 
We then reweight the average loss or reward with number\_of\_cores / number\_of\_feasible\_architectures to obtain an unbiased estimate.

\subsubsection{More details on the NAS method comparison plot (Figure~\ref{fig:comparison_with_random_criteo_3_layers})}
\label{appsec:distributional_performance_comparison_details}

For each architecture below, we report its number of parameters and mean $\pm$ std logistic loss across 5 stand-alone training runs in brackets.

We have the reference architecture 32-144-24 (41,153 parameters, 0.4454 $\pm$ 0.0003) for NAS methods to match.
In the search space with $20^3=8000$ candidate architectures:
\begin{itemize}[leftmargin=2em,topsep=0pt,partopsep=1ex,parsep=0ex]
\item TabNAS trials with no fewer than 2,048 Monte-Carlo samples and the RL learning rate $\eta$ among \{0.001, 0.005, 0.01\} consistently finds either the reference architecture itself, or an architectures that is qualitatively the same as the reference, like 32-112-32 (40,241 parameters, 0.4456 $\pm$ 0.0003). 
\item NAS with the Abs Reward: After grid search over RL learning rate $\eta$ (among \{0.0001, 0.0005, 0.001, 0.005, 0.01, 0.015, 0.02, 0.025, 0.03, 0.04, 0.05, 0.06, 0.07, 0.08, 0.09, 0.1, 0.15, 0.2, 0.25, 0.3, 0.4, 0.5, 0.75, 1.0, 1.5, 2.0\}) and $\beta$ (among \{-0.0005, -0.001, -0.005, -0.01, -0.05, -0.1, -0.5, -0.75, -1.0, -1.25, -1.5, -2.0, -3.0\}), the RL controller finds 32-64-96 (41,345 parameters, 0.4461 $\pm$ 0.0003) or 32-80-64 (40,785 parameters, 0.4459 $\pm$ 0.0002) among over $90\%$ trials that eventually find an architecture within $\pm 5\%$ of the target number of parameters 41,153.
\end{itemize}

\subsection{Difficulty in using the MnasNet reward}
With the MnasNet reward, only fewer than 1\% NAS trials in our hyperparameter grid search (the ones with a medium $\beta$) can find an architecture whose number of parameters is within $\pm 5\%$ of the reference, and among which none or only one (out of tens) can match the reference performance.
In contrast, TuNAS with the Abs Reward finds an architecture with number of parameters within $\pm 5\%$ of the reference among over 50\% of the grid search trials described in Appendix~\ref{appsec:distributional_performance_comparison_details}, and TabNAS with the rejection-based reward consistently finds such architectures at medium RL learning rates $\eta$ and decently large numbers of MC samples $N$.
This means it is significantly more difficult to use the MnasNet reward than competing approaches in the practice of resource-constrained tabular NAS.

\section{More failure cases of the Abs Reward, and performance of TabNAS}
\label{appsec:more_failure_cases}
For each architecture below, we report its number of parameters and mean $\pm$ std loss across 5 stand-alone training runs (logistic loss for Criteo, balanced error for the others) in brackets.

\begin{figure}[t]
\centering
\subfigure[Criteo, 4-layer search space]{\label{fig:comparison_with_random_criteo_4_layers}\includegraphics[width=.47\linewidth]{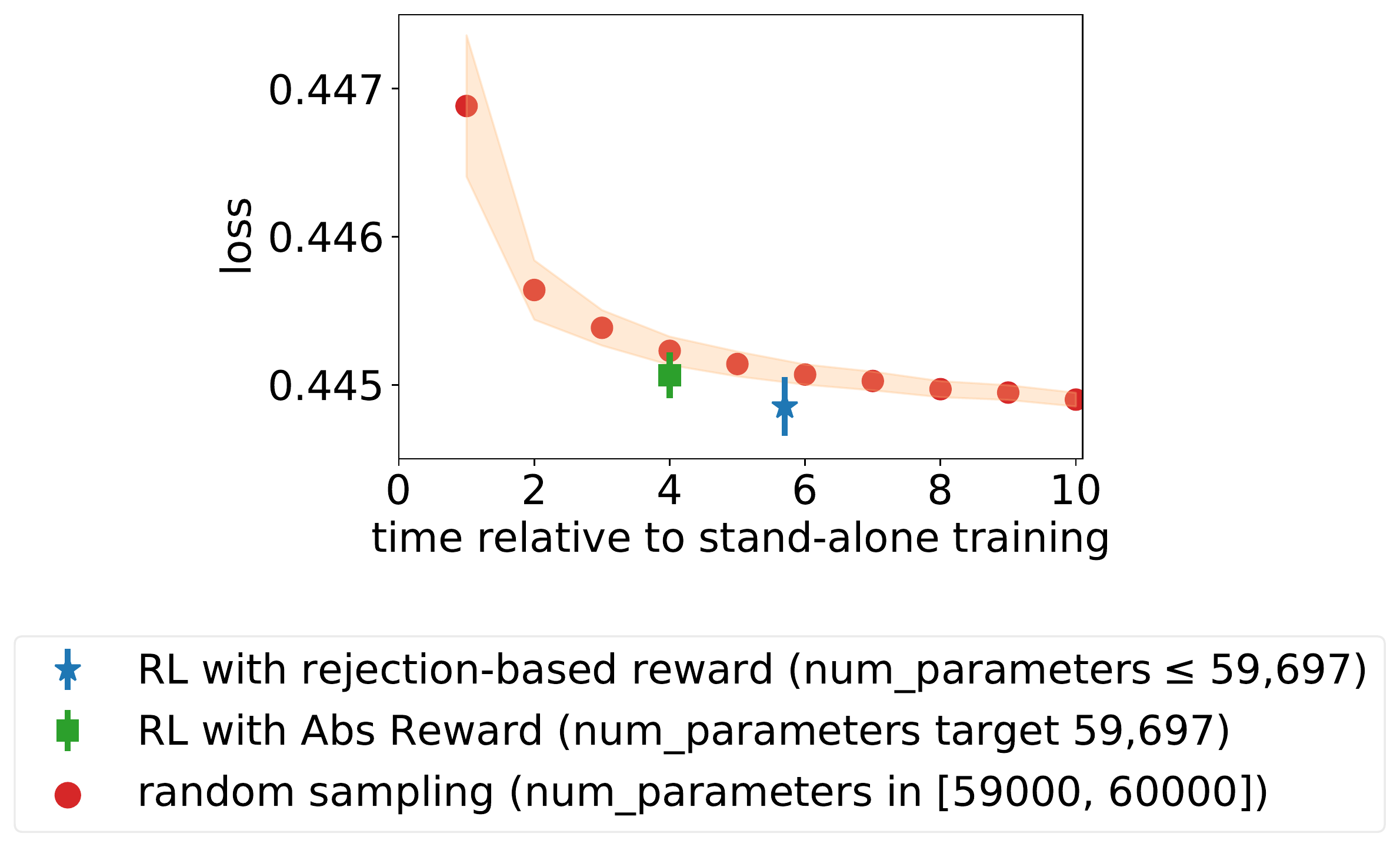}}
\hspace{.03\linewidth}
\subfigure[Criteo, 5-layer search space]{\label{fig:comparison_with_random_criteo_5_layers}\includegraphics[width=.47\linewidth]{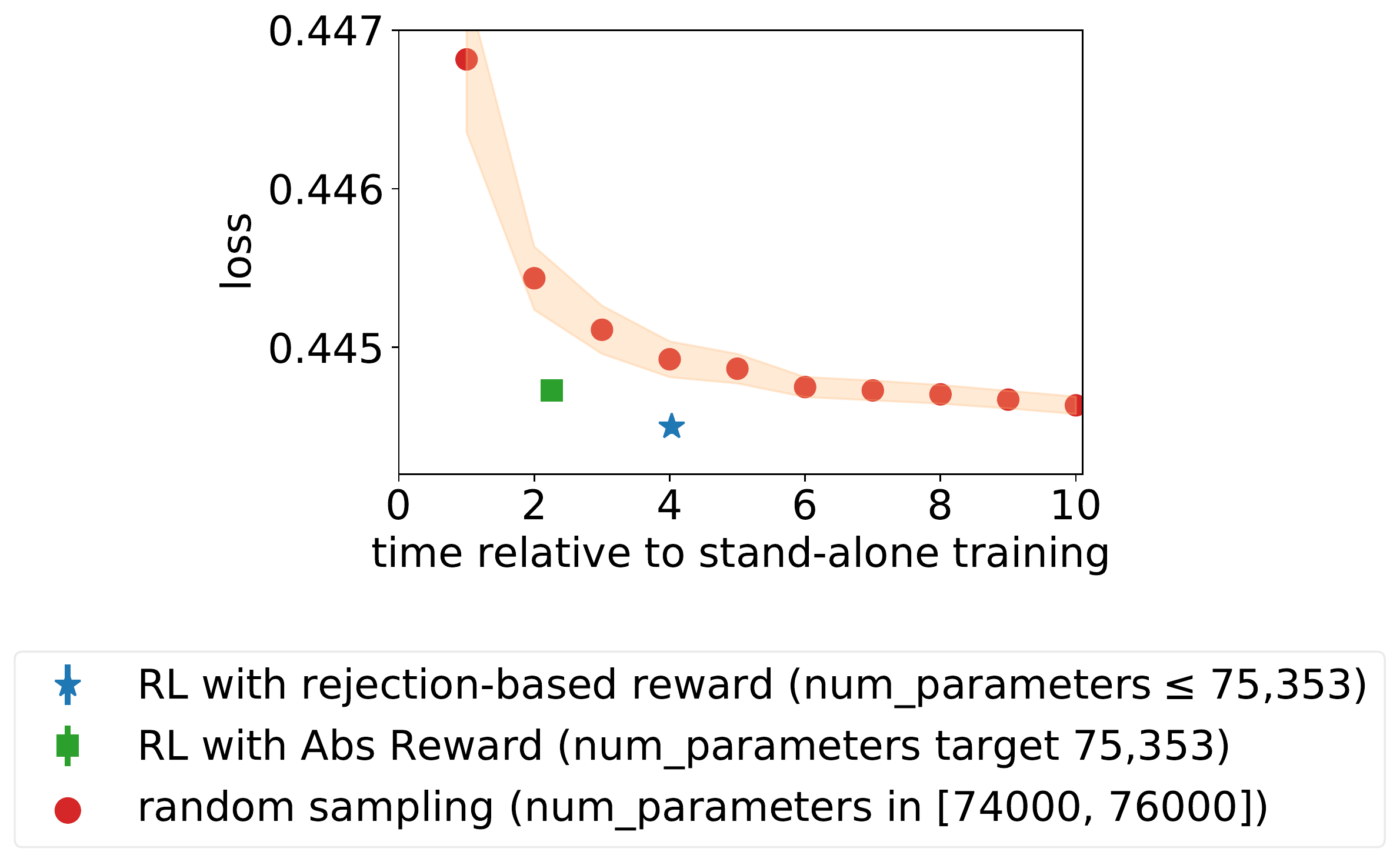}}

\subfigure[Volkert, 4-layer search space]{\label{fig:comparison_with_random_volkert_4_layers}\includegraphics[width=.47\linewidth]{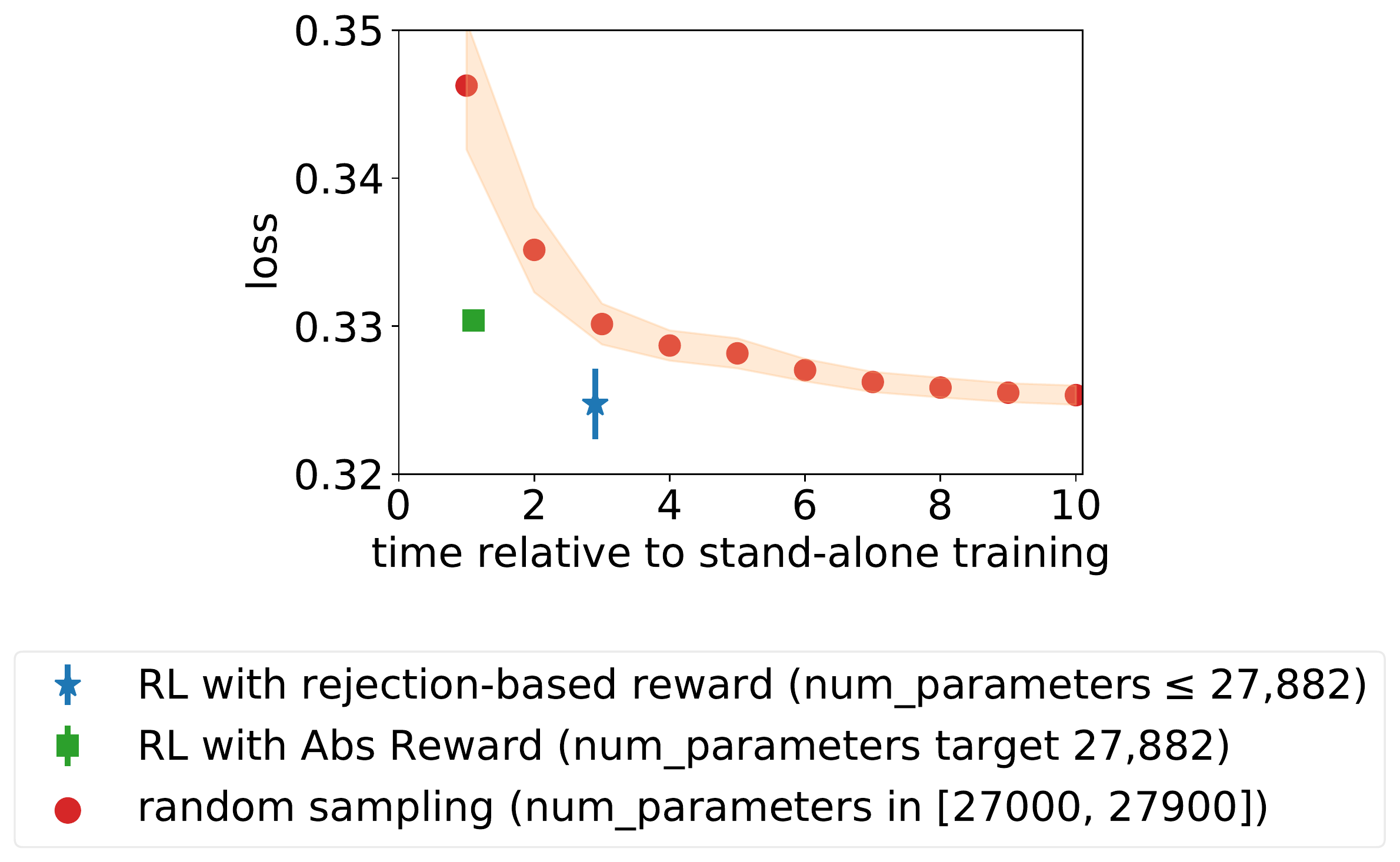}}
\hspace{.03\linewidth}
\subfigure[Volkert, 9-layer search space]{\label{fig:comparison_with_random_volkert_9_layers}\includegraphics[width=.47\linewidth]{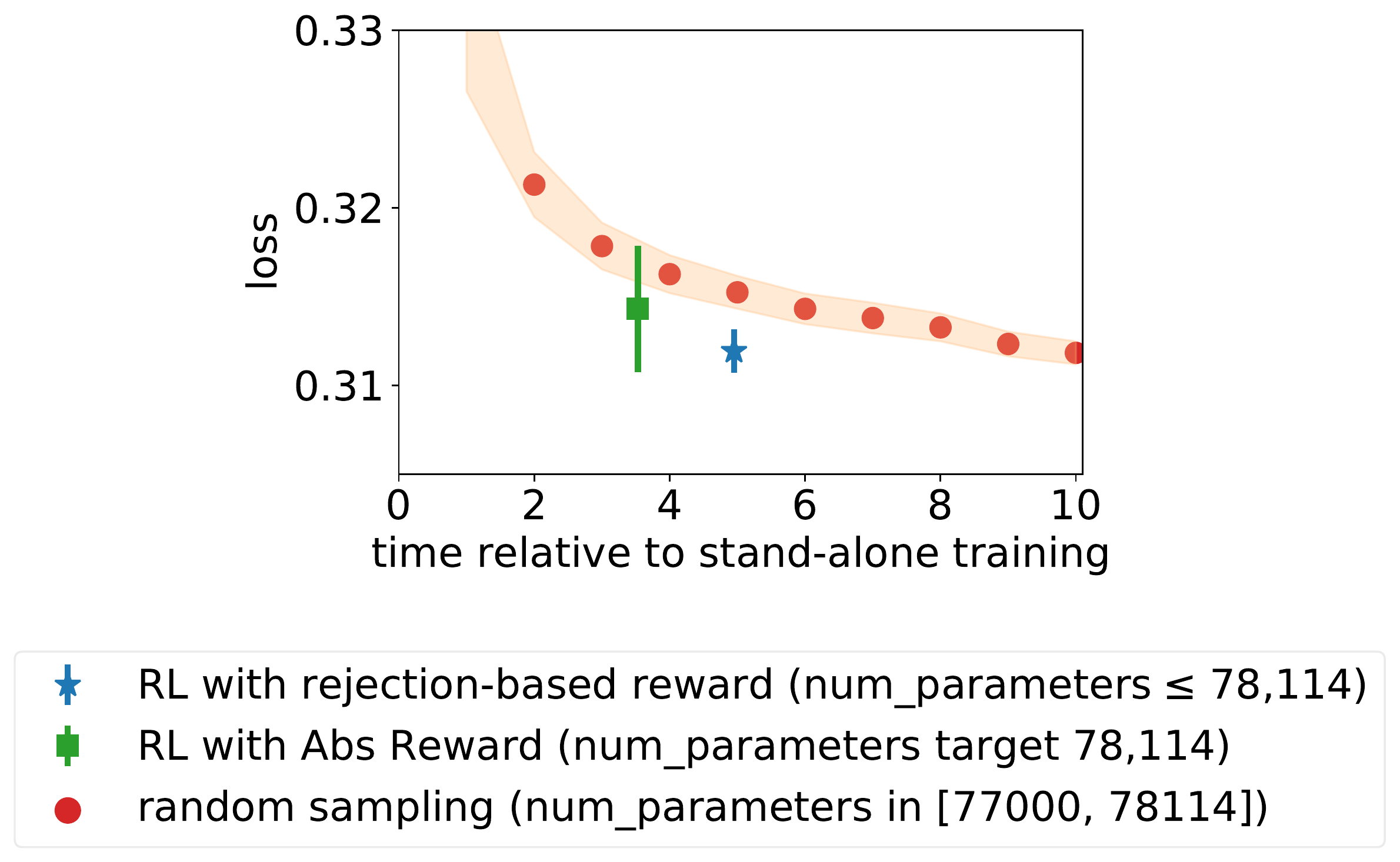}}

\caption{Rejection-based reward distributionally outperforms random search and resource-aware Abs Reward in a number of search spaces.
The points and error bars have the same meaning as in Figure~\ref{fig:comparison_with_random_criteo_3_layers}.
The time taken for each stand-alone training run (the unit length for x axes) is 2.5 hours on Criteo (Figure~\ref{fig:comparison_with_random_criteo_3_layers} in the main paper and \subref{fig:comparison_with_random_criteo_4_layers}, ~\subref{fig:comparison_with_random_criteo_5_layers}), 10 minutes on Volkert with 4-layer FFNs (Figure~\subref{fig:comparison_with_random_volkert_4_layers}), and 22-25 minutes on Volkert with 9-layer FFNs (Figure~\subref{fig:comparison_with_random_volkert_9_layers}).
}
\label{fig:comparison_with_random_appendix}
\end{figure}

\myparagraph{On Criteo, in the 4-layer search space}
We have the reference architecture 48-128-16-112 (59,697 parameters, 0.4451 $\pm$ 0.0002) for NAS to match in the search space (shown as Figure~\ref{fig:tradeoff_criteo_4_layers}).
Similar to Figure~\ref{fig:comparison_with_random_criteo_3_layers}, we show similar results on NAS with rejection-based reward (TabNAS) and NAS with the Abs Reward (TuNAS) in Figure~\ref{fig:comparison_with_random_criteo_4_layers}.
In the search space with $20^4=1.6 \times 10^5$ candidate architectures:
\begin{itemize}[leftmargin=2em,topsep=0pt,partopsep=1ex,parsep=0ex]
\item TabNAS with 32,768 Monte-Carlo samples and RL learning rate $\eta$ among \{0.001, 0.005, 0.01\} consistently finds architectures qualitatively the same as the reference. 
Example results include 48-128-24-32 (59,545 parameters, 0.4449 $\pm$ 0.0002), 48-144-16-48 (59,585 parameters, 0.4448 $\pm$ 0.0001), 48-112-16-144 (59,233 parameters, 0.4448 $\pm$ 0.0002) and the reference architecture itself. 
\item NAS with the Abs Reward successfully finds the reference architecture 48-128-16-112 in 3 out of 338 hyperparameter settings on a $\beta$-$\eta$ grid. 
Other found architectures include 48-80-32-112 (59,665 parameters, 0.4452 $\pm$ 0.0002), 32-128-80-144 (59,249 parameters, 0.4453 $\pm$ 0.0003) and 48-160-8-48 (58,953 parameters, 0.4448 $\pm$ 0.0003), among which the first two are inferior to the TabNAS-found counterparts.
\end{itemize}

\myparagraph{On Criteo, in the 5-layer search space}
We have the reference architecture 48-240-24-256-8 (75,353 parameters, 0.4448 $\pm$ 0.0002) for NAS methods to match in the search space (shown as Figure~\ref{fig:tradeoff_criteo_5_layers}).
Similar to Figure~\ref{fig:comparison_with_random_criteo_3_layers}, we have similar results on the comparison among random sampling, NAS with rejection-based reward (TabNAS), and NAS with the Abs Reward as Figure~\ref{fig:comparison_with_random_criteo_5_layers}.
In the search space with $20^5=3.2 \times 10^6$ candidate architectures:
\begin{itemize}[leftmargin=2em,topsep=0pt,partopsep=1ex,parsep=0ex]
\item TabNAS with 32,768 Monte-Carlo samples and the RL learning rate $\eta=0.005$ consistently finds architectures qualitatively the same as the reference. 
Example results include 48-176-64-16-256 (74,945 parameters, 0.4445 $\pm$ 0.0002), 48-208-48-48-64 (75,121 parameters, 0.4444 $\pm$ 0.0001), 48-256-32-80-24 (74,721 parameters, 0.4446 $\pm$ 0.0003) and 48-176-80-16-96 (75,153 parameters, 0.4445 $\pm$ 0.0002). 
\item NAS with the Abs Reward finds 64-80-48-8-8 (75,353 parameters, 0.4448 $\pm$ 0.0001), 64-80-24-16-112 (75,353 parameters, 0.4447 $\pm$ 0.0001), 48-144-96-16-192 (75,329 parameters, 0.4446 $\pm$ 0.0001) and 64-96-8-32-64 (75,273 parameters, 0.4445 $\pm$ 0.0001) that are mostly inferior to the TabNAS-found architectures. 
\end{itemize}

\myparagraph{On Volkert, in the 4-layer search space}
We have the reference architecture 48-160-32-144 (27,882 parameters, 0.3244 $\pm$ 0.0040) for NAS to match in the search space (shown as Figure~\ref{fig:tradeoff_volkert_4_layers}).
Similar to Figure~\ref{fig:comparison_with_random_criteo_3_layers}, we draw the comparison plot among random sampling, NAS with rejection-based reward (TabNAS), and NAS with the Abs Reward as Figure~\ref{fig:comparison_with_random_volkert_4_layers}.
In the search space with $1.6 \times 10^5$ candidate architectures:
\begin{itemize}[leftmargin=2em,topsep=0pt,partopsep=1ex,parsep=0ex]
\item TabNAS with $10^4$ Monte-Carlo samples and the RL learning rate $\eta \in \{0.001, 0.005, 0.01, 0.05\}$ consistently finds either the reference architecture itself or other architectures qualitatively the same. 
Examples include 
64-128-48-16 (27,050 parameters, 0.3237 $\pm$ 0.0040),  80-48-112-32 (27,802 parameters, 0.3274 $\pm$ 0.0037), 64-96-80-24 (27,778 parameters, 0.3279 $\pm$ 0.0005), and 64-144-32-48 (27,658 parameters, 0.3204 $\pm$ 0.0038).
\item NAS with the Abs Reward finds 96-64-32-48 (27,738 parameters, 0.3302 $\pm$ 0.0042), 96-48-32-96 (27,738 parameters, 0.3305 $\pm$ 0.0047), 96-80-16-48 (27,738 parameters, 0.3302 $\pm$ 0.0050), 112-48-24-24 (27,722 parameters, 0.3301 $\pm$ 0.0034) and 80-80-48-48 (27,690 parameters, 0.3309 $\pm$ 0.0022) that are inferior. 
\end{itemize}

\myparagraph{On Volkert, in the 9-layer search space}
We further do NAS on Volkert in the 9-layer search space to test the ability of TabNAS in searching among significantly deeper FFNs.
The tradeoff between loss and number of parameters in the search space is shown in Figure~\ref{fig:tradeoff_volkert_9_layers}.
We have the reference architecture 144-128-112-16-16-48-144-24-160 (78,114 parameters, 0.3126 $\pm$ 0.0050) for NAS to match.
We compare random sampling, NAS with rejection-based reward (TabNAS), and NAS with the Abs Reward in Figure~\ref{fig:comparison_with_random_volkert_9_layers}.
In the search space with $5.2 \times 10^9$ candidate architectures (which is nearly impossible for exhaustive search):
\begin{itemize}[leftmargin=2em,topsep=0pt,partopsep=1ex,parsep=0ex]
\item TabNAS with $5 \times 10^6$ Monte-Carlo samples and the RL learning rate $\eta \in \{0.002, 0.005\}$ consistently finds architectures that are qualitatively the same as the reference.
These architectures are found when the RL controller is far from converged and when $\probP(V)$ slightly decreases after RL starts. 
Example results include 
144-144-112-64-24-16-128-8-128 (78,026 parameters, 0.3120 $\pm$ 0.0049), 128-160-96-32-24-64-64-32-160 (77,890 parameters, 0.3127 $\pm$ 0.0040), 128-144-112-32-64-64-80-16-128 (77,834 parameters, 0.3094 $\pm$ 0.0012), 160-128-96-32-48-64-48-24-112 (78,002 parameters, 0.3137 $\pm$ 0.0021), and 144-112-160-24-112-16-16-128-48 (77,986 parameters, 0.3119 $\pm$ 0.0029).
\item NAS with the Abs Reward finds 144-96-80-80-48-64-96-80-32 (78,170 parameters, 0.3094 $\pm$ 0.0039), 160-80-160-24-80-16-80-64-128 (78,114 parameters, 0.3158 $\pm$ 0.0020), 128-96-80-80-64-64-80-80-80 (78,106 parameters, 0.3128 $\pm$ 0.0020), and 144-128-80-16-16-16-96-160-24 (78,050 parameters, 0.3192 $\pm$ 0.0014).
Interestingly, all architectures except 144-96-80-80-48-64-96-80-32 are inferior to the TabNAS-found architectures despite having slightly more parameters, and 144-96-80-80-48-64-96-80-32 does not have an evident bottleneck structure like the other architectures found here. 
\end{itemize}

\section{Tabulated performance of different RL rewards on all datasets}
\label{appsec:tabulated_performance_of_rewards_on_datasets}
Table~\ref{table:tabulated_results_of_rewards_on_datasets} summarizes among results of RL with the rejection-based reward, the Abs Reward, two MNasNet rewards and the reward in RENA~\cite{zhou2018resource} Equation 3.
Bold results in each column indicate architectures that are on par with the best for the corresponding dataset.
The rejection-based reward in TabNAS gets the best architectures overall.

\begin{table}
\scriptsize
\caption{Architectures found by RL with 5 reward functions on 5 tabular datasets in Table~\ref{table:dataset_details}.
Each reward-dataset combination has 3 repeated NAS runs.
The bracket ``(2 times)'' indicates the same architecture was found by the corresponding reward function twice.
}
\begin{tabular}{P{3cm}P{1.8cm}P{1.8cm}P{1.8cm}P{1.8cm}P{1.8cm}}
\toprule
& Criteo & Volkert & Aloi & Connect-4 & Higgs \\
\midrule
reference architecture & \textbf{32-144-24 (41,153 parameters, 0.4454 $\pm$ 0.0003)} & \textbf{48-160-32-144 (27,882 parameters, 0.3244 $\pm$ 0.0040)} & 160-64-512-64 (162,056 parameters, 0.0470 $\pm$ 0.0004) & \textbf{32-22-4-20 (3,701 parameters, 0.2691 $\pm$ 0.0021)} & \textbf{16-144-16 (5,265 parameters, 0.2794 $\pm$ 0.0013)} \\\hline
RL with rejection-based reward & \textbf{32-144-24 (2 times) (41,153 parameters, 0.4454 $\pm$ 0.0003) 32-112-32 (40,241 parameters, 0.4456 $\pm$ 0.0003)} & \textbf{64-128-48-16 (27,050 parameters, 0.3237 $\pm$ 0.0040) 48-160-32-144 (27,882 parameters, 0.3244 $\pm$ 0.0040) 80-48-112-32 (27,802 parameters, 0.3274 $\pm$ 0.0037)} & \textbf{176-144-144-80 (2 times) (161,672 parameters, 0.0458 $\pm$ 0.0007) 192-112-176-80 (161,432 parameters, 0.0461 $\pm$ 0.0012)} & 20-32-14-48 (3,701 parameters, 0.2827 $\pm$ 0.0062) 20-30-24-22 (3,693 parameters, 0.2795 $\pm$ 0.0049) 28-10-12-56 (3,701 parameters, 0.2819 $\pm$ 0.0052) & 72-24-48 (2 times) (5,161 parameters, 0.2911 $\pm$ 0.0031) 80-32-8 (5,265 parameters, 0.2893 $\pm$ 0.0022) 64-16-128 (5,265 parameters, 0.2870 $\pm$ 0.0033) \\\hline
RL with Abs Reward & 32-64-96 (2 times) (41,345 parameters, 0.4461 $\pm$ 0.0003) 32-80-64 (40,785 parameters, 0.4459 $\pm$ 0.0002) & 96-64-32-48 (27,738 parameters, 0.3302 $\pm$ 0.0042) 96-48-32-96 (27,738 parameters, 0.3305 $\pm$ 0.0047) 96-80-16-48 (27,738 parameters, 0.3302 $\pm$ 0.0050) & 144-112-144-96 (3 times) (162,008 parameters, 0.0473 $\pm$ 0.0004) 128-96-96-112 (4 times) (162,072 parameters, 0.0488 $\pm$ 0.0012) 112-112-96-112 (161,816 parameters, 0.0502 $\pm$ 0.0007) & \textbf{30-22-12-12 (3,703 parameters, 0.2702 $\pm$ 0.0027)} 26-20-18-26 (2 times) (3,703 parameters, 0.2807 $\pm$ 0.0029) 26-18-20-26 (3,703 parameters, 0.2765 $\pm$ 0.0046) & 104-16-24 (2 times) (5,233 parameters, 0.2896 $\pm$ 0.0036) 80-16-88 (2 times) (5,281 parameters, 0.2887 $\pm$ 0.0021) 88-16-64 (2 times)  (5,217 parameters, 0.2874 $\pm$ 0.0027)\\\hline
MNasNet (reward type 1: $Q(y) (T(y) / T_0)^\beta$) & 32-64-16 (36,065 parameters, 0.4464 $\pm$ 0.0003) 32-64-8 (35,537 parameters, 0.4466 $\pm$ 0.0002) 32-24-64 (35,353 parameters, 0.4466 $\pm$ 0.0003) & only found one feasible architecture: 32-32-224-24 (19,890 parameters, 0.3392 $\pm$ 0.0043) & 160-128-112-96 (163,544 parameters, 0.0469 $\pm$ 0.0009) 128-96-96-112 (162,072 parameters, 0.0488 $\pm$ 0.0012) 128-176-128-80 (153,192 parameters, 0.0473 $\pm$ 0.0007) & 24-18-16-44 (3,677 parameters, 0.2788 $\pm$ 0.0046) 28-12-32-14 (3,651 parameters, 0.2781 $\pm$ 0.0052) 22-32-16-22 (3,577 parameters, 0.2818 $\pm$ 0.0032) & 72-24-32 (4,745 parameters, 0.2898 $\pm$ 0.0019) 64-32-16 (4,545 parameters, 0.2888 $\pm$ 0.0018) 72-24-24 (2 times) (4,537 parameters, 0.2903 $\pm$ 0.0036)\\\hline
MNasNet (reward type 2: $Q(y) \max\{1, (T(y) / T_0)^\beta\}$) & 24-128-16 (29,953 parameters, 0.4468 $\pm$ 0.0005) 16-176-48 (27,985 parameters, 0.4478 $\pm$ 0.0006) 24-16-160 (27,953 parameters, 0.4479 $\pm$ 0.0003) & only found one feasible architecture: 80-24-24-24 (17,874 parameters, 0.3521 $\pm$ 0.0013) & 112-144-240-64 (145,944 parameters, 0.0480 $\pm$ 0.0005) 160-112-128-64 (126,392 parameters, 0.0497 $\pm$ 0.0019) 256-64-224-32 (104,232 parameters, 0.0548 $\pm$ 0.0015) & 20-12-44-24 (3,679 parameters, 0.2871 $\pm$ 0.0069) 28-16-10-52 (3,745 parameters, 0.2776 $\pm$ 0.0061) 14-40-28-20 (3,581 parameters, 0.2886 $\pm$ 0.0047) & 16-8-16 (2 times) (777 parameters, 0.2929 $\pm$ 0.0020) 8-8-24 (553 parameters, 0.3003 $\pm$ 0.0031) \\\hline
RENA~\cite{zhou2018resource} & 32-16-48 (34,289 parameters, 0.4472 $\pm$ 0.0002) 24-48-160 (33,873 parameters, 0.4468 $\pm$ 0.0001) 8-64-96 (15,137 parameters, 0.4523 $\pm$ 0.0004) & only found one feasible architecture: 32-48-24-512 (26,482 parameters, 0.3389 $\pm$ 0.0039) & didn’t find any (close to) feasible architecture & 24-26-20-16 (3,617 parameters, 0.2806 $\pm$ 0.0068) 30-18-10-32 (3,749 parameters, 0.2752 $\pm$ 0.0075) 26-26-18-8 (3,577 parameters, 0.2769 $\pm$ 0.0031) & only found one feasible architecture: 80-8-152 (4,569 parameters, 0.2882 $\pm$ 0.0016)\\
\bottomrule
\end{tabular}
\label{table:tabulated_results_of_rewards_on_datasets}
\end{table}

\section{Comparison with Bayesian optimization and evolutionary search in one-shot NAS}
\label{appsec:comparison_with_bo_and_es}

Bayesian optimization (BO) and evolutionary search (ES) are popular strategies for NAS (e.g., \cite{jin2019auto,kandasamy2018neural,white2019bananas,zhou2019bayesnas,shi2020bridging}; \cite{liu2018progressive,elsken2018efficient,awad2020differential,guerrero2021bag}).
We are not aware of any work that successfully applies BO to one-shot NAS for tabular data; \citet{guo2020single} proposed an ES approach for one-shot NAS on vision datasets.
BO and ES omit the extra forward passes for RL controller training, but need extra forward passes to evaluate child networks from SuperNet weights (Figure~\ref{fig:supernet_illustration} in the main paper). 
Thus we control the number of forward passes for a fair comparison\footnote{RL also does backward passes to optimize over the logits $\{\ell_{ij}\}_{i \in [L], j \in [C_i]}$, but the number of logits in our setting is much fewer than the size of the validation set (100 vs. 4,582,432), which means the cost of forward passes dominates.}.
We design the following one-shot NAS methods to compare with an $M$-epoch TabNAS that interleaves weight and RL updates in its latter $75\%$ iterations: 
\begin{itemize}[leftmargin=2em,topsep=0pt,partopsep=1ex,parsep=0ex]
\item \textbf{train-then-search-with-RL}: Train the SuperNet for $M$ epochs as in TabNAS, then do NAS for $0.75M$ epochs with rejection-based RL (with each iteration as Algorithm~\ref{alg:mc}).
\item \textbf{train-then-search-with-BO}: Train the SuperNet for $M$ epochs as in TabNAS, then do BO (by Gaussian processes~\cite{rasmussen2003gaussian} with expected improvement~\cite{movckus1975bayesian,jones1998efficient}) in the set of feasible architectures with a similar number of SuperNet forward passes for child network evaluation.
\item \textbf{train-then-search-with-ES}: Train the SuperNet for $M$ epochs as in TabNAS, then do ES in the set of feasible architectures as Algorithm~1 in~\citet{guo2020single} with a similar number of SuperNet forward passes for child network evaluation: in each iteration, start with population size $P$, pick top-$k$ architectures, crossover and mutate these top-$k$ to each get $P/2$ architectures, then combine the crossover and mutation results to get the population for the next iteration.
\end{itemize}

On Criteo, the cost of forward passes for RL is comparable to evaluating 405 child networks on the validation set.
The search space of 5-layer FFNs has 340,590 feasible architectures below the 75,353 parameters limit (corresponding reference architecture is 48-240-24-256-8) in Figure~\ref{fig:tradeoff_criteo_5_layers}. 
In BO, we tune RBF kernel length scale, number of initially sampled architectures, and number of new architectures to sample in each step; in ES, we tune population size and $k$.
We can see from the results in Table~\ref{table:comparison_with_bo_and_es} that:
\begin{itemize}[leftmargin=2em,topsep=0pt,partopsep=1ex,parsep=0ex]
\item RL-based methods (TabNAS or train-then-search-with-RL) stably finds architectures that are qualitatively the best.
\item There is a large variance across architectures found by each hyperparameter setting of BO or ES.
The search results are sensitive to initialization and are worse than those found by RL in over $2/3$ trials.
We observe that each of the BO and ES searches quickly gets stuck at an architecture that is close to the best architecture in the initial set of samples.
\item The local optima that BO and ES get stuck at still often have bottleneck structures, but the number of parameters is often significantly below the limit (e.g., 63,817 parameters in the searched model vs.~a limit of 75,353 parameters).
Model performance suffers as a result.
\end{itemize}

Many interesting questions on BO and ES for one-shot NAS remain open for future work, including how to control the initialization randomness (for better exploration of the search space) and how to design methods that properly interleave weight training and NAS steps under such large exploration randomness (for better exploitation of promising architectures).

\begin{table}
\scriptsize
\caption{TabNAS vs.~train-then-search-with-RL/BO/ES for one-shot NAS on Criteo, among 5-layer FFNs.
The reference architecture 48-240-24-256-8 has 75,353 parameters and validation loss 0.4448 $\pm$ 0.0002.
Bold results below indicate architectures that are on par with the best validation loss 0.4444 $\pm$ 0.0002.
We run each method 3 times, except for TabNAS, which was detailed in Appendix~\ref{appsec:more_failure_cases}.
The square bracket ``[3 times]'' indicates that the same architecture was found by the corresponding method three times.
}
\begin{center}
\begin{tabular}{P{3cm}P{6.5cm}P{3.2cm}}
\toprule
method & found architectures (number of parameters, mean $\pm$ std loss) & NAS cost\\
\midrule
TabNAS ($\eta=0.005$ or $0.001$, $N=32768$) & \textbf{48-176-64-16-256 (74,945 parameters, 0.4445 $\pm$ 0.0002)} \textbf{48-208-48-48-64 (75,121 parameters, 0.4444 $\pm$ 0.0001)} \textbf{48-176-80-16-96 (75,153 parameters, 0.4445 $\pm$ 0.0002)} & $1.86 \times 10^9$ forward passes (405 child network evaluations)\\[8ex]
train-then-search-with-RL ($\eta=0.0005$ or $0.001$, $N=32768$) & \textbf{48-64-224-16-256 (75,249 parameters, 0.4446 $\pm$ 0.0002)} \textbf{48-240-16-24-384 (75,353 parameters, 0.4445 $\pm$ 0.0002)} \textbf{48-176-64-16-256 (74,945 parameters, 0.4445 $\pm$ 0.0002)} & $1.86 \times 10^9$ forward passes (405 child network evaluations)\\[8ex]
train-then-search-with-BO (RBF kernel L=10, white noise variance 1, 50 initial samples, 20 new samples per BO iteration) & 64-80-8-16-16 (72,073 parameters, 0.4447 $\pm$ 0.0002) \textbf{48-384-16-16-16 (74,881 parameters, 0.4444 $\pm$ 0.0002)} 48-256-16-8-240 (68,537 parameters, 0.4447 $\pm$ 0.0002) & 153 child network evaluations, then gets stuck\\[12ex]
train-then-search-with-BO (RBF kernel L=1, white noise variance 1, 100 initial samples, 10 new samples per BO iteration) & 48-80-48-96-96 (71,265 parameters, 0.4451 $\pm$ 0.0003) 48-128-8-64-8 (57,753 parameters, 0.4451 $\pm$ 0.0002) \textbf{64-96-16-32-16 (74,673 parameters, 0.4446 $\pm$ 0.0002)} & 420 child network evaluations \\ [12ex]
train-then-search-with-BO (RBF kernel L=1, white noise variance 1, 50 initial samples, 20 new samples per BO iteration) & \textbf{64-112-16-8-16 (75,177 parameters, 0.4445 $\pm$ 0.0001)} 48-32-256-8-240 (63,817 parameters, 0.4454 $\pm$ 0.0003) 64-96-16-24-48 (75,241 parameters, 0.4447 $\pm$ 0.0002) & 430 child network evaluations\\ [12ex]
train-then-search-with-BO (RBF kernel L=0.1, white noise variance 1, 50 initial samples, 20 new samples per BO iteration) & [3 times] 64-96-16-24-48 (75,241 parameters, 0.4447 $\pm$ 0.0002) & 430 child network evaluations\\ [12ex]
train-then-search-with-BO (RBF kernel L=0.1, white noise variance 1, 100 initial samples, 10 new samples per BO iteration) & [3 times] 64-96-16-24-48 (75,241 parameters, 0.4447 $\pm$ 0.0002) & 430 child network evaluations\\[12ex]
train-then-search-with-ES (population $50$, $k=10$) & 48-96-32-224-16 (68,161 parameters, 0.4450 $\pm$ 0.0003) 48-48-8-160-64 (63,897 parameters, 0.4455 $\pm$ 0.0003) 48-96-32-24-64 (59,609 parameters, 0.4448 $\pm$ 0.0001) & 424 child network evaluations\\[7ex]
train-then-search-with-ES (population $25$, $k=20$) & 48-144-32-16-144 (64,161 parameters, 0.4447 $\pm$ 0.0002) 48-80-16-128-64 (65,057 parameters, 0.4453 $\pm$ 0.0002) 48-96-32-48-64 (61,937 parameters, 0.4451 $\pm$ 0.0001) & 469 child network evaluations\\
\bottomrule
\end{tabular}
\end{center}
\label{table:comparison_with_bo_and_es}
\end{table}

\section{Rejection-based reward outperforms Abs Reward in NATS-Bench size search space (full version)}
\label{appsec:comparison_in_natsbench_size_search_space}

 In the vision domain, the NATS-Bench~\cite{dong2021nats} size search space has convolutional network architectures with a predefined skeleton and 8 candidate sizes \{8, 16, 24, 32, 40, 48, 56, 64\} for each of its 5 layers. 
 In this search space with $8^5=32768$ candidate architectures, we use the true number of floating point operations (\#FLOPs) as our cost metric, validation accuracy on CIFAR-100~\cite{krizhevsky2009learning} as RL quality reward, and test error (1 - test accuracy) on CIFAR-100 as final performance metric (the use of error instead of accuracy is consistent with other results here).
 We use 75M \#FLOPs as the resource limit, so that there are 13,546 (41.3\%) feasible architectures in the search space. 
 This experiment setting is the same as in the toy example (Figure~\ref{fig:toy_example}), where we regard the network weights as given and only compare the RL controllers. 
 We do grid search over NAS hyperparameters: \{0.01, 0.05, 0.1, 0.5\} for the RL learning rate for both the rejection-based reward and the Abs Reward, and \{10, 5, 2, 1, 0.5, 0.2, 0.1\} for $|\beta|$ in the Abs Reward. We show \#FLOPs and test error statistics (mean $\pm$ std) of architectures found by the rejection-based RL reward and the Abs Reward across 500 NAS repetitions of each experiment setting.

Detailed results of \#FLOPs and test errors of architectures found by either of the RL rewards at different NAS hyperparameters are listed in Table~\ref{table:flops_in_nats} and~\ref{table:errors_in_nats}, respectively.

\begin{table}[t]
\footnotesize
\caption{\#FLOPs (M) of architectures found by RL with the rejection-based reward and the Abs Reward in NATS-Bench size search space.}
\begin{tabular}{P{4cm}P{1.9cm}P{1.9cm}P{1.9cm}P{1.9cm}}
\hline
\toprule
RL learning rate & 0.01 & 0.05 & 0.1 & 0.5 \\ 
\midrule
rejection-based reward, $N$=200 & 74.7 $\pm$ 0.3 (median 74.7) & 74.7 $\pm$ 0.2 (median 74.7) & 74.6 $\pm$ 0.3 (median 74.7) & 70.7 $\pm$ 4.5 (median 72.4)  \\[4ex]
Abs Reward, $\beta$=-10 & 77.0 $\pm$ 12.7 (median 76.4)   & 75.1 $\pm$ 4.5 (median 74.8) & 75.2 $\pm$ 1.7 (median 75.1) & 75.7 $\pm$ 3.8 (median 75.2)  \\[4ex]
Abs Reward, $\beta$=-5 & 78.1 $\pm$ 13.1 (median 77.0)   & 75.6 $\pm$ 4.3 (median 75.5) & 75.4 $\pm$ 1.7 (median 75.2) & 76.0 $\pm$ 4.4 (median 75.3)  \\[4ex]
Abs Reward, $\beta$=-2 & 80.6 $\pm$ 13.8 (median 78.5)   & 76.3 $\pm$ 4.2 (median 76.1) & 75.7 $\pm$ 1.8 (median 75.5) & 75.9 $\pm$ 5.4 (median 75.4)  \\[4ex]
Abs Reward, $\beta$=-1 & 83.0 $\pm$ 13.3 (median 81.9)   & 77.4 $\pm$ 4.4 (median 77.1) & 76.0 $\pm$ 2.1 (median 75.8) & 76.7 $\pm$ 5.3 (median 75.6)  \\[4ex]
Abs Reward, $\beta$=-0.5 & 87.1 $\pm$ 12.7 (median 86.7)   & 78.8 $\pm$ 4.5 (median 78.7) & 77.0 $\pm$ 2.6 (median 76.6) & 76.9 $\pm$ 6.0 (median 76.1)  \\[4ex]
Abs Reward, $\beta$=-0.2 & 95.5 $\pm$ 12.8 (median 95.3)   & 80.6 $\pm$ 5.2 (median 80.1) & 78.5 $\pm$ 3.6 (median 77.9) & 78.5 $\pm$ 9.2 (median 76.8)  \\[4ex]
Abs Reward, $\beta$=-0.1 & 101.8 $\pm$ 13.3 (median 103.0) & 84.2 $\pm$ 7.6 (median 83.0) & 81.1 $\pm$ 5.8 (median 80.5) & 80.1 $\pm$ 10.6 (median 78.6) \\
\bottomrule
\end{tabular}
\label{table:flops_in_nats}
\end{table}

\begin{table}[t]
\footnotesize
\caption{Test errors of architectures found by RL with the rejection-based reward and the Abs Reward in NATS-Bench size search space.}
\begin{tabular}{ccccc}
\hline
\toprule
RL learning rate & 0.01 & 0.05 & 0.1 & 0.5 \\ 
\midrule
rejection-based reward, $N$=200 & 0.468 $\pm$ 0.050 & 0.435 $\pm$ 0.023 & 0.425 $\pm$ 0.014 & 0.420 $\pm$ 0.008\\
Abs Reward, $\beta$=-10 & 0.483 $\pm$ 0.056 & 0.468 $\pm$ 0.034 & 0.473 $\pm$ 0.046 & 0.490 $\pm$ 0.074\\
Abs Reward, $\beta$=-5 & 0.474 $\pm$ 0.049 & 0.461 $\pm$ 0.029 & 0.463 $\pm$ 0.039 & 0.481 $\pm$ 0.061\\
Abs Reward, $\beta$=-2 & 0.462 $\pm$ 0.039 & 0.450 $\pm$ 0.022 & 0.455 $\pm$ 0.028 & 0.468 $\pm$ 0.046\\
Abs Reward, $\beta$=-1 & 0.449 $\pm$ 0.027 & 0.442 $\pm$ 0.016 & 0.445 $\pm$ 0.020 & 0.456 $\pm$ 0.035\\
Abs Reward, $\beta$=-0.5 & 0.436 $\pm$ 0.019 & 0.434 $\pm$ 0.014 & 0.435 $\pm$ 0.015 & 0.444 $\pm$ 0.023\\
Abs Reward, $\beta$=-0.2 & 0.422 $\pm$ 0.013 & 0.424 $\pm$ 0.011 & 0.426 $\pm$ 0.012 & 0.433 $\pm$ 0.016 \\
Abs Reward, $\beta$=-0.1 & 0.414 $\pm$ 0.009 & 0.416 $\pm$ 0.008 & 0.418 $\pm$ 0.009 & 0.426 $\pm$ 0.014 \\
\bottomrule
\end{tabular}
\label{table:errors_in_nats}
\end{table}

We can see that:
\begin{itemize}[leftmargin=2em,topsep=0pt,partopsep=1ex,parsep=0ex]
\item RL with the rejection-based reward finds architectures that are within the 75M \#FLOPs limit.
\item When $|\beta|$ is large, the architectures found by RL with the Abs Reward are within the 75M \#FLOPs limit, but they are inferior in quality (have larger test errors). 
\item When $|\beta|$ is small, the architectures found by RL with the Abs Reward are similar or better in quality, but they exceed the 75M \#FLOPs limit by >5\%.
\end{itemize}
These observations are similar to what we saw in the toy example, showing the rejection-based reward outperforms in tasks from both tabular and vision domains.

Taking a closer look at the architectures that are Pareto-optimal and those found by each RL reward, we can see that:
\begin{itemize}[leftmargin=2em,topsep=0pt,partopsep=1ex,parsep=0ex]
\item The Pareto-optimal architectures often have bottleneck structures like their counterparts in tabular tasks; examples include (in NATS-Bench format, and same below) 8:32:16:40:48, 8:56:40:56:64, and 16:56:64:64:56 that have a bottleneck in their first channel.
\item Bottleneck structures occur less often in the architectures found by RL with the Abs Reward. Examples include 24:24:64:40:32, 24:40:56:40:48, and 32:48:40:32:32.
The architectures found by RL with the rejection-based reward (our method) have more similar appearances as the Pareto-optimal architectures; examples include 8:56:64:64:64, 8:40:56:64:56, and 16:64:48:64:64.
\end{itemize}
This means the co-adaptation problem (described in Section~\ref{sec:intro} of the main paper) also occurs in RL-based NAS with resource-aware rewards in the image domain.

\section{Difficulty of hyperparameter tuning}
\label{appsec:hyperparameter_tuning}
Hyperparameter tuning has always been a headache for machine learning.
In the design of NAS approaches, the hope is that the NAS hyperparameters are much easier to tune than the architectures NAS search over.
We denote the RL learning rate and the number of MC samples by $\eta$ and $N$, respectively.
The three resource-aware rewards (in MnasNet and TuNAS) have both $\eta$ and $\beta$ as hyperparameters; our TabNAS with the rejection-based reward has $\eta$ and $N$ to tune.

\begin{figure}
\centering
\begin{minipage}[t]{.55\linewidth}
\centering
\subfigure[Tuning $\beta$ in Abs Reward]{\label{fig:beta_tuning}\includegraphics[width=.25\linewidth]{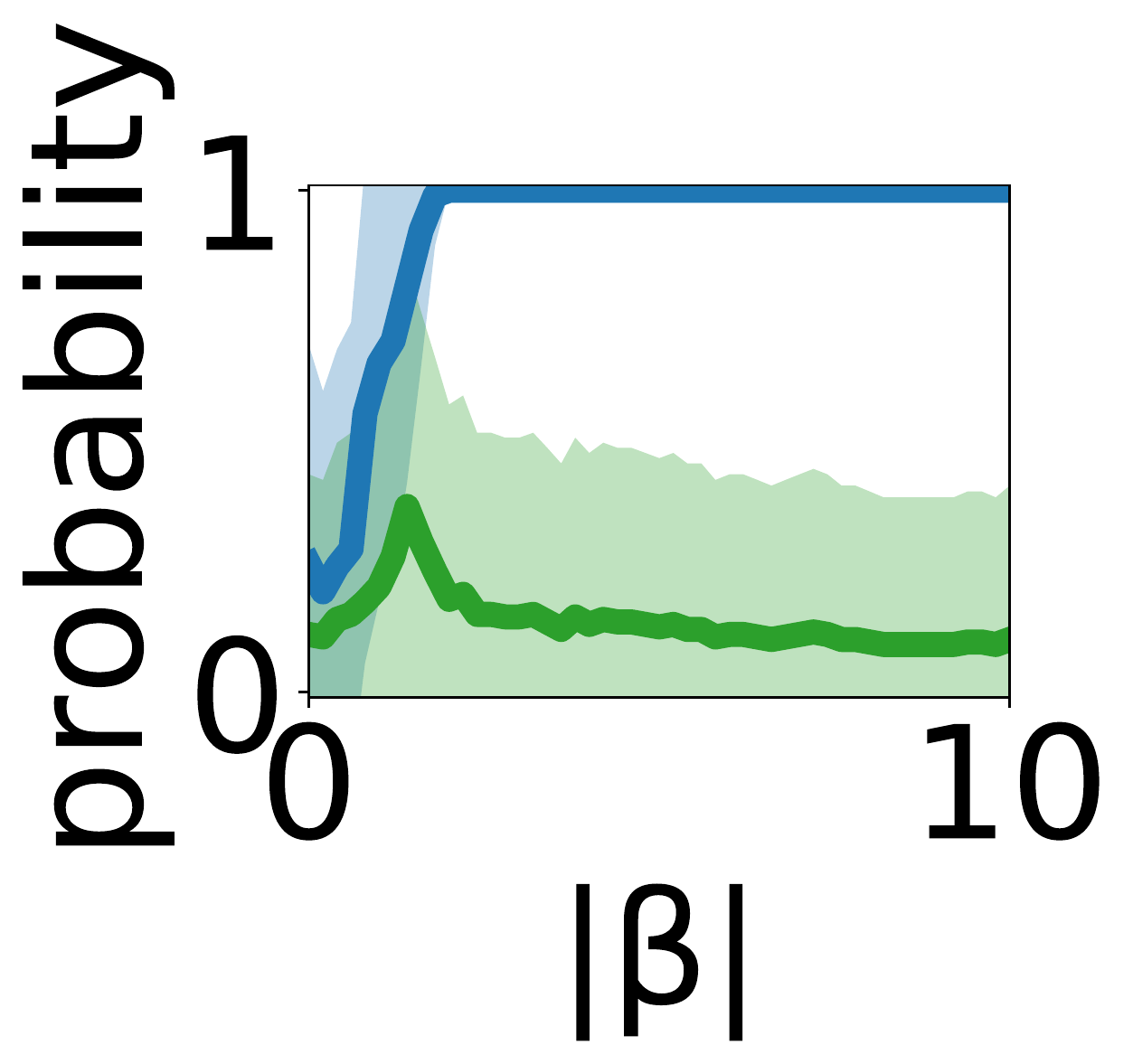}}
\hspace{.01\linewidth}
\subfigure[Tuning $N$ in rejection]{\label{fig:N_tuning}\includegraphics[width=.25\linewidth]{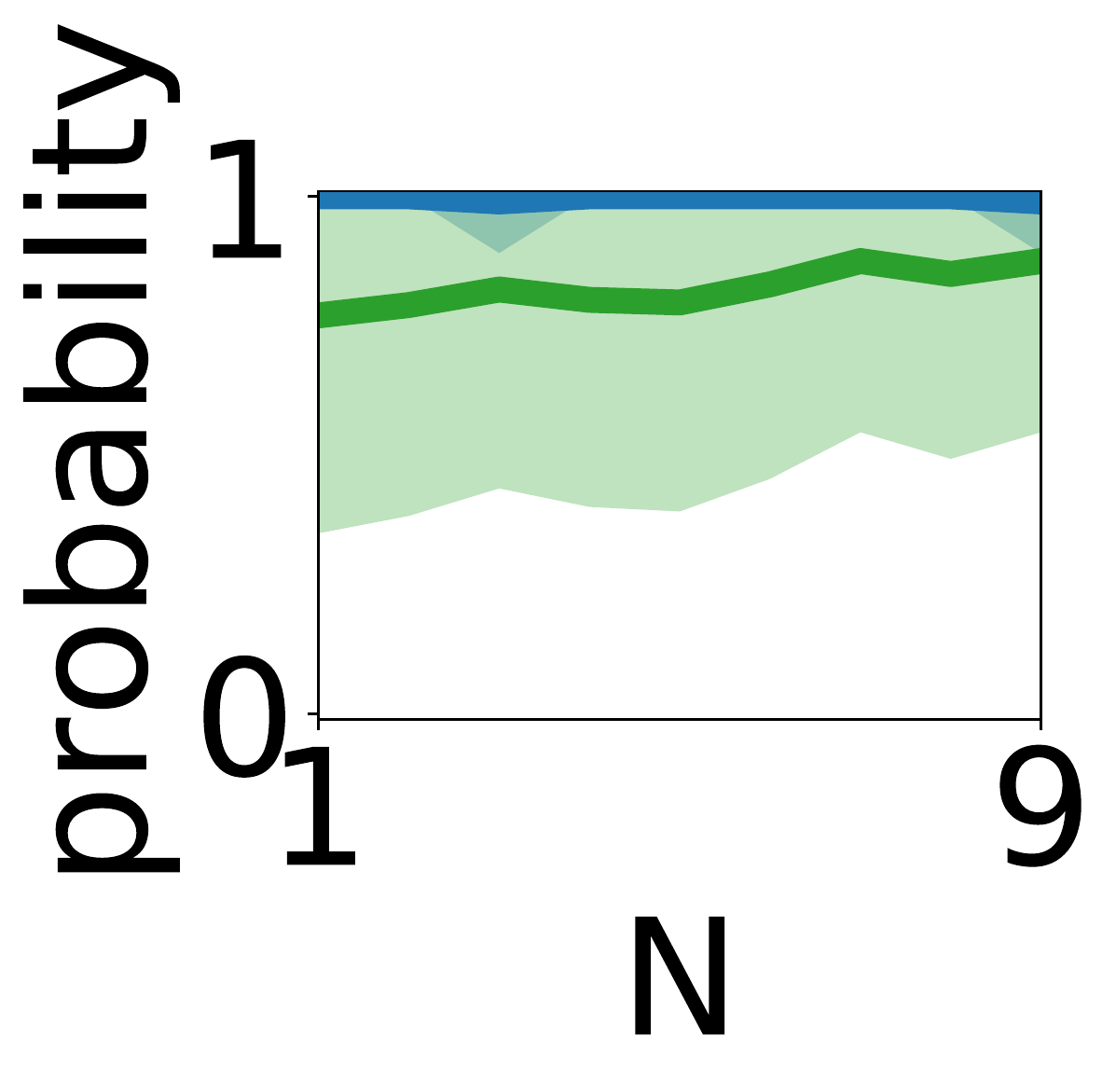}}
\stackunder[1pt]{\includegraphics[width=.35\linewidth]{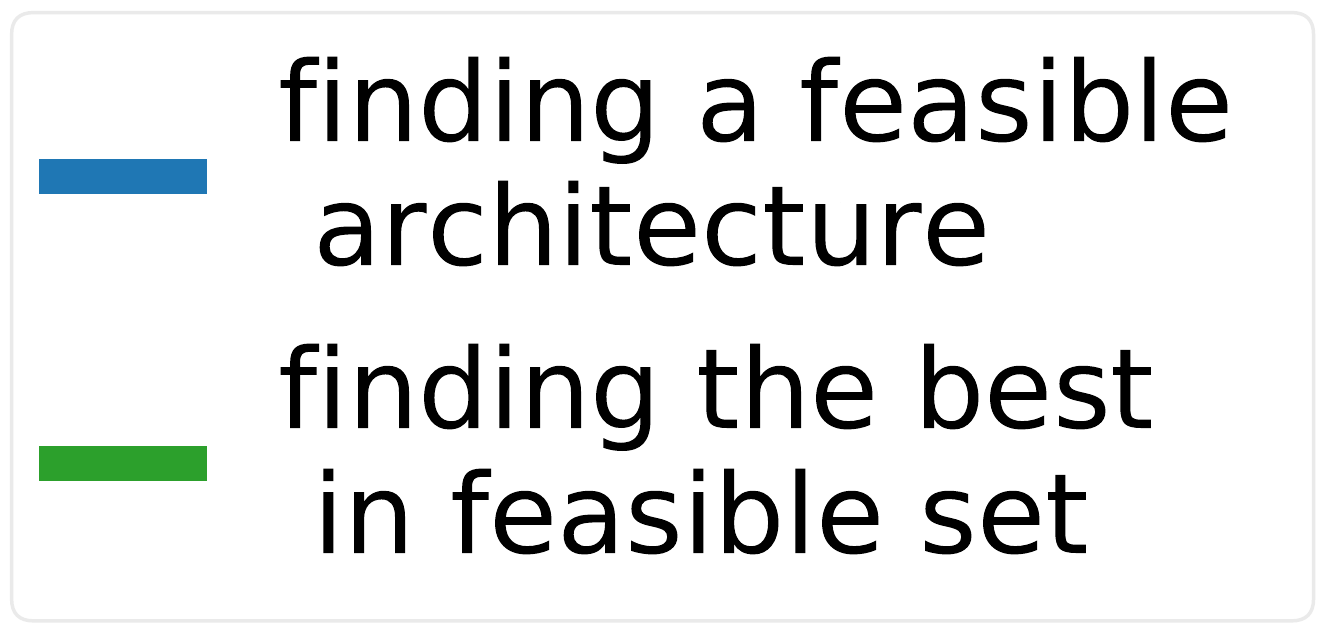}}{}

\vspace{-.5em}
\caption{\textbf{Tuning $\beta$ and $N$ on the toy example (Figure~\ref{fig:toy_example})}: the number of MC samples $N$ in rejection-based reward is easier to tune than $\beta$ in Abs Reward, and is easier to succeed.
The lines and shaded regions are mean and standard deviation across 200 independent runs, respectively. 
}
\label{fig:hyperparameter_tuning}
\end{minipage}
\hspace{.02\linewidth}
\begin{minipage}[t]{.4\linewidth}
\centering
\subfigure{\includegraphics[width=\linewidth]{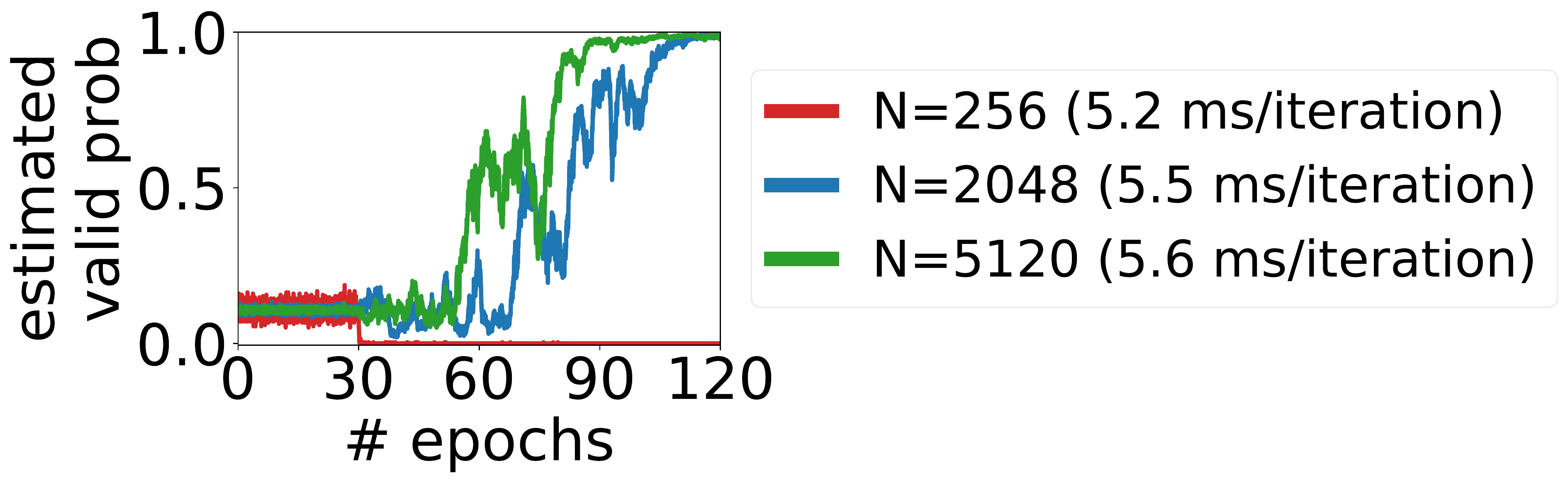}}
\vspace{-1em}
\caption{\textbf{Tuning $N$ on Criteo}: the change of $\widehat{\probP}(V)$ when the number of Monte-Carlo samples $N$ is 256, 2,048 or 5,120, and the time taken for each iteration. 
We show results with RL learning rate $\eta=0.005$; those other $\eta$ values have similar failure patterns.
}
\label{fig:tuning_N}
\end{minipage}
\end{figure}

\subsection{Resource hyperparameter $\beta$}
$\beta$ is difficult to tune in experiments: the best value varies by dataset and lies in the middle of its search space. 
Since $\beta<0$, we discuss its absolute value.
In a NAS search space, the architecture that is feasible and can match the reference performance often has the number of parameters that is more than $98\%$ of the reference.
A too small $|\beta|$ is not powerful enough to enforce the resource constraint, in which case NAS finds an architecture that is far from the target number of parameters and makes the search nearly unconstrained (e.g., the Abs Reward with $|\beta| = 1$ in the toy example, shown in Figure~\ref{fig:toy_example} and towards the left end in Figure~\ref{fig:beta_tuning}).
A too large $|\beta|$ severely penalizes the violation of the resource constraint, in which case the RL controller would always give an architecture close to the reference, with much bias (e.g., the Abs Reward with $|\beta| = 2$ in Figure~\ref{fig:toy_example}, and towards the right end in Figure~\ref{fig:beta_tuning}). 
Thus practitioners seek a medium $|\beta|$ in hyperparameter tuning to both obey the resource constraint and achieve a better result.
In our experiments, such ``appropriate'' medium values vary largely across datasets: 1 on Criteo with the 32-144-24 reference architecture (41,153 parameters), 2 on Volkert with the 48-160-32-144 reference architecture (27,882 parameters), and 25 on Aloi with the 64-192-48-32 reference architecture (64,568 parameters).  

\subsection{RL learning rate $\eta$}
The RL learning rate $\eta$ is easier to tune and more generalizable across datasets than $\beta$.
With a large $\eta$, the RL controller quickly converges right after the first $25\%$ epochs of layer warmup; with a small $\eta$, the RL controller converges slowly or may not converge, although there may still be enough signal from the layerwise probabilities to get the final result. 
It is thus straightforward to tune $\eta$ by observing the convergence behavior of sampling probabilities. 
In our experiments, the appropriate value of $\eta$ does not significantly vary across tasks: a constant $\eta \in [0.001, 0.01]$ is appropriate for all datasets and all number of parameter limits.

\subsection{Number of MC samples $N$}
\label{appsec:hyperparameter_tuning_N}
The number of MC samples $N$ is also easier to tune than $\beta$.
Resource permitting, $N$ is the larger, the better (Figure~\ref{fig:N_tuning}), so that $\probP(V)$ can be better estimated. 
When $N$ is too small, the MC sampling has a high chance of missing the valid architectures in the search space, and thus incurs large bias and variance for the estimate of $\nabla \log[\probP(y \given y \in V)]$.
In such cases, $\widehat{\probP}(V)$ may miss all valid architectures at the beginning of RL and quickly converge to 0.
$\widehat{\probP}(V)$ being equal or close to 0 is a bad case for our rejection-based algorithm: the single-step RL objective $J(y)$ that has a $-\log(\widehat{\probP}(V))$ term grows extremely large and gives an explosive gradient to stuck the RL controller in the current choice.
Consequently, the criterion for choosing $N$ is to choose the largest that can afford, and hopefully, at least choose the smallest that can make $\widehat{\probP}(V)$ steadily increase during RL.
Figure~\ref{fig:tuning_N} shows the changes of $\widehat{\probP}(V)$ on Criteo with the 32-144-24 reference in the search space of 8,000 architectures at three $N$ values. 
The NAS succeeds when $N \geq 2048$, same as the threshold that makes $\widehat{\probP}(V)$ increase.

Overall, the RL controller with our rejection-based reward has hyperparameters that are easier to tune than with resource-aware rewards in MnasNet and TuNAS.

\section{Ablation studies}
\label{appsec:ablation}
We do the ablation studies on Criteo with the 32-144-24 reference. 
The behavior on other datasets with other reference architectures are similar.

\myparagraph{Whether to use $\widehat{\probP}(V)$ instead of $\probP(V)$}
\label{sec:ablation_whether_to_estimate}
The Monte-Carlo (MC) sampling estimates $\probP(V)$ with $\widehat{\probP}(V)$ to save resources.
Such estimations are especially efficient when the sample space is large.
Empirically, the $\widehat{\probP}(V)$ estimated with enough MC samples (as described in Appendix~\ref{appsec:hyperparameter_tuning}) enables the RL controller to find the same architecture as $\probP(V)$, because the $\widehat{\probP}(V)$ estimated with a large enough number of samples is accurate enough (e.g., Figure~\ref{fig:valid_prob} and~\ref{fig:valid_prob_criteo_3_layers}).

\myparagraph{Whether to skip infeasible architectures in weight updates}
\label{sec:ablation_not_skipping_infeasible_in_training}
In each iteration of one-shot training and REINFORCE (Appendix~\ref{appsec:pseudocode} Algorithm~\ref{alg:reinforce}) with the rejection mechanism (Appendix~\ref{appsec:pseudocode} Algorithm~\ref{alg:mc}), we train the weights in the sampled child network $x$ regardless of whether $x$ is feasible.
Instead, we may update the weights only when $x$ is feasible, in a similar rejection mechanism as the RL step.
We find this mechanism may mislead the search because of insufficiently trained weights: the rejection-based RL controller can still find qualitatively the best architectures on Criteo with the 32-144-24 or 48-240-24-256-8 reference, but fails with the 48-128-16-112 reference.
In the latter case, although the RL controller still finds architectures with bottleneck structures (e.g., 32-384-8-144), the first layer sizes of the found architectures are much smaller, leading to suboptimal performance.

\myparagraph{Whether to differentiate through $\widehat{\probP}(V)$}
\label{sec:ablation_no_differentiation_through_PV}
Recall that REINFORCE with rejection has the objective
\begin{equation*}
    J(y) = \stopgrad{Q(y) - \overline{Q}} \cdot \log \left[ \probP(y) / \probP(V) \right].
\end{equation*}
To update the RL controller's logits, we compute $\nabla J(y)$, which requires a differentiable approximation of $\probP(V)$.
From a theoretical standpoint, omitting the extra term $\probP(V)$ -- or using a non-differentiable approximation -- will result in biased gradient estimates. 
Empirically, we ran experiments with multiple variants of our algorithm where we omitted the term $\probP(V)$, but found that the quality of the searched architectures was significantly worse.

\begin{figure}[H]
\centering
\subfigure[Layer 2, not skipping infeasible in training]{\label{fig:criteo_3_layers_rejection_sampling_prob_heatmap_layer_2_no_diff_no_skipping}\includegraphics[width=.21\linewidth]{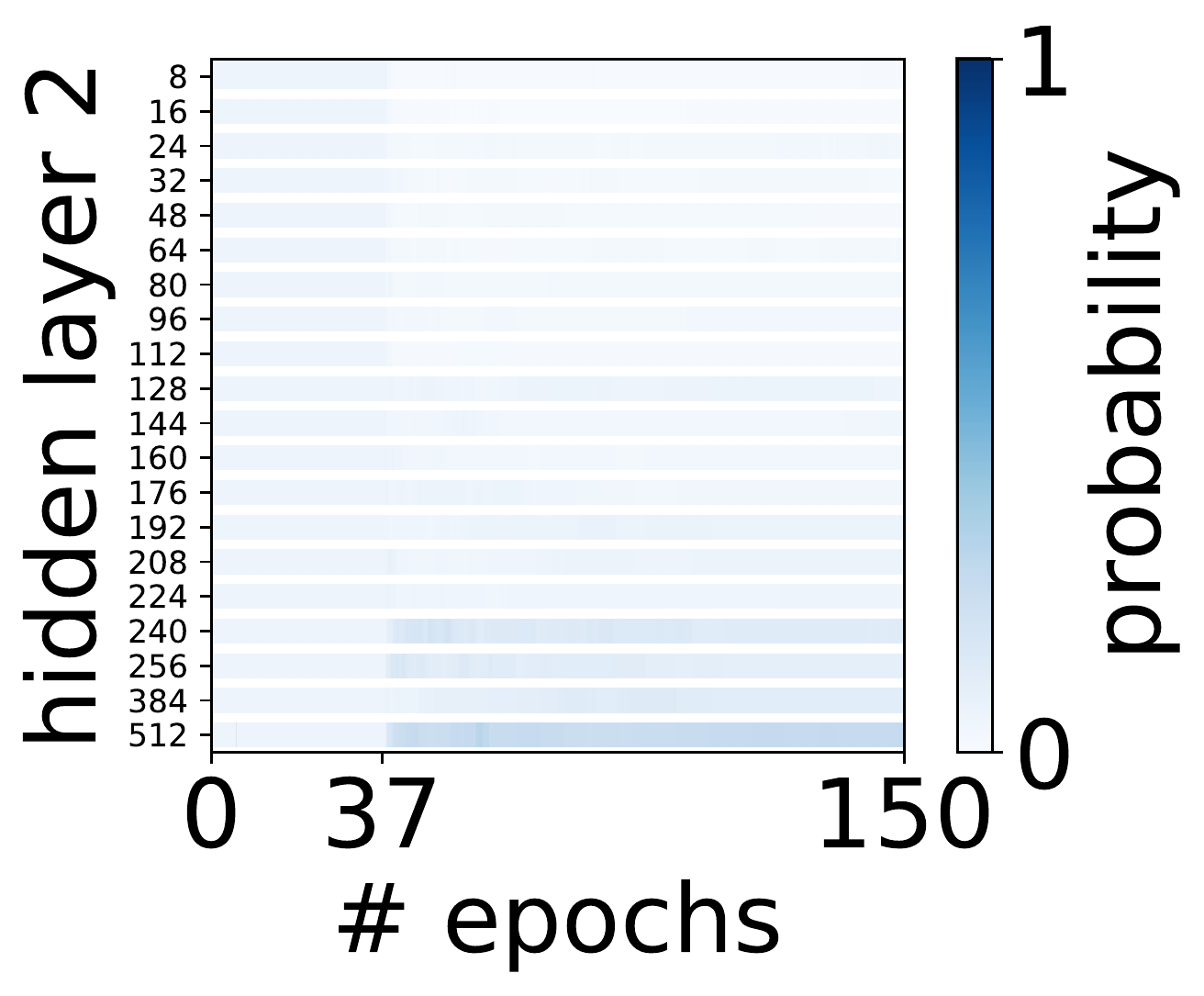}}
\hspace{.03\linewidth}
\subfigure[Valid probabilities, not skipping infeasible in training]{\label{fig:valid_prob_no_differentiation_no_skipping}\includegraphics[width=.21\linewidth]{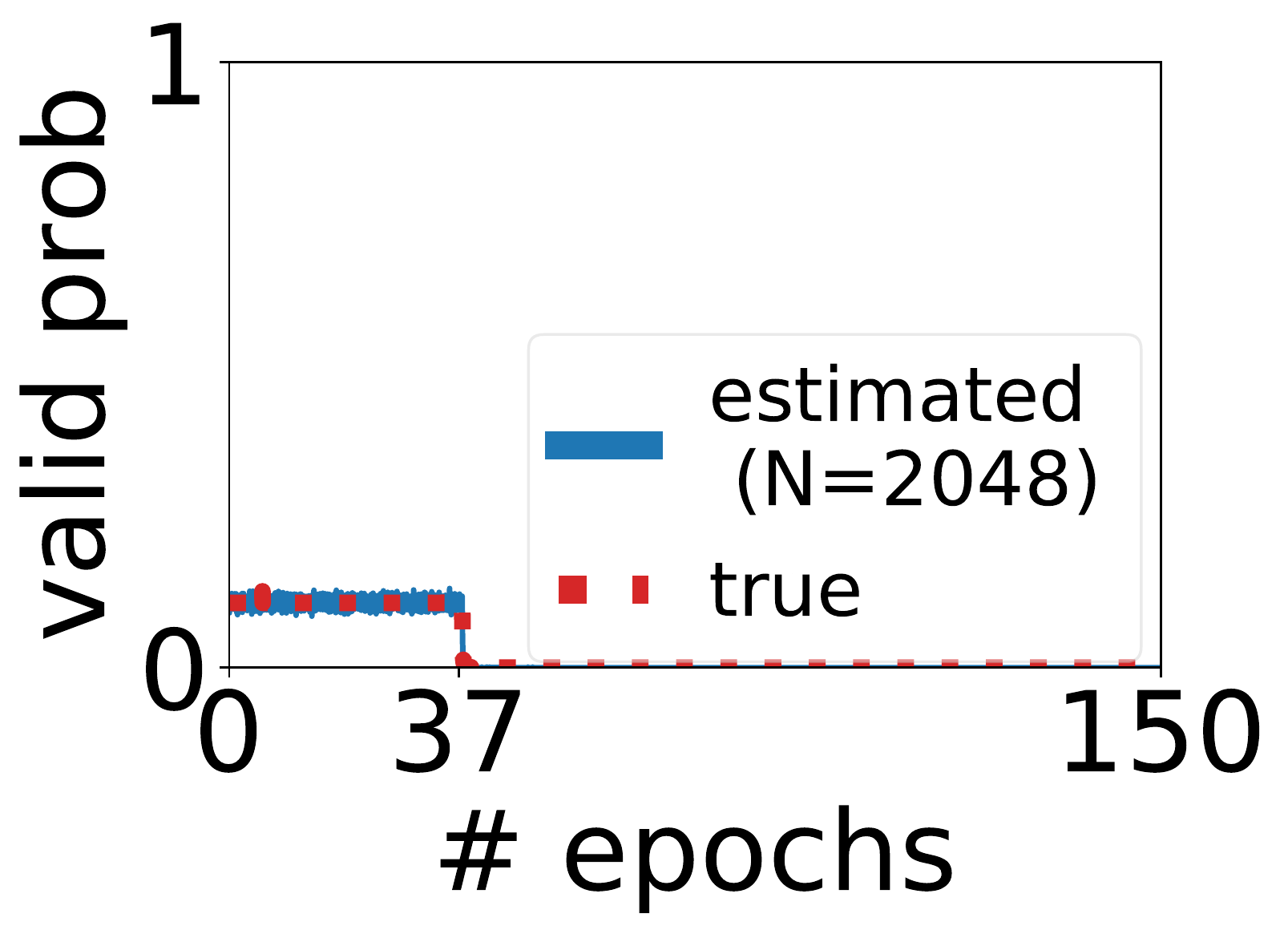}}
\hspace{.01\linewidth}
\subfigure[Layer 2, skipping infeasible in training]{\label{fig:criteo_3_layers_rejection_sampling_prob_heatmap_layer_2_no_diff_with_skipping}\includegraphics[width=.21\linewidth]{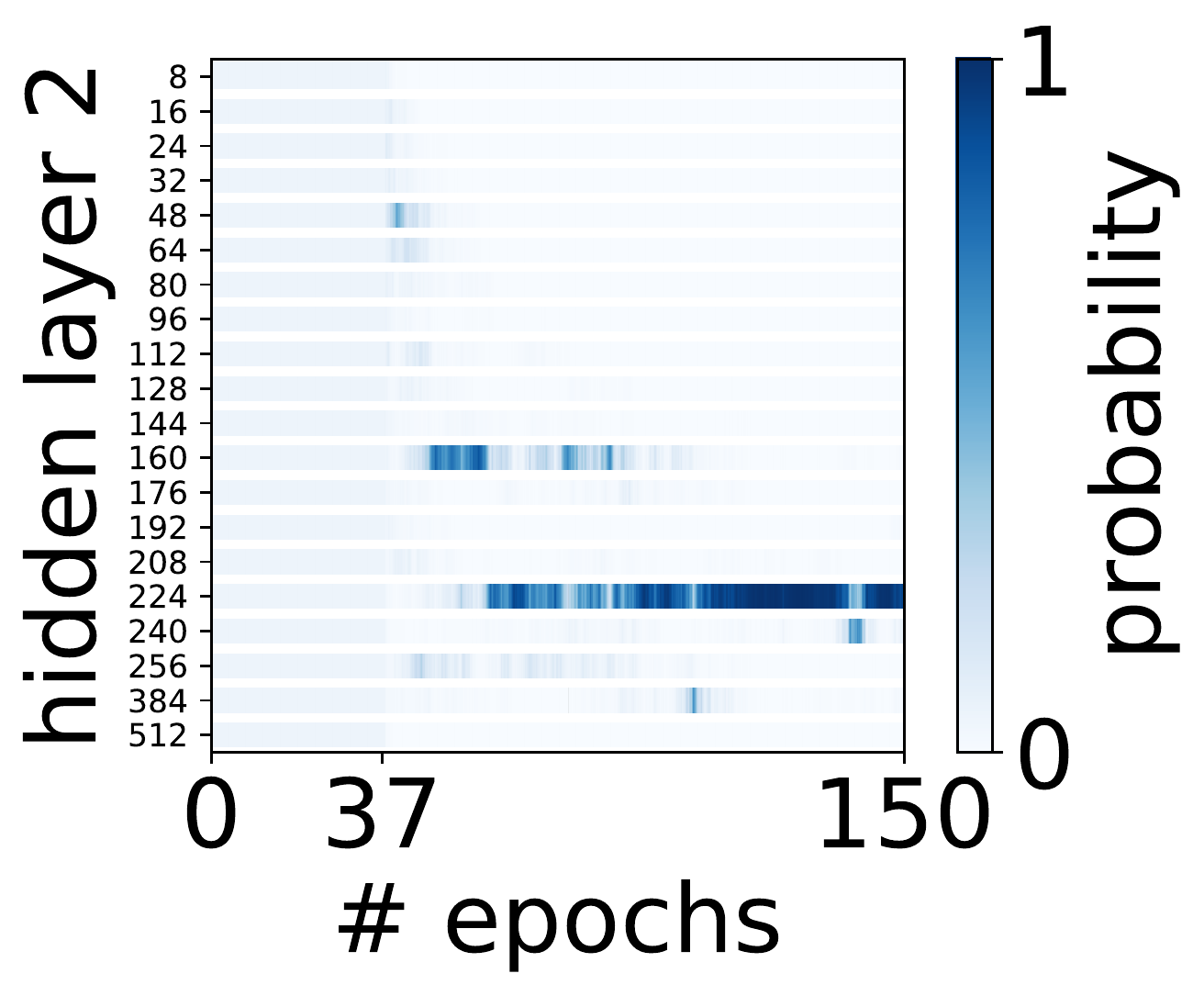}}
\hspace{.01\linewidth}
\subfigure[Valid probabilities, skipping infeasible in training]{\label{fig:valid_prob_no_differentiation_with_skipping}\includegraphics[width=.23\linewidth]{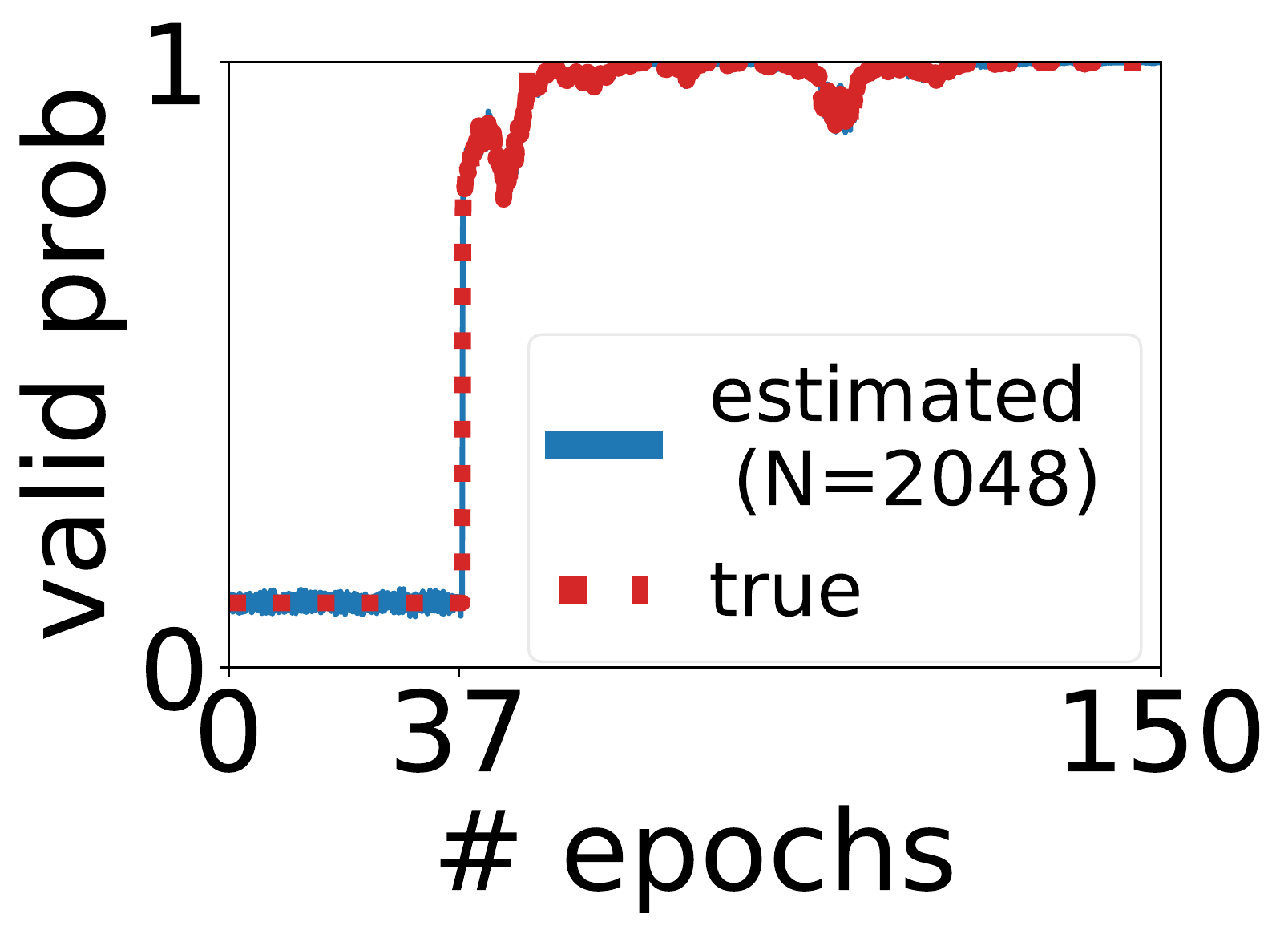}}

\caption{\textbf{Failure cases in ablation when $\widehat{\probP}(V)$ is non-differentiable}.
We show results with RL learning rate $\eta=0.005$; those under other $\eta$ values are similar.
}
\label{fig:ablation_non_differentiable}
\end{figure}

Below are our experimental findings:
\begin{itemize}[leftmargin=2em,topsep=0pt,partopsep=1ex,parsep=0ex]
\item In the case that we do not skip infeasible architectures in weight updates, the largest hidden layer sizes may gain and maintain the largest sampling probabilities soon after RL starts.
This is because most architectures in the 3-layer Criteo search space are above the number of parameters limit 41,153. 
When RL starts, the sampled feasible architectures underperform the moving average, thus their logits are severely penalized, making the logits of the infeasible architectures (which often have wide hidden layers) quickly dominate (Figure~\ref{fig:criteo_3_layers_rejection_sampling_prob_heatmap_layer_2_no_diff_no_skipping}).
Accordingly, the (estimated) valid probability 
$\probP(V)$ (or $\widehat{\probP}(V)$) quickly decrease to 0 (Figure~\ref{fig:valid_prob_no_differentiation_no_skipping}), and the RL controller gets stuck (as described in Appendix~\ref{appsec:hyperparameter_tuning_N}) in these large choices for hidden layer sizes.
\item In the case that we skip infeasible architectures in both weight and RL updates, the RL controller eventually picks feasible architectures with bottleneck structures, but the found architectures are almost always suboptimal: when RL starts, the controller severely boosts the logits of the sampled feasible architectures without much exploration in the search space, and quickly gets stuck there. 
For example, the search in Figure~\ref{fig:criteo_3_layers_rejection_sampling_prob_heatmap_layer_2_no_diff_with_skipping}) finds 24-384-16 (40,449 parameters) that is feasible but suboptimal; $\probP(V)$ and $\widehat{\probP}(V)$ quickly increase to 1 after RL starts (Figure~\ref{fig:valid_prob_no_differentiation_with_skipping}).
\end{itemize}

\myparagraph{Strategy for choosing the final architecture after search}
\label{sec:ablation_choosing_final_arch}
When RL finishes, instead of biasing towards architectures with more parameters (Appendix~\ref{appsec:pseudocode} Algorithm~\ref{alg:sample_to_get_solution}), we may also bias towards those that are feasible and have larger sampling probabilities.
We find that when the final distributions are less deterministic, the architectures found by the latter strategy to perform worse: for example, the top 3 feasible architectures found with the final distribution in Figure~\ref{fig:sampling_prob_and_valid_prob_criteo_3_layers_rejection} are 32-128-16, 32-160-16 and 32-128-8, and they are all inferior to 32-144-24.

\section{Proofs}
\label{appsec:proofs}
\subsection{$\widehat{\probP}(V)$ is an unbiased and consistent estimate of $\probP(V)$}
Within the search space $S$, recall the definitions of $\probP(V)$ and $\widehat{\probP}(V)$:

\begin{itemize}[leftmargin=2em,topsep=0pt,partopsep=1ex,parsep=0ex]
\item $\probP(V) = \sum\limits_{z^{(i)} \in S} p^{(i)} \indicator{z^{(i)} \in V}$
\item $\widehat{\probP}(V) = \frac{1}{N} \sum\limits_{k \in [N],z^{(k)} \in V} \frac{p^{(k)}}{q^{(k)}} = \frac{1}{N} \sum\limits_{k \in [N]} \frac{p^{(k)}}{q^{(k)}} \indicator{z^{(k)} \in V}$
\end{itemize}

\myparagraph{Unbiasedness}
With $N$ architectures sampled from the proposal distribution $q$, we take the expectation with respect to $N$ sampled architectures:
\begin{equation}
\begin{aligned}
\Expect [\widehat{\probP}(V)] & = \frac{1}{N} \Expect \Bigg[\sum_{k \in [N],z^{(k)} \in V} \frac{p^{(k)}}{q^{(k)}}\Bigg] \\
& = \frac{1}{N} \Expect \Bigg[\sum_{k \in [N]} \frac{p^{(k)}}{q^{(k)}} \indicator{z^{(k)} \in V} \Bigg]\\
& = \frac{1}{N} \sum_{k \in [N]} \Expect \Bigg[\frac{p^{(k)}}{q^{(k)}} \indicator{z^{(k)} \in V}\Bigg], \\
\end{aligned}
\nonumber
\end{equation}
in which each summand
\begin{equation}
\begin{aligned}
\Expect \Bigg[\frac{p^{(k)}}{q^{(k)}} \indicator{z^{(k)} \in V}\Bigg] & = \sum\limits_{z^{(k)} \in S} q^{(k)} \frac{p^{(k)}}{q^{(k)}} \indicator{z^{(k)} \in V}\\
& = \probP(V),
\end{aligned}
\nonumber
\end{equation}
Thus $\Expect [\widehat{\probP}(V)] = \probP(V)$.

\myparagraph{Consistency}
We first show the variance of $\probP(V)$ converges to 0 as the number of MC samples $N \rightarrow \infty$. 
Because of independence among samples, 
\begin{equation}
\begin{aligned}
\Var [\widehat{\probP}(V)] & = \frac{1}{N} \sum_{k \in [N]} \Var \Bigg[\frac{p^{(k)}}{q^{(k)}} \indicator{z^{(k)} \in V}\Bigg],
\end{aligned}
\nonumber
\end{equation}
in which each summand
\begin{equation}
\begin{aligned}
\Var \Bigg[\frac{p^{(k)}}{q^{(k)}} \indicator{z^{(k)} \in V}\Bigg] & = \Expect \Big[ \frac{p^{(k)}}{q^{(k)}} \indicator{z^{(k)} \in V} - \probP(V)\Big]\\
& = \sum\limits_{z^{(k)} \notin V} q^{(k)} \probP(V)^2 + \sum_{z^{(k)} \in V} q^{(k)} \Big[\frac{p^{(k)}}{q^{(k)}} - \probP(V)\Big]^2\\
& = -\probP(V)^2 + \sum_{z^{(k)} \in V} \frac{(p^{(k)})^2}{q^{(k)}},\\
\end{aligned}
\label{eq:single_summand_variance}
\end{equation}
thus the variance
\begin{equation}
\begin{aligned}
\Var [\widehat{\probP}(V)] & = \frac{1}{N} \sum_{k \in [N]} \Var \Big[\frac{p^{(k)}}{q^{(k)}} \indicator{z^{(k)} \in V}\Big] \\
& \frac{1}{N} \Big[-\probP(V)^2 + \sum_{z^{(k)} \in V} \frac{(p^{(k)})^2}{q^{(k)}}\Big],\\
\end{aligned}
\nonumber
\end{equation}
which goes to 0 as $N \rightarrow \infty$.
It worths noting that when we set $q = \stopgrad{p}$, the single-summand variance (Equation~\ref{eq:single_summand_variance}) becomes $\probP(V) - \probP(V)^2$, which is the variance of a Bernoulli distribution with mean $\probP(V)$.

The Chebyshev's Inequality states that for a random variable $X$ with expectation $\mu$, for any $a > 0$, $\probP (|X - \mu| > a) \leq \frac{\Var(X)}{a^2}$.
Thus $\lim\limits_{N \rightarrow \infty} \Var(X) = 0$ implies that $\lim \limits_{N \rightarrow \infty}\probP (|X - \mu| > a)=0$ for any $a > 0$, indicating consistency.

\subsection{$\nabla \log [\probP(y) / \widehat{\probP}(V)]$ is a consistent estimate of $\nabla \log[\probP(y \given y \in V)]$}
Since $\probP(y \given y \in V) = \frac{\probP(y)}{\probP(V)}$, we show $\plim\limits_{N \rightarrow \infty} \nabla \log \widehat{\probP}(V) = \nabla \log \probP(V)$ below to prove consistency, in which $\plim\limits_{N \rightarrow \infty}$ denotes convergence in probability.

Recall $p^{(i)}$ is the probability of sampling the $i$-th architecture $z^{(i)}$ within the search space $S$, and the definitions of $\probP(V)$ and $\widehat{\probP}(V)$ are:
\begin{itemize}[leftmargin=2em,topsep=0pt,partopsep=1ex,parsep=0ex]
\item $\probP(V) = \sum\limits_{z^{(i)} \in S} p^{(i)} \indicator{z^{(i)} \in V}$,
\item $\widehat{\probP}(V) = \frac{1}{N} \sum\limits_{k \in [N],z^{(k)} \in V} \frac{p^{(k)}}{q^{(k)}} = \frac{1}{N} \sum\limits_{k \in [N]} \frac{p^{(k)}}{q^{(k)}} \indicator{z^{(k)} \in V}$, in which each $p^{(k)}$ is differentiable with respect to all logits $\{\ell_{ij}\}_{i \in [L], j \in [C_i]}$.
\end{itemize}

Thus we have
\begin{equation}
\begin{aligned}
\plim_{N \rightarrow \infty} \widehat{\probP}(V) & = \plim_{N \rightarrow \infty} \frac{1}{N} \sum\limits_{k \in [N]} \frac{p^{(k)}}{q^{(k)}} \indicator{z^{(k)} \in V} \\
& = \frac{1}{N} \sum_{z^{(k)} \in S}\frac{p^{(k)}}{q^{(k)}} \indicator{z^{(k)} \in V} N q^{(k)}\\
& = \sum_{z^{(k)} \in S} p^{(k)} \indicator{z^{(k)} \in V} = \probP(V),\\ 
\end{aligned}
\nonumber
\end{equation}

and
\begin{equation}
\begin{aligned}
\plim_{N \rightarrow \infty} \nabla \widehat{\probP}(V) & = \plim_{N \rightarrow \infty} \frac{1}{N} \sum\limits_{k \in [N]} \frac{\nabla p^{(k)}}{q^{(k)}} \indicator{z^{(k)} \in V} \\
& = \frac{1}{N} \sum_{z^{(k)} \in S}\frac{\nabla p^{(k)}}{q^{(k)}} \indicator{z^{(k)} \in V} N q^{(k)}\\
& = \sum_{z^{(k)} \in S} \nabla p^{(k)} \indicator{z^{(k)} \in V} = \nabla \probP(V).\\ 
\end{aligned}
\nonumber
\end{equation}

Together with the condition that $\probP(V) > 0$ (the search space contains at least one feasible architecture), we have the desired result for consistency as $\plim\limits_{N \rightarrow \infty} \nabla \log \widehat{\probP}(V) = \plim\limits_{N \rightarrow \infty} \frac{\nabla \widehat{\probP}(V)}{\widehat{\probP}(V)} =  \frac{\plim\limits_{N \rightarrow \infty} \nabla \widehat{\probP}(V)}{\plim\limits_{N \rightarrow \infty} \widehat{\probP}(V)} = \frac{\nabla \probP(V)}{\probP(V)} = \nabla \log \probP(V)$, in which the equalities hold due to the properties of convergence in probability.

\end{document}